\documentclass[12pt]{article}

\newcommand{\blind}{1}
\usepackage{floatpag}
\usepackage{amsmath}
\usepackage{amssymb}
\usepackage{bbm}
\usepackage{amsbsy}
\usepackage{graphicx}
\usepackage{caption}
\usepackage{amsfonts}
\usepackage{bm}
\usepackage{url}
\usepackage{natbib}
\usepackage{enumerate}
\usepackage{booktabs}
\usepackage{multirow}
\usepackage{makecell}
\usepackage{mathtools}
\usepackage{dsfont}
\usepackage{setspace}
\usepackage{threeparttable}
\usepackage{tablefootnote}
\usepackage{adjustbox}
\usepackage[normalem]{ulem}
\usepackage[ruled,linesnumbered]{algorithm2e}
\newcolumntype{V}{!{\vrule width 1pt}}
\mathtoolsset{showonlyrefs}  
\usepackage{placeins}
\usepackage{tocloft}
\usepackage[title]{appendix}
\usepackage{textcomp}
\usepackage{eufrak}
\usepackage{hyperref}
\hypersetup{colorlinks=true,linkcolor=blue, citecolor=blue, linktocpage}
\allowdisplaybreaks

\makeatletter
\newcommand*{\Rom}[1]{\expandafter\@slowromancap\romannumeral #1@}
\makeatother
\newcommand*{\rom}[1]{\romannumeral #1}

\newcommand{\tp}{\mathbb{P}}

\newcommand{\te}{\mathbb{E}}

\newcommand{\btheta}{\bm{\theta}}

\newcommand{\bbeta}{\bm{\beta}}
\newcommand{\hbbeta}{\hat{\bm{\beta}}}

\newcommand{\infnorm}[1]{\|#1\|_{\infty}}
\newcommand{\twonorm}[1]{\|#1\|_{2}}
\newcommand{\onenorm}[1]{\|#1\|_{1}}

\newcommand{\norma}[1]{\left|#1\right|}
\newcommand{\norm}[1]{|#1|}
\newcommand{\bx}{\bm{x}}
\newcommand{\bw}{\bm{w}}
\newcommand{\blambda}{\bm{\lambda}}
\newcommand{\hblambda}{\hat{\bm{\lambda}}}
\newcommand{\hlambda}{\hat{\lambda}}
\newcommand{\balpha}{\bm{\alpha}}
\newcommand{\hpi}{\hat{\pi}}
\newcommand{\hbpi}{\hat{\bm{\pi}}}
\newcommand{\bpi}{\bm{\pi}}
\newcommand{\hphi}{\hat{\phi}}

\newcommand{\bmu}{\bm{\mu}}

\newcommand{\argmax}{\mathop{\arg\max}}
\newcommand{\argmin}{\mathop{\arg\min}}
\newcommand{\mball}[2]{\mathcal{B}_{#1}(#2)}
\newcommand{\clmball}[2]{\bar{\mathcal{B}}_{#1}(#2)}
\newcommand{\convp}{\xrightarrow[]{p}}
\newcommand{\inlineeqnum}{\refstepcounter{equation}~~\mbox{(\theequation)}}

\newcommand{\hL}{\widehat{L}}

\def\boxit#1{\vbox{\hrule\hbox{\vrule\kern6pt\vbox{\kern6pt#1\kern6pt}\kern6pt\vrule}\hrule}}

\newcommand\smallo{
  \mathchoice
    {{\scriptstyle\mathcal{O}}}
    {{\scriptstyle\mathcal{O}}}
    {{\scriptscriptstyle\mathcal{O}}}
    {\scalebox{.7}{$\scriptscriptstyle\mathcal{O}$}}
  }

\usepackage{multibib}
\newcites{app}{References}

\newtheorem{theorem}{Theorem}
\newtheorem{lemma}{Lemma} 
\newtheorem{proposition}{Proposition} 
\newtheorem{remark}{Remark}
\newtheorem{corollary}{Corollary}

\newtheorem{assumption}{Assumption}

\newtheorem{manualassumptioninner}{Assumption}
\newenvironment{manualassumption}[1]{%
  \renewcommand\themanualassumptioninner{#1}%
  \manualassumptioninner
}{\endmanualassumptioninner}

\begin{document}
\def\spacingset#1{\renewcommand{\baselinestretch}%
{#1}\small\normalsize} \spacingset{1}

\title{\bf Neyman-Pearson Multi-class Classification via Cost-sensitive Learning}
  \date{}
\if1\blind
{
 
  \author{Ye Tian\\
    Department of Statistics\\ Columbia University\\
    and \\
    Yang Feng \\
    Department of Biostatistics, School of Global Public Health\\New York University}
   
  \maketitle
} \fi

\if0\blind
{
  \bigskip
  \bigskip
  \bigskip

  \maketitle
} \fi

\maketitle

\begin{abstract}
Most existing classification methods aim to minimize the overall misclassification error rate. However, in applications such as loan default prediction, different types of errors can have varying consequences. To address this asymmetry issue, two popular paradigms have been developed: the Neyman-Pearson (NP) paradigm and the cost-sensitive (CS) paradigm.  Previous studies on the NP paradigm have primarily focused on the binary case, while the multi-class NP problem poses a greater challenge due to its unknown feasibility. In this work, we tackle the multi-class NP problem by establishing a connection with the CS problem via strong duality and propose two algorithms. We extend the concept of NP oracle inequalities, crucial in binary classifications, to NP oracle properties in the multi-class context.  Our algorithms satisfy these NP oracle properties under certain conditions. Furthermore, we develop practical algorithms to assess the feasibility and strong duality in multi-class NP problems, which can offer practitioners the landscape of a multi-class NP problem with various target error levels. Simulations and real data studies validate the effectiveness of our algorithms. To our knowledge, this is the first study to address the multi-class NP problem with theoretical guarantees. The proposed algorithms have been implemented in the R package \texttt{npcs}, which is available on CRAN.
\end{abstract}

\vspace{0.5cm}

\noindent
{\it Keywords: multi-class classification, Neyman-Pearson paradigm, cost-sensitive learning, duality, feasibility, confusion matrix.} 
\vfill

\newpage
\spacingset{1.83}

\addtocontents{toc}{\protect\setcounter{tocdepth}{0}}

\section{Introduction}\label{sec: introduction}
\subsection{Asymmetric classification errors and an example in loan default prediction}\label{subsec: assymetry}
Classification is one of the central tasks in machine learning, in which we train a classifier on training data to accurately predict the labels of unseen test data based on predictors. In practice, we rarely achieve a perfect classifier that can correctly classify all the unknown data. There are different types of errors that a classifier can make. In binary classification with classes 1 and 2, denote the predictor vector $X \in \mathcal{X} \subseteq \mathbb{R}^p$ and the label $Y \in \{1, 2\}$. For any classifier $\phi: \mathcal{X} \rightarrow \{1, 2\}$, we usually define type-\Rom{1} error $R_1 = \tp_{X|Y=1}(\phi(X)\neq 1)$  and type-\Rom{2} error $R_2 = \tp_{X|Y=2}(\phi(X)\neq 2)$, where $\tp_{X|Y=k}$ represents the probability measure induced by the conditional distribution of $X$ given $Y=k$, $k = 1$ or 2. The overall misclassification error can be viewed as a weighted sum of type-\Rom{1} and type-\Rom{2} errors.

In many classification approaches, classifiers are designed to minimize the overall misclassification error. However, in many scenarios, different types of errors can have varying degrees of consequences, rendering the overall misclassification error minimization inappropriate. One such example is loan default prediction, where a default borrower is denoted as class 1 and a borrower who pays the full amount on time as class 2. In this context, making a type-\Rom{1} error, i.e., misclassifying a default borrower as a non-default borrower and lending money to them, is typically more serious than making a type-\Rom{2} error, i.e., misclassifying a non-default borrower person as a default one and refusing to lend money to them. In such cases, the criterion of overall misclassification error minimization may need to be revised. Consequently, researchers developed two paradigms -- the Neyman-Pearson paradigm and the cost-sensitive learning paradigm -- to address this error asymmetry. In the following two subsections, we introduce these paradigms separately.

\subsection{Neyman-Pearson paradigm}\label{subsec: np}
The Neyman-Pearson (NP) paradigm changes the classical classification framework by prioritizing different types of errors differently. In binary classification, the NP paradigm seeks the classifier $\phi$ that solves the following optimization problem
\begin{align}
	&\min_{\phi} \quad  \tp_{X|Y=2}(\phi(X)\neq 2) \\
	&\text{s.t.} \quad \tp_{X|Y=1}(\phi(X)\neq 1) \leq \alpha_1, \label{eq: NP binary}
\end{align}
with a given target error level $\alpha_1 \in [0, 1]$. 

There have been many studies on the binary NP paradigm, and researchers have developed many useful tools to solve problem \eqref{eq: NP binary}. \cite{cannon2002learning} initiated the theoretical analysis of NP classification. \cite{scott2005neyman} proved theoretical properties of the empirical error minimization (ERM) approach, including the so-called NP oracle inequalities. \cite{scott2007performance} combined two types of errors to measure the performance under the NP paradigm. \cite{rigollet2011neyman} transformed the original problem into a convex problem through some convex surrogates. They solved the new problem and proved that the optimal classifier could successfully control the type-\Rom{1} error with high probability. \cite{tong2013plug} tackled this problem by combining the Neyman-Pearson lemma with the kernel density estimation and developed the so-called plug-in method, which enjoys the NP oracle inequalities. \cite{zhao2016neyman} extended the NP framework into the high-dimensional case via na\"ive Bayes classifier, where the number of predictors can grow with the sample size. More recently, \cite{tong2018neyman} proposed an umbrella NP algorithm that can adapt to any scoring-type classifier, including linear discriminant analysis (LDA), support vector machines (SVM), and random forests. Using order statistics and some thresholding strategy, the umbrella algorithm can provide high probability control for all classifiers under some sample size requirements. \cite{tong2020neyman} further studied both parametric and non-parametric ways to adjust the classification threshold for an LDA classifier, which were proved to solve \eqref{eq: NP binary} with NP oracle inequalities. More recently, \cite{wang2021non} introduced an LDA-based NP classifier that does not depend on sample splitting. \cite{scott2019generalized} proposed a generalized Neyman-Pearson criterion and argued that a broader class of transfer learning problems could be solved under this criterion. \cite{li2020bridging} first connected binary NP problems with CS problems and proposed a way to construct a CS classifier with type-\Rom{1} error control. \cite{xia2021intentional} applied the NP umbrella method proposed by \cite{tong2018neyman} into a social media text classification problem. \cite{li2021flexible} proposed a model-free feature ranking method based on the NP framework. The works we list may be incomplete. We refer interested readers to the survey paper by \cite{tong2016survey} and another recent paper discussing the relationship between hypothesis testing and NP binary classification by \cite{li2020statistical}.

However, all the works mentioned above primarily focus on the binary NP paradigm. In many real-world scenarios, for example, the loan default prediction problem, there may be more than two possible outcomes, such as default, fully paid, and late payment but not default.  Controlling errors under certain target levels in the multi-class scenario is a less explored yet more practically relevant problem. In this paper, we consider such a \textit{multi-class} classification problem and propose algorithms to solve it under the NP paradigm. Suppose there are $K$ classes ($K \geq 2$), and we denote them as classes $1$ to $K$. The training sample $\{(\bm{x}_i, y_i)\}_{i=1}^n$ are i.i.d. copies of $(X, Y) \subseteq \mathcal{X} \otimes \{1, \ldots, K\}$, where $\mathcal{X} \subseteq \mathbb{R}^p$. Denote $\pi_k^* = \tp(Y=k)$ and we assume $\pi_k^* \in (0, 1)$ for all $k$'s. Also denote $\bpi^* = (\pi_1^*, \ldots, \pi_K^*)^T$. To formulate a multi-class NP problem, we need to extend the two types of errors in binary classification to the multi-class case. We now introduce two possible formulations.
\begin{itemize}
	\item \cite{mossman1999three} and \cite{dreiseitl2000comparing} extended binary receiver operating characteristic (ROC) to multi-class ROC by considering $\tp_{X|Y=k}(\phi(X)\neq k|Y=k)$ as the $k$-th error rate of classifier $\phi$ for any $k \in \{1, \ldots, K\}$. Then the NP problem can be constructed to minimize a weighted sum of $\{\tp_{X|Y=k}(\phi(X)\neq k)\}_{k=1}^K$ while controlling $\tp_{X|Y=k}(\phi(X)\neq k)$ for $k \in \mathcal{A} \subseteq \{1, \ldots, K\}$.
	\item Another way is to consider the confusion matrix $\Gamma = [\Gamma_{rk}]_{K \times K}$, where $\Gamma_{rk} = \tp_{X|Y=k}(\phi(X) = r)$ for $r \neq k$ \citep{edwards2004ideal}. Then we can formulate the NP problem as minimizing a weighted sum of $\{\tp_{X|Y=k}(\phi(X)= r)\}_{r,k=1}^K$ while controlling $\Gamma_{rk}$ for $(r,k) \in \mathcal{A} \subseteq [K] \otimes [K]$.
\end{itemize}
To begin, we focus on the first formulation, which aims to minimize a weighted sum of $\{\tp_{X|Y=k}(\phi(X)\neq k)\}_{k=1}^K$ and controls $\tp_{X|Y=k}(\phi(X)\neq k)$ for $k \in \mathcal{A}$, where $\mathcal{A} \subseteq \{1, \ldots, K\}$. The more general confusion matrix control problem is more complicated and will be discussed in Section \ref{sec: general confusion matrix problem} of the supplementary materials due to space constraints. We formally present the Neyman-Pearson \textit{multi-class} classification (NPMC) problem as 
\begin{align}
	&\min_{\phi} \quad J(\phi) = \sum_{k=1}^K w_k \tp_{X|Y=k}(\phi(X)\neq k) \\
	&\text{s.t.} \quad \tp_{X|Y=k}(\phi(X)\neq k) \leq \alpha_k, \quad k \in \mathcal{A}, \label{eq: NPMC}
\end{align}
where $\phi: \mathcal{X} \rightarrow \{1, \ldots, K\}$ is a (measurable) classifier, $\alpha_k \in [0, 1]$, $w_k \geq 0$, $\sum_{k=1}^K w_k = 1$ \footnote{This is without loss of generality as we can always normalize the weights $\{w_k\}_{k=1}^K$ by $\sum_{k=1}^K w_k$.}, and $\mathcal{A} \subseteq \{1, \ldots, K\}$. The linear combination format of the objective function $J(\phi)$ is chosen for ease of interpretation. Here, $w_k$ represents the ``cost" of misclassifying an observation from class $k$. If we set $w_k = \pi_k^*$ for all $k$, then $J(\phi)$ equals the overall misclassification error rate $\tp(\phi(X) \neq Y)$. Furthermore, our analysis and proposed algorithms can be extended to the case of $J(\phi) = \max_{k \notin \mathcal{A}}\tp_{X|Y=k}(\phi(X)\neq k)$, which represents the worst performance among classes not in $\mathcal{A}$ \footnote{We thank one of the reviewers for pointing this out.}, and the details can be found in Section \ref{sec: NPMC-max supp} in the supplements.

The formulation of \eqref{eq: NPMC} is closely connected to the distributional hypothesis testing problem with a composite null hypothesis consisting of finite arguments. For example, suppose that we have collected data $X_n = (x_1, \ldots, x_n)^T \sim$ some distribution $\tp$ and we would like to test $H_0: \tp \in \{\tp^{(k)}\}_{k=1}^K \, \text{v.s. } H_1: \tp = \tp^{(K+1)}$. The goal is to find the optimal deterministic testing function $\varphi: X_n \mapsto \{0, 1\}$ that maximizes the statistical power $\tp^{(K+1)}(\varphi(X_n)=1)$ and controls the type-I error rate under level $\alpha$, i.e., $\max_{k=1:K}\tp^{(k)}(\varphi(X_n)=1) \leq \alpha$. These two problems are interconnected, and both necessitate control over multiple errors. However, there are some intrinsic differences between these two problems. First, in the hypothesis testing problem, $\tp^{(k)}$ is \textit{known}, whereas in the NP problem \eqref{eq: NPMC}, the distribution of $X$ given $Y = k$ is \textit{unknown}. Second, multiple $\tp^{(k)}$'s belong to the same null hypothesis $H_0$, inherently constituting a \textit{binary} problem. Consequently, the hypothesis testing problem is always feasible.  However, in the NP problem \eqref{eq: NPMC}, $K$ classes are distinct and are associated with potentially different target control levels $\alpha_k$'s, rendering it a \textit{multi-class} problem where feasibility is not guaranteed (as elaborated later). More comparisons between the hypothesis testing and NP problems can be found in \cite{li2020statistical}. Additional discussions will be provided in Section \ref{subsec: discussion np ht} of supplementary materials.

Previously, there have been few works on solving the NPMC problem. \cite{landgrebe2005neyman} proposed a general empirical method to solve the NPMC problem, which relies on the multi-class ROC estimation. Our work tackles the NPMC problem by linking it with the cost-sensitive learning problem (to be introduced), which is partly motivated by their approach. However, there are notable differences between our work and theirs. First, their algorithm requires a grid search to determine the appropriate cost parameters. When dealing with a large number of classes $K$ and demanding high accuracy,  the computation cost will be too high to be affordable. Despite the efficient multi-class ROC approximation via decomposition and sensitivity analysis proposed in a later work \citep{landgrebe2008efficient}, it remains somewhat restrictive without a formal connection to a cost-sensitive learning problem. Our algorithms connect the NPMC problem to cost-sensitive learning by duality and search the optimal costs in cost-sensitive learning by a direct optimization procedure, which is much more straightforward than their method. Second, their approach lacks theoretical guarantees, whereas we prove the multi-class NP oracle properties for our methods under certain conditions. More recently, \cite{ma2020quadratically} developed a regularized sub-gradient method on non-convex optimization problems, which can be applied to solve the NPMC problem with specific linear classifiers with non-convex losses. Their method is only suitable for linear classifiers with certain loss functions, while our methods are adaptable to any classification method. To our knowledge, our work is the first to solve the NPMC problem via cost-sensitive learning techniques with theoretical guarantees.

Compared to the binary NP problem \eqref{eq: NP binary}, the multi-class version \eqref{eq: NPMC} is significantly more challenging to solve. One of the major challenges lies in the fact that the binary NP problem \eqref{eq: NP binary} is always feasible (in the most extreme case, all observations can be classified to the class whose error rate is to be controlled) while the problem \eqref{eq: NPMC} can be infeasible. To provide readers with insight into how feasibility interacts with target error levels and the conditional distribution of $X$ given $Y$, let's consider a simple example: a 3-class NPMC problem with $X|Y = k \sim N(\bmu_k, \bm{I}_p)$ for $k = 1, 2, 3$, $\mathcal{A} = \{1,2\}$, and the target levels $\alpha_1$, $\alpha_2$. Even in this basic setup, characterizing the feasibility condition remains challenging because problem \eqref{eq: NPMC} encompasses all deterministic classifiers.  However, thanks to our Theorem \ref{thm: strong duality} (to be introduced in Section \ref{subsec: strong duality}), we can derive the following lemma, which explicitly provides the feasibility condition.
\begin{lemma}\label{lem: feasibility toy example}
	The 3-class NPMC problem \eqref{eq: NPMC} with $X|Y = k \sim N(\bmu_k, \bm{I}_p)$ for $k = 1, 2, 3$, $\mathcal{A} = \{1,2\}$, and the target levels $\alpha_1$, $\alpha_2 \in [0,1]$, is feasible if and only if
	\vspace*{-0.5cm}
	\begin{equation}
		\twonorm{\bmu_1 - \bmu_2} \geq \Phi^{-1}(1-\alpha_1) + \Phi^{-1}(1-\alpha_2),
	\end{equation}
	\vspace*{-1.3cm}
	
	\noindent where $\Phi^{-1}$ is the inverse CDF function of $N(0, 1)$.
\end{lemma}

We observe a trade-off between $\alpha_1$ and $\alpha_2$ given $\bmu_1$ and $\bmu_2$, indicating that we cannot make both arbitrarily small. In general, it is difficult to characterize the feasibility condition on the joint distribution of $(X, Y)$ for the NPMC problem \eqref{eq: NPMC}. 

\subsection{Cost-sensitive learning}\label{subsec: cs}
As discussed in Section \ref{subsec: assymetry}, cost-sensitive learning (CS) provides another approach to addressing asymmetric errors in classification. There are two types of cost-sensitive learning problems where the cost is associated with features or classes, respectively \citep{fernandez2018cost}. Here, we focus on the second type, where the cost is associated with different classes. \cite{ling2008cost} further divided methods dealing with this type of CS problem into two categories: direct and meta-learning methods. Direct methods design the algorithm structure for specific classifiers, e.g., support vector machines \citep{katsumata2015robust}, $k$-nearest neighbors \citep{qin2013cost}, and neural networks \citep{zhou2005training}. Meta-learning methods create a wrapper that converts an existing classifier into a cost-sensitive one. Instances of this type of approach include rescaling \citep{domingos1999metacost, zhou2010multi}, thresholding \citep{elkan2001foundations, sheng2006thresholding, tian2019thors}, and weighted-likelihood methods \citep{dmochowski2010maximum}, among others.

Similar to the multi-class NP problem, there are also two ways to formulate the multi-class CS problem. One is to consider per-class error rates $\tp_{X|Y=k}(\phi(X)\neq k|Y=k)$ for $k = 1, \ldots, K$, and the other one is to consider the confusion matrix. In this paper, we would like to connect \eqref{eq: NPMC} to the following cost-sensitive (CS) multi-class classification problem
\begin{equation}\label{eq: general cs problem}
	\min_{\phi} \quad \textup{Cost}(\phi) = \sum_{k=1}^K \pi_k^* c_k \tp_{X|Y=k}(\phi(X)\neq k),
\end{equation}
where $\phi: \mathcal{X} \rightarrow \{1, \ldots, K\}$, $\pi^*_k = \tp(Y=k)$, and $\{c_k\}_{k=1}^K$ are the costs associated with each class. The relationship between the NPMC problem with the confusion matrix control and the CS problem will be discussed in Section \ref{sec: general confusion matrix problem} of supplementary materials.

The following lemma shows that CS problem \eqref{eq: general cs problem} has an explicit solution.

\begin{lemma}\label{lemma: cs oracle classifier}
	Define classifier $\bar{\phi}^*: \bx \mapsto \argmax_k \{c_k\tp_{Y|X=\bx}(Y = k)\}$ \footnote{If there is a tie, let $\bar{\phi}^*(\bx)$ be the smallest index within the tie.}. Then $\bar{\phi}^*$ is an optimal classifier of \eqref{eq: general cs problem} in the following sense: For any classifier $\phi$, $\textup{Cost}(\bar{\phi}^*) \leq \textup{Cost}(\phi)$.
\end{lemma}

\subsection{Multi-class NP oracle properties}\label{subsec: np oracle properties}
In this section, we extend the NP oracle inequalities proposed in \cite{scott2005neyman} to the multi-class case for problem \eqref{eq: NPMC}. We call them the multi-class NP oracle properties. Algorithms satisfying these two properties satisfied are desirable. For any classifier $\phi$, we denote $R_k(\phi) = \tp_{X|Y=k}(\phi(X) \neq k)$. 

\vspace{0.2cm}
\noindent\textbf{Multi-class NP oracle properties for the NPMC problem}:\vspace{-0.2cm}
\begin{enumerate}[(i)]
	\item If the NPMC problem is feasible and has an optimal solution $\phi^*$, then the algorithm outputs a solution $\hphi$ that satisfies
		\begin{enumerate}
			\item $R_k(\hphi) \leq \alpha_k + \mathcal{O}_{\tp}(\epsilon(n))$, $\forall k\in \mathcal{A}$;
			\item $J(\hphi) \leq J(\phi^*) + \mathcal{O}_{\tp}(\epsilon_J(n))$,
		\end{enumerate}
		where $\epsilon(n)$ and $\epsilon_J(n) \rightarrow 0$ as $n \rightarrow \infty$.
	\item Denote the event that the algorithm indicates infeasibility of NPMC problem given $\{(\bx_i, y_i)\}_{i=1}^n$ as $\mathcal{G}_n$. If the NPMC problem is infeasible, then $\tp(\mathcal{G}_n) \rightarrow 1$, as $n \rightarrow \infty$.
\end{enumerate}

It is important to remark that multi-class NP oracle properties can only guarantee an ``approximate" control for problem \eqref{eq: NPMC}, in the sense that the actual error rate could fluctuate around the target level, and the scale of fluctuation vanishes with high probability as the sample size $n \rightarrow \infty$. This form is motivated from the NP oracle inequalities in the binary case used in literature (e.g., \citealp{cannon2002learning, scott2005neyman, scott2019generalized, kalan2023tight, kalan2024distribution}). Therefore, our goal is to obtain a classifier $\phi$ which can control $\tp_{X|Y=k}(\phi(X) \neq k)$ around $\alpha_k$ with high probability for all $k \in \mathcal{A}$.

\subsection{Organization}
We organize the rest of this paper as follows. In Section \ref{sec: method}, we develop two algorithms to solve the NPMC problem \eqref{eq: NPMC}, denoted as NPMC-CX (ConveX) and NPMC-ER (Empirical Risk), respectively. In Section \ref{sec: theory}, we show that NPMC-CX enjoys multi-class NP oracle properties under Rademacher classes, and NPMC-ER satisfies multi-class NP oracle properties under a broader class of models, as long as the model can fit the data well enough. We validate the effectiveness of our approaches via simulations and real data experiments in Section \ref{sec: numerical}. Section \ref{sec: discussion} summarizes our contributions and points out a few potential future research directions. Due to the page limit, some additional discussions, extra numerical results, and all the proofs are provided in the supplementary materials.

\subsection{Notations}
Before closing the introduction, we summarize the notations used throughout this paper. For any set $D$, $\norm{D}$ represents its cardinality. For any real number $a$, $\lfloor a \rfloor$ denotes the maximum integer no larger than $a$. Define the non-negative half space in $\mathbb{R}^p$ as $\mathbb{R}_+^p = \{\bx =(x_1, \ldots, x_p)^T \in (\mathbb{R} \cup \{+\infty\})^p: \min_j x_j \geq 0\}$. For a $p$-dimensional vector $\bx = (x_1, \ldots, x_p)^T$, its $\ell_2$-norm is defined as $\twonorm{\bx} = \sqrt{\sum_{j=1}^p x_j^2}$. For a $p \times p$ matrix $A$, $\lambda_{\max}(A)$ and $\lambda_{\min}(A)$ represent its maximum and minimum eigenvalues, respectively. We mean $A$ is positive-definite or negative-definite by writing $A \succ  0$ or $A \prec 0$, respectively. For a function $f: \mathcal{X} \rightarrow \mathbb{R}$ where $\mathcal{X}$ is some metric space, we define its sup-norm as $\infnorm{f} = \sup_{\bx \in \mathcal{X}}\norm{f(\bx)}$. For the empty set $\emptyset$, we define $\min_{\bx \in \emptyset}f(\bx) = +\infty$. For two non-zero real sequences $\{a_n\}_{n=1}^{\infty}$ and $\{b_n\}_{n=1}^{\infty}$, we denote $\sup_n |a_n/b_n| < \infty$ by $a_n \lesssim b_n$. For two random sequences $\{a_n\}_{n=1}^{\infty}$ and $\{b_n\}_{n=1}^{\infty}$, $a_n = \mathcal{O}_{\tp}(b_n)$ indicates that for any $\epsilon > 0$, there exists a positive constant $M$ such that $\sup_n \tp(|a_n/b_n| > M) \leq \epsilon$. We use $\tp$ and $\te$ to represent probabilities and expectations. Sometimes we add subscripts to emphasize the source of randomness. For example, $\tp_{Y|X=\bx}(Y=k)$ means the probability of $Y = k$ given $X = \bx$. $\te_X$ means the expectation is taken w.r.t. the distribution of $X$. If there is no subscript, we mean the probability and expectation are calculated w.r.t. all randomness.

\section{Methodology}\label{sec: method}

\subsection{The first algorithm: NPMC-CX}\label{subsec: npmc-cx}
Before formally introducing our first algorithm, we would like to derive it through heuristic calculations. For problem \eqref{eq: NPMC}, consider its Lagrangian function as
\begin{align}
	&L(\blambda, \phi) = \sum_{k \notin \mathcal{A}}w_k\tp_{X|Y=k}(\phi(X)\neq k) + \sum_{k \in \mathcal{A}}(w_k + \lambda_k)\tp_{X|Y=k}(\phi(X)\neq k) -\sum_{k \in \mathcal{A}}\lambda_k\alpha_k \\
	&= -\sum_{k \notin \mathcal{A}}w_k\tp_{X|Y=k}(\phi(X)= k) - \sum_{k \in \mathcal{A}}(w_k + \lambda_k)\tp_{X|Y=k}(\phi(X)= k) +\sum_{k=1}^K w_k +\sum_{k \in \mathcal{A}}\lambda_k(1-\alpha_k), \label{eq: f eq 1}
\end{align}
where $\blambda = \{\lambda_k\}_{k \in \mathcal{A}}$. Then, the dual problem of \eqref{eq: NPMC} can be written as
\begin{equation}\label{eq: dual of NPMC}
	\max_{\blambda \in \mathbb{R}_{+}^{|\mathcal{A}|}} \min_{\phi} L(\blambda, \phi).
\end{equation}
We can see that \eqref{eq: dual of NPMC} looks for a lower bound of the objective function in \eqref{eq: NPMC}, i.e., $\sup_{\blambda \in \mathbb{R}_{+}^{|\mathcal{A}|}} \allowbreak \min_{\phi} L(\blambda, \phi) \leq  \inf_{\phi \in \mathfrak{C}} \sum_{k=1}^K w_k \tp_{X|Y=k}(\phi(X)\neq k)$, where $\mathfrak{C}$ includes all feasible classifiers for problem \eqref{eq: NPMC}. We often call this fact as \textit{weak duality}. In many cases, the exact equality holds, which is called \textit{strong duality}. Under strong duality, \eqref{eq: NPMC} and \eqref{eq: dual of NPMC} can be seen as two different approaches to address the same problem. If one has an optimal solution, the other one will have an optimal solution as well. If the original NPMC problem \eqref{eq: NPMC} is infeasible, then \eqref{eq: dual of NPMC} must be unbounded from above, and vice versa. Another key observation is that, for a given $\blambda \in \mathbb{R}_{+}^{|\mathcal{A}|}$, looking for $\phi$ that minimizes $L(\blambda, \phi)$ in \eqref{eq: f eq 1} effectively translates into a CS problem \eqref{eq: general cs problem} with costs
\begin{equation}
	c_k(\blambda, \bpi^*) = \begin{cases}
		w_k/\pi_k^*, \quad &k \notin \mathcal{A};\\
		(w_k+\lambda_k)/\pi_k^*, \quad &k \in \mathcal{A}.
	\end{cases}
\end{equation}
This observation motivates our first algorithm, where we endeavor to solve the more tractable CS problem \eqref{eq: dual of NPMC} to address the more challenging original problem \eqref{eq: NPMC}. 

To derive our first algorithm, let's rewrite \eqref{eq: f eq 1} as
\begin{equation}\label{eq: ex dual npmc switch}
	L(\blambda, \phi) = -\te_X\left[c_{\phi(X)}(\blambda, \bpi^*)\cdot\tp_{Y|X}(Y = \phi(X))\right] +\sum_{k=1}^K w_k +\sum_{k \in \mathcal{A}}\lambda_k(1-\alpha_k).
\end{equation}
Then by Lemma \ref{lemma: cs oracle classifier}, we can define 
\begin{align}
	\phi^*_{\blambda}: \bx &\mapsto \argmax_k \{c_k(\blambda, \bpi^*)\tp_{Y|X=\bx}(Y = k)\} \in \argmin_\phi L(\blambda, \phi), \label{eq: phi star lambda} \\
	G(\blambda) &= \min_{\phi}L(\blambda, \phi) = L(\blambda, \phi^*). \label{eq: G population}
\end{align}
Therefore, on the population level, we can find $\blambda$ which maximizes $G(\blambda)$, then plug $\blambda$ into \eqref{eq: phi star lambda} to obtain the final classifier. On the other hand, due to weak duality, since the objective function in \eqref{eq: NPMC} is no larger than $1$ when it's feasible, we must have $\sup_{\blambda \in \mathbb{R}_{+}^{|\mathcal{A}|}}G(\blambda) \leq 1$. Thus, if $\sup_{\blambda \in \mathbb{R}_{+}^{|\mathcal{A}|}}G(\blambda) > 1$, the original NP problem \eqref{eq: NPMC} must be infeasible.

In practice, there is no access to $L(\blambda, \phi)$ and $G(\blambda)$ since we do not know the true model. We estimate $L(\blambda, \phi)$ by training data as
\begin{equation}\label{eq: F lambda hat}
	\hL^{\textup{CX}}(\blambda, \phi) = -\frac{1}{n}\sum_{i=1}^n c_{\phi(\bx_i)}(\blambda, \hbpi)\widehat{\tp}_{Y|X=\bx_i}(Y = \phi(\bx_i)) +\sum_{k=1}^K w_k +\sum_{k \in \mathcal{A}}\lambda_k(1-\alpha_k),
\end{equation}
where 
\begin{equation}
	c_k(\blambda, \hbpi) = \begin{cases}
		w_k/\hpi_k, \quad &k \notin \mathcal{A};\\
		(w_k+\lambda_k)/\hpi_k, \quad &k \in \mathcal{A},
	\end{cases}
\end{equation}
$\hpi_k = n_k/n$ with $n_k = \#\{i: y_i =k\}$, $\hbpi = \{\hpi_k\}_{k=1}^K$, and $\widehat{\tp}_{Y|X}$ is the estimated conditional probability. $\widehat{\tp}_{Y|X}$ can be obtained from any function class by fitting the data, and we do not impose any conditions on it here. Here are two examples.
\begin{itemize}
	\item For a parametric example, we may use the data to fit a multinomial logistic regression model and obtain the estimates of $(K-1)$ contrast coefficients $\{\hat{\bbeta}^{(k)}\}_{k=1}^{K-1}$ with $\hat{\bbeta}^{(k)} \in \mathbb{R}^p$. Then $\widehat{\tp}_{Y|X=\bx}(Y=k) = \frac{\exp\{\bx^T\hat{\bbeta}^{(k)}\}}{\sum_{k=1}^K\exp\{\bx^T\hat{\bbeta}^{(k)}\}}$ where $\hat{\bbeta}^{(K)} = \bm{0}_p$. 
	\item For a non-parametric example, we may use the $k$-nearest neighbors ($k$NN) to obtain the estimate $\widehat{\tp}_{Y|X=\bx}$. Given such an $\bx$ and the number of the nearest neighbors $k_0$, we can use the proportion of training observations of class $k$ among $k_0$ nearest neighbors to $\bx$ as an estimate $\widehat{\tp}_{Y|X=\bx}(Y=k)$.
\end{itemize}

Similar to Lemma \ref{lemma: cs oracle classifier}, it is easy to show that one of the optimal classifiers that minimize $\hL^{\textup{CX}}(\blambda, \phi)$ for a given $\blambda$ is 
\begin{equation}\label{eq: cs classifier given lambda empirical}
	\hphi_{\blambda}: \bx \mapsto \argmax_k \{c_k(\blambda, \hbpi)\widehat{\tp}_{Y|X=\bx}(Y = k)\} \in \argmin_\phi \hL^{\textup{CX}}(\blambda, \phi).
\end{equation}
Denote 
\begin{equation}\label{eq: G emipical}
	\widehat{G}^{\textup{CX}}(\blambda) \coloneqq \widehat{G}^{\textup{CX}}(\blambda;\widehat{\tp}_{Y|X}, \hbpi) = \min_{\phi}\hL^{\textup{CX}}(\blambda, \phi) = \hL^{\textup{CX}}(\blambda, \hphi_{\blambda}),
\end{equation}
which is a well-defined function of $\blambda$ given $\widehat{\tp}_{Y|X}$ and $\hbpi$. Similar to \eqref{eq: dual of NPMC}, we solve 
\begin{equation}\label{eq: dual of NPMC empirical}
	\max_{\blambda \in \mathbb{R}_{+}^{|\mathcal{A}|}} \min_{\phi} \hL^{\textup{CX}}(\blambda, \phi) = \max_{\blambda \in \mathbb{R}_{+}^{|\mathcal{A}|}} \hL^{\textup{CX}}(\blambda, \hphi_{\blambda}) = \max_{\blambda \in \mathbb{R}_{+}^{|\mathcal{A}|}}\widehat{G}^{\textup{CX}}(\blambda)
\end{equation}
to find solution $\hblambda$, then plug it in \eqref{eq: cs classifier given lambda empirical} to obtain the final solution $\hphi_{\hblambda}$ to the original NPMC problem \eqref{eq: NPMC}. On the other hand, considering the estimation error, if $\sup_{\blambda \in \mathbb{R}_{+}^{|\mathcal{A}|}}\widehat{G}^{\textup{CX}}(\blambda) > 1+\delta$ with a small positive constant $\delta$, we declare that the NPMC problem \eqref{eq: NPMC} is infeasible.

\begin{algorithm}[!h]
\setstretch{1.2}
\caption{NPMC-CX}
\label{algo}
\KwIn{training data $\{(\bm{x}_i, y_i)\}_{i = 1}^n$, target upper bounds of errors $\balpha$, the weighting vector of objective function $\bw$, a function class $\mathcal{M}$ to estimate $\tp_{Y|X}$, a small constant $\delta > 0$}
\KwOut{the fitted classifier $\hphi$ or report the NP problem as infeasible}
$\widehat{\tp}_{Y|X}, \hbpi \leftarrow$ the estimates of $\tp_{Y|X}$ (chosen from $\mathcal{M}$) and $\bpi^*$ on training data $\{(\bm{x}_i, y_i)\}_{i = 1}^n$ \\
\uIf{$\sup_{\blambda \in \mathbb{R}_{+}^{|\mathcal{A}|}} \widehat{G}^{\textup{CX}}(\blambda; \widehat{\tp}_{Y|X}, \hbpi) \leq 1+\delta$}{
$\hblambda \in \argmax_{\blambda \in \mathbb{R}_{+}^{|\mathcal{A}|}} \widehat{G}^{\textup{CX}}(\blambda; \widehat{\tp}_{Y|X}, \hbpi) \hspace{7cm} \inlineeqnum\label{eq: algo 1 hat lambda opt}$   \\
Report the NP problem as feasible and output the solution $\hphi(\bx) = \argmax_k \{c_k(\hblambda, \hbpi)\widehat{\tp}_{Y|X=\bx}(Y = k)\}$\\
}
\Else{Report the NP problem as infeasible}
\end{algorithm}

Note that $\widehat{G}^{\textup{CX}}(\blambda)$ is a concave function (as we will show in Proposition \ref{prop: concave G CX}), which implies that the optimization problem \eqref{eq: algo 1 hat lambda opt} is convex. Therefore, we refer to the algorithm above as NPMC-CX, summarized in Algorithm \ref{algo}. It is worth noting that $\widehat{G}^{\textup{CX}}(\blambda)$ is also a piecewise linear function on $\mathbb{R}_{+}^{|\mathcal{A}|}$. In practice, despite the concavity of $\widehat{G}^{\textup{CX}}(\blambda)$, the common convex optimization methods are difficult to apply due to the difficulty in calculating the gradient of $\widehat{G}^{\textup{CX}}(\blambda)$ w.r.t. $\blambda$. Instead, we implement the optimization step via direct search methods like the Hooke-Jeeves method \citep{hooke1961direct} and Nelder-Mead method \citep{nelder1965simplex}. More implementation details will be described in Section \ref{sec: numerical} and Section \ref{sec: additional numerical results supp} of the supplementary materials.

\subsection{The second algorithm: NPMC-ER}\label{subsec: npmc-er}
In Section \ref{subsec: npmc-cx}, we introduced an estimator \eqref{eq: F lambda hat} for the Lagrangian function \eqref{eq: f eq 1}. In the literature on NP classification, a more popular estimator is constructed using empirical error rates on a separate data set \citep{landgrebe2005neyman, tong2013plug}. In this section, we develop a new algorithm, NPMC-ER, based on a different estimator for \eqref{eq: f eq 1} using empirical error rates. We will compare NPMC-CX and NPMC-ER both theoretically (Section \ref{sec: theory}) and empirically (Section \ref{sec: numerical}). Some take-away messages will be summarized in Section \ref{sec: discussion}. 

For convenience, throughout this section, we assume the training sample size to be $2n$. Consider the following procedure. First, we divide the training data randomly into two parts of size $n$: $\mathcal{D}_1 = \{(\bx_i, y_i)\}_{i=1}^{n}$ and $\mathcal{D}_2 = \{(\bx_i, y_i)\}_{i=n+1}^{2n}$. $\mathcal{D}_1$ will be used to calculate the value of $\hL^{\textup{ER}}(\blambda, \phi)$ (to be defined),  and $\mathcal{D}_2$ will be used to estimate $\widehat{\tp}_{Y|X}$ and $\hbpi$. We estimate \eqref{eq: f eq 1} on $\mathcal{D}_1 = \{(\bm{x}_i, y_i)\}_{i = 1}^{n} = \{\{(\bm{x}_i^{(k)}, y_i^{(k)})\}_{i = 1}^{n_k}\}_{k=1}^K$ by
\begin{align}
	\hL^{\textup{ER}}(\blambda, \phi) &= -\sum_{k \notin \mathcal{A}}w_k\cdot \frac{1}{n_k}\sum_{i=1}^{n_k}\mathds{1}(\phi(\bx_i^{(k)})=k) - \sum_{k \in \mathcal{A}}(w_k+\lambda_k)\cdot \frac{1}{n_k}\sum_{i=1}^{n_k}\mathds{1}(\phi(\bx_i^{(k)})=k)\\
	&\quad + \sum_{k=1}^K w_k +  \sum_{k \in \mathcal{A}}\lambda_k(1-\alpha_k), \label{eq: f eq 2}
\end{align}
where $\{(\bm{x}_i^{(k)}, y_i^{(k)})\}_{i = 1}^{n_k}$ are the observations from class $k$ in $\mathcal{D}_1$. Then, similar to \eqref{eq: dual of NPMC empirical}, we solve 
\begin{equation}\label{eq: dual of NPMC empirical error rate}
	 \hblambda \in \argmax_{\blambda \in \mathbb{R}_{+}^{|\mathcal{A}|}} \hL^{\textup{ER}}(\blambda, \hphi_{\blambda}), \footnote{This $\hblambda$ is different from the $\hblambda$ estimated in NPMC-CX. We ignore the superscript for simplicity.}
\end{equation}
where $\hphi_{\blambda}$ is defined as in \eqref{eq: cs classifier given lambda empirical} while $\widehat{\tp}_{Y|X}$ and $\hbpi$ are calculated by data in $\mathcal{D}_2$.  Define 
\begin{equation}\label{eq: G empirical npmc-er}
	\widehat{G}^{\textup{ER}}(\blambda) \coloneqq \widehat{G}^{\textup{ER}}(\blambda;\widehat{\tp}_{Y|X}, \hbpi)= \hL^{\textup{ER}}(\blambda, \hphi_{\blambda}).
\end{equation}
Note that in NPMC-CX, given any $\blambda$, $\hphi_{\blambda}$ is a minimizer of $\hL^{\textup{CX}}(\blambda, \phi)$ w.r.t. any classifier $\phi$. In contrast, for NPMC-ER, given $\blambda$, we still define $\hphi_{\blambda}$ as in \eqref{eq: cs classifier given lambda empirical}, which is not necessarily a minimizer of $\hL^{\textup{ER}}(\blambda, \phi)$, and $\widehat{G}^{\textup{ER}}(\blambda)$ is not necessarily equal to $\sup_{\blambda \in \mathbb{R}_{+}^{|\mathcal{A}|}} \min_{\phi} \hL^{\textup{ER}}(\blambda, \phi)$. The remaining steps are the same as NPMC-CX.

The reason we do not define $\hphi_{\blambda}$ as $\argmin_{\phi} \hL^{\textup{ER}}(\blambda, \phi)$ is that there might be many (even infinitely many) minimizers, leading to instability in the estimated model. This issue often arises when fitting models via minimizing the training error. For instance, rescaling all coefficient components in logistic regression does not change the classification results and error rates.

\begin{algorithm}[!h]
\setstretch{1.2}
\caption{NPMC-ER}
\label{algo: erm}
\KwIn{training data $\{(\bm{x}_i, y_i)\}_{i = 1}^{2n}$, target upper bound of errors $\balpha$, the weighting vector of objective function $\bw$, a search range $R > 0$, a function class $\mathcal{M}$ to estimate $\tp_{Y|X}$, a small constant $\delta > 0$}
\KwOut{the fitted classifier $\hphi$ or report the NP problem as infeasible}
Randomly divide the whole training data (and reindex them) into $\mathcal{D}_1 \bigcup \mathcal{D}_2 = \{(\bm{x}_i, y_i)\}_{i = 1}^{n} \bigcup \{(\bm{x}_i, y_i)\}_{i = n+1}^{2n}$\\
$\widehat{\tp}_{Y|X}, \hbpi \leftarrow$ the estimates of $\tp_{Y|X}$ (chosen from $\mathcal{M}$) and $\bpi^*$ on $\mathcal{D}_{2} = \{(\bm{x}_i, y_i)\}_{i = n+1}^{2n}$\\
$\hblambda \leftarrow \argmax_{\blambda \in \mathbb{R}_{+}^{|\mathcal{A}|}, \infnorm{\blambda} \leq R} \widehat{G}^{\textup{ER}}(\blambda; \widehat{\tp}_{Y|X}, \hbpi)$, where $\widehat{G}^{\textup{ER}}$ is estimated on $\mathcal{D}_1 = \{(\bm{x}_i, y_i)\}_{i = 1}^{n} \hspace{9cm}\inlineeqnum\label{eq: algo 2 hat lambda opt}$ \\
\uIf{$\widehat{G}^{\textup{ER}}(\hblambda) \leq 1+\delta$}{
Report the NP problem as feasible and output the solution $\hphi(\bx) = \argmax_k \{c_k(\hblambda, \hbpi)\widehat{\tp}_{Y|X=\bx}(Y = k)\}$\\
}
\Else{Report the NP problem as infeasible}
\end{algorithm}

We name the second algorithm NPMC-ER because it uses the empirical error to estimate the true error rate, and we summarize it as Algorithm \ref{algo: erm}. Similar to $\widehat{G}^{\textup{CX}}(\blambda)$ defined in \eqref{eq: G emipical}, $\widehat{G}^{\textup{ER}}(\blambda)$ in \eqref{eq: G empirical npmc-er} is also a piecewise linear function of $\blambda$. However, it is not necessarily concave. Similar to NPMC-CX, we use the direct search method to conduct the optimization step \eqref{eq: algo 2 hat lambda opt} in practice. Note that since $\widehat{G}^{\textup{ER}}(\blambda)$ is not necessarily concave, for technical reasons, we need to restrict the search range of the best $\blambda$ to a bounded region. Hence, compared to NPMC-CX (Algorithm \ref{algo}), there is an additional argument representing the search range $R$ in NPMC-ER (Algorithm \ref{algo: erm}). The condition on $R$ in the theoretical analysis will be described in the next section. The empirical results are not very sensitive to the choice of $R$, and we pick $R=1000$ in all numerical studies.

\section{Theory}\label{sec: theory}
In this section, we delve into the theoretical properties of the two algorithms introduced in Section \ref{sec: method}. We begin with Section \ref{subsec: strong duality}, where we establish sufficient and necessary conditions for strong duality, shedding light on the circumstances under which it holds. Sections \ref{subsec: theory npmc-cx} and \ref{subsec: theory npmc-er} are dedicated to presenting the theoretical foundations of NPMC-CX and NPMC-ER, respectively. In Section \ref{subsec: comparison theory}, we undertake a theoretical comparison of the two algorithms, unearthing additional insights that encompass discussions on the assumptions and other essential properties of NP algorithms. The additional details omitted in this section can be found in Section \ref{sec: additional NPMC supp} of the supplementary materials.

\subsection{Checking strong duality and feasibility}\label{subsec: strong duality}
As described in the heuristic arguments in Section \ref{subsec: npmc-cx}, strong duality between the original NPMC problem \eqref{eq: NPMC} and the dual problem \eqref{eq: dual of NPMC} is vital for our algorithms to work well. Therefore, we formalize the requirement of strong duality through the following assumption.
\begin{assumption}[Strong duality for the NPMC problem]\label{asmp: strong duality}
	It holds that
	\vspace{-0.4cm}
	\begin{equation}
		\inf_{\phi \in \mathfrak{C}}J(\phi) = \sup_{\blambda \in \mathbb{R}_{+}^{|\mathcal{A}|}}G(\blambda),
	\end{equation}
	where $\mathfrak{C}$ includes all feasible classifiers for the NPMC problem \eqref{eq: NPMC}. If $\mathfrak{C} \neq \emptyset$, the infimum over $\phi \in \mathfrak{C}$ is achievable, and the supremum over $\blambda \in \mathbb{R}_{+}^{|\mathcal{A}|}$ can be attained  at a finite $\blambda$.
\end{assumption}

There are various sufficient conditions for strong duality in literature, e.g., Slater's condition \citep{luenberger1997optimization, boyd2004convex}. However, most of them are applicable only to convex problems, while the original NPMC problem \eqref{eq: NPMC} is not necessarily convex. The following theorem elucidates a tight relationship between the feasibility of the induced classifier from the dual CS problem \eqref{eq: dual of NPMC} and the strong duality in the NPMC problem \eqref{eq: NPMC}.

\begin{theorem}[Sufficient and necessary conditions for NPMC strong duality]\label{thm: strong duality}
Suppose $\{X|Y=k\}_{k=1}^K$ are continuous random variables (i.e. have Lebesgue density). 
\begin{enumerate}[(i)]
	\item When the NPMC problem \eqref{eq: NPMC} is feasible, the strong duality holds if and only if there exists $\blambda^{(0)} = \{\lambda_k^{(0)}\}_{k \in \mathcal{A}}$ such that $\phi_{\blambda^{(0)}}^*$ is feasible for the NPMC problem \eqref{eq: NPMC}, i.e., $\tp_{X|Y=k}(\phi_{\blambda^{(0)}}^*(X) \neq k) \leq \alpha_k$ for all $k \in \mathcal{A}$.
	\item Suppose $\tp_{Y|X=\bx}(Y=k) \geq a > 0$ for a.s. $\bx$ (w.r.t. the distribution of $X$) and all $k \in \mathcal{A}$. When the NPMC problem \eqref{eq: NPMC} is infeasible, the strong duality holds (i.e., $\sup_{\blambda \in \mathbb{R}_{+}^{|\mathcal{A}|}}G(\blambda)$ is unbounded from above) if and only if for an arbitrary $\blambda \in \mathbb{R}^{|\mathcal{A}|}_+$, $\phi_{\blambda}^*$ is infeasible for NPMC problem \eqref{eq: NPMC}, i.e., $\exists$ at least one $k \in \mathcal{A}$ such that $\tp_{X|Y=k}(\phi_{\blambda}^*(X) \neq k) > \alpha_k$.
\end{enumerate}
\end{theorem}

Building upon Theorem \ref{thm: strong duality}, we derive the following corollary, which proves to be very useful in practical assessments of feasibility and strong duality.
 
\begin{corollary}\label{cor: df check}
	Suppose $\{X|Y=k\}_{k=1}^K$ are continuous random variables (i.e. have Lebesgue density).  The following equivalences hold:
	\begin{enumerate}[(i)]
		\item The NPMC problem is feasible, and strong duality holds $\Leftrightarrow$ $\exists$ a finite $\blambda^* \in \argmax_{\blambda \in \mathbb{R}_{+}^{|\mathcal{A}|}}G(\blambda)$ and $\phi^*_{\blambda^*}$ is feasible;
		\item The NPMC problem is infeasible, and strong duality holds $\Leftrightarrow$ $\exists$ an infinite $\blambda^*$ and $G(\blambda^*) = +\infty$;
		\item \begin{itemize}
				\item The NPMC problem is feasible, and strong duality fails $\Rightarrow$ For any $\blambda^* \in \argmax_{\blambda \in \mathbb{R}_{+}^{|\mathcal{A}|}}G(\blambda)$, $\lambda^*$ is infinite \footnote{When we say infinite $\blambda^*$, we refer to a sequence $\{(\blambda^*)^{(m)}\}_{m=1}^{\infty}$ s.t. $\infnorm{(\blambda^*)^{(m)}}\rightarrow +\infty$, $\lim_{m \rightarrow \infty} G((\blambda^*)^{(m)}) = \sup_{\blambda \in \mathbb{R}_+^{|\mathcal{A}|}}G(\blambda)$ exists and is denoted as $G(\blambda^*)$.} or $\blambda^*$ is finite but $\phi^*_{\blambda^*}$ is infeasible, and 	$G(\blambda^*) \leq 1$;
				\item For any $\blambda^* \in \argmax_{\blambda \in \mathbb{R}_{+}^{|\mathcal{A}|}}G(\blambda)$, $\lambda^*$ is infinite or $\blambda^*$ is finite but $\phi^*_{\blambda^*}$ is infeasible, and $G(\blambda^*) \leq 1 \Rightarrow$ strong duality fails, and the NPMC problem can be either feasible or infeasible;
			 \end{itemize}
		\item \begin{itemize}
 				\item The NPMC problem is infeasible, and strong duality fails $\Rightarrow$ For any $\blambda^* \in \argmax_{\blambda \in \mathbb{R}_{+}^{|\mathcal{A}|}}G(\blambda)$, $\blambda^*$ is infinite or $\blambda^*$ is finite but $\phi^*_{\blambda^*}$ is infeasible, and $G(\blambda^*) < +\infty$.
 				\item For any $\blambda^* \in \argmax_{\blambda \in \mathbb{R}_{+}^{|\mathcal{A}|}}G(\blambda)$, $\blambda^*$ is infinite or $\blambda^*$ is finite but $\phi^*_{\blambda^*}$ is infeasible, and $1 < G(\blambda^*) < +\infty \Rightarrow$ strong duality fails, the NPMC problem is infeasible.
 			  \end{itemize}
	\end{enumerate}
\end{corollary}

Corollary \ref{cor: df check} establishes a connection between NPMC strong duality and feasibility with the optimal $\blambda^*$ and the value of $G(\blambda^*)$. In practice, $\blambda^*$ and $G(\blambda^*)$ can be estimated by $\hblambda$ and $\widehat{G}^{\textup{CX}}(\hblambda)$ from NPMC-CX or $\hblambda$ and $\widehat{G}^{\textup{ER}}(\hblambda)$ from NPMC-ER. The equivalences in Corollary \ref{cor: df check} can then be used to assess whether feasibility and strong duality hold. Due to space constraints, further details are provided in Section \ref{subsec: dk check NPMC supp} in the supplements, while related empirical results will be discussed in Section \ref{sec: numerical}.

\subsection{Analysis on NPMC-CX}\label{subsec: theory npmc-cx}
It is well-known that regardless of the primal problem, the Lagrangian dual function is always concave \citep{luenberger1997optimization, boyd2004convex}, implying that $G(\blambda)$ in \eqref{eq: G population} is concave w.r.t. $\blambda$. For NPMC-CX, the empirical version $\widehat{G}(\blambda)$ in \eqref{eq: G emipical} is a concave function as well, making \eqref{eq: algo 1 hat lambda opt} a convex optimization problem. 

\begin{proposition}\label{prop: concave G CX}
	$G(\blambda)$ and $\widehat{G}^{\textup{CX}}(\blambda)$ are concave and continuous on $\mathbb{R}^{|\mathcal{A}|}_+$.
\end{proposition}

Suppose we estimate $\tp_{Y|X=\bx}(Y = k)$ with a function class $\mathcal{M}$ that can be indexed by an index $\bbeta \in \mathcal{B}$, where $\mathcal{B}$ is some metric space. Suppose the data dimension $p$ is fixed.

To prove the NP oracle properties of NPMC-CX, we impose the following assumptions.

\begin{assumption}[Model consistency]\label{asmp: posterior distance}
	$\max_k \te |\widehat{\tp}_{Y|X}(Y = k)-\tp_{Y|X}(Y = k)| \rightarrow 0$ as $n \rightarrow \infty$.
\end{assumption}

\begin{assumption}[Strict concavity]\label{asmp: second order info}
	 $G(\blambda)$ is continuously twice-differentiable at $\blambda^*$ and $\nabla^2 G(\blambda^*) \prec 0$, where $\blambda^* = \argmax_{\blambda \in \mathbb{R}_+^{|\mathcal{A}|}} G(\blambda)$ is unique.
\end{assumption}

\begin{assumption}[Rademacher classes]\label{asmp: rademacher}
	The function class for estimating conditional probability $\mathcal{M}=\{\{\widehat{\tp}_{Y|X=\bx}(Y=k;\bbeta)\}_{k=1}^K: \bbeta \in \mathcal{B}\}$ has a vanishing Rademacher complexity 
	\begin{equation}
		C_{\textup{Rad}}(n) \coloneqq \max_{k=1:K}\te \sup_{\bbeta \in \mathcal{B}}\bigg|\frac{1}{n}\sum_{i=1}^n \epsilon_i\widehat{\tp}_{Y|X=\bx_i}(Y=k; \bbeta)\bigg| \rightarrow 0,
	\end{equation}
	as $n \rightarrow \infty$, where $\bm{\epsilon} = (\epsilon_1, \ldots, \epsilon_n)^T$ is a vector of independent Rademacher variables.
\end{assumption}

\begin{assumption}[Margin condition]\label{asmp: marginal finite sample}
	Denote the function characterizing the decision boundary of class $k$ as $\varphi_k(\bx) = c_k(\blambda^*, \bpi^*)\tp_{Y|X=\bx}(Y=k) -\max_{j \neq k}\{c_j(\blambda^*, \bpi^*)\tp_{Y|X=\bx}(Y=j)\}$, where $\blambda^* = \argmax_{\blambda \in \mathbb{R}_+^{|\mathcal{A}|}} G(\blambda)$ is unique. It holds that
	\begin{equation}
		\max_{k=1:K}\tp_{X|Y=k}(\norm{\varphi_k(X)} \leq \tau) \lesssim \tau^{\bar{\gamma}},
	\end{equation}
	with some $\bar{\gamma} > 0$ and any non-negative $\tau$ smaller than some constant $C \in (0, 1)$.
\end{assumption}

\begin{remark}
	Assumption \ref{asmp: posterior distance} ensures that the conditional probability can be accurately estimated. Assumption \ref{asmp: second order info} is motivated by the second-order information condition used in proving MLE consistency \citep{wald1949note, van2000asymptotic}. 
	
	Algorithm \ref{asmp: rademacher} restricts the model complexity \footnote{More precisely, such a restriction also depends on $\tp_X$ because $\te$ is w.r.t. all the random ness.}. Many parametric model classes fulfill this condition, such as the multinomial logistic model with bounded coefficients when $\tp_X$ has second-order moments. Additionally, certain non-parametric classes also satisfy this requirement, such as Lipschitz function classes with $\widehat{\tp}_{Y|X=\bx}(Y=k;\bbeta)$ Lipschitz in $\bx$ when $\tp_X$ is supported on a bounded set of $\mathbb{R}^p$. Note that the function class $\mathcal{M}$ does not necessarily correspond to the underlying true model, and we do not require the true model to belong to a Rademacher class. 

	Assumption \ref{asmp: marginal finite sample} is commonly referred to as ``margin condition" in literature \citep{audibert2007fast, tong2013plug, zhao2016neyman}, and it requires most data points to be away from the optimal decision boundary. In many cases, this assumption can lead to convergence rates faster than $\mathcal{O}_{\tp}(n^{-1/2})$. Previous binary NP classification papers such as \cite{tong2013plug}, \cite{zhao2016neyman} and \cite{tong2020neyman} do not require it when arbitrary convergence rates are acceptable. Besides, it is often employed with an opposite condition called ``detection condition" \citep{tong2013plug, zhao2016neyman, tong2020neyman} to aid in accurately estimating the optimal classification threshold. Here, we do not need such a detection condition, but Assumption \ref{asmp: marginal finite sample} is crucial and required to hold. 
	
	More discussions can be found in Section \ref{subsec: discussions on assumptions} of the supplementary materials. 
\end{remark}

Next, we establish that NPMC-CX satisfies the multi-class NP oracle properties under the conditions above.

\begin{theorem}[Multi-class NP oracle properties of NPMC-CX]\label{thm: error rate bound}
NPMC-CX satisfies multi-class NP oracle properties in the following senses.
	\begin{enumerate}[(i)]
		\item When the NPMC problem \eqref{eq: NPMC} is feasible, if Assumptions \ref{asmp: strong duality}-\ref{asmp: marginal finite sample} hold, and $\delta \gtrsim \big[R_{\textup{Rad}}(n) + \max_k \te\norm{\widehat{\tp}_{Y|X}(Y=k)-\tp_{Y|X}(Y=k)}\big]^{\bar{\gamma}/2}$, then there exist a solution $\phi^*$ and a constant $C > 0$ such that
			\begin{equation}
				\max_k\tp(\norm{R_k(\hphi) - R_k(\phi^*)} > \tau) \lesssim \exp\{-Cn\tau^{4/\bar{\gamma}}\}  + \tau^{-\frac{2\vee (1+\bar{\gamma})}{\bar{\gamma}}}\max_k \te\norma{\widehat{\tp}_{Y|X}(Y=k)-\tp_{Y|X}(Y=k)}, 
			\end{equation}
		when $1 \geq \tau \gtrsim \big[C_{\textup{Rad}}(n) + \max_{k}\te\norm{\widehat{\tp}_{Y|X}(Y=k)-\tp_{Y|X}(Y=k)}\big]^{\bar{\gamma}/2}$.
		\item When the NPMC problem \eqref{eq: NPMC} is infeasible, if Assumptions \ref{asmp: strong duality}, \ref{asmp: posterior distance} and \ref{asmp: rademacher} hold, and $\delta \gtrsim \big[R_{\textup{Rad}}(n) + \max_k \te\norm{\widehat{\tp}_{Y|X}(Y=k)-\tp_{Y|X}(Y=k)}\big]^{1/2}$, then there exists a constant $C > 0$ such that
			\begin{equation}
				\tp\left(\widehat{G}^{\textup{CX}}(\hblambda) \leq 1+\delta\right) \lesssim \exp\{-Cn\} +\max_k \te\norma{\widehat{\tp}_{Y|X}(Y=k)-\tp_{Y|X}(Y=k)},
			\end{equation}
			where $\delta$ is an input parameter in Algorithm \ref{algo}.
	\end{enumerate}
\end{theorem}

\begin{remark}\label{rmk: oracle npmc-cx}
	Observe that $J(\hphi) - J(\phi^*)$ is a linear combination of $\{R_k(\hphi) - R_k(\phi^*)\}_{k=1}^K$. Hence, when the NPMC problem \eqref{eq: NPMC} is feasible,
	\begin{align}
		R_k(\hphi) -  \alpha_k &\leq R_k(\hphi) -  R_k(\phi^*) \leq \mathcal{O}_{\tp}(\epsilon(n)), \quad \forall k \in \mathcal{A}, \\
		J(\hphi) - J(\phi^*) &\leq \mathcal{O}_{\tp}(\epsilon(n)),
	\end{align}
	where $\epsilon(n) = n^{-\bar{\gamma}/4} + \big(\max_k \te\norm{\widehat{\tp}_{Y|X}(Y=k)-\tp_{Y|X}(Y=k)}\big)^{\bar{\gamma}/(2\vee (1+\bar{\gamma}))} + (C_{\textup{Rad}}(n))^{\bar{\gamma}/2} \rightarrow 0$.
	Theorem \ref{thm: error rate bound} verifies multi-class NP oracle properties as defined in Section \ref{subsec: np oracle properties}. 
\end{remark}

\subsection{Analysis on NPMC-ER}\label{subsec: theory npmc-er}
One advantage of NPMC-ER over NPMC-CX is that it does not require $\widehat{\tp}_{Y|X=\bx}(Y=k)$ to belong to a Rademacher class. We will explain the intuition in the next subsection.

Unlike NPMC-CX, for NPMC-ER, the empirical dual function $\widehat{G}(\blambda)$ in \eqref{eq: G empirical npmc-er} is not necessarily concave. This discrepancy arises from the ``mismatch" of $\hL^{\textup{ER}}(\blambda, \phi)$ and $\hphi_{\blambda}$. As discussed in Section \ref{subsec: npmc-er}, given $\blambda$, $\hphi_{\blambda}$ is not necessarily a minimizer of $\hL^{\textup{ER}}(\blambda, \phi)$, leading to a dual function not of the ``max-min" type and hence not necessarily concave. Nonetheless, the multi-class NP oracle properties still hold under similar conditions.

\begin{theorem}[Multi-class NP oracle properties of NPMC-ER]\label{thm: error bound alg 2}
	NPMC-ER satisfies multi-class NP oracle properties in the following senses.
	\begin{enumerate}[(i)]
		\item When the NPMC problem \eqref{eq: NPMC} is feasible, if Assumptions \ref{asmp: strong duality}, \ref{asmp: posterior distance}, \ref{asmp: second order info} and \ref{asmp: marginal finite sample} hold, $\delta \gtrsim n^{-\bar{\gamma}/4}$, and $R \geq \infnorm{\blambda^*}$ with $\blambda^* = \argmax_{\blambda \in \mathbb{R}^{|\mathcal{A}|}_+} G(\blambda)$, then there exist a solution $\phi^*$ and a constant $C > 0$ such that
			\begin{equation}
				\max_k\tp(\norm{R_k(\hphi) - R_k(\phi^*)} > \tau) \lesssim   \exp\{-Cn\tau^{4/\bar{\gamma}}\} + \tau^{-\frac{2\vee (1+\bar{\gamma})}{\bar{\gamma}}}\max_k \te\norma{\widehat{\tp}_{Y|X}(Y=k)-\tp_{Y|X}(Y=k)},
			\end{equation}
		when $1 \geq \tau \gtrsim n^{-\bar{\gamma}/4}$.
		\item When the NPMC problem \eqref{eq: NPMC} is infeasible, if Assumptions \ref{asmp: strong duality} and \ref{asmp: posterior distance} hold, $\delta \gtrsim n^{-1/4}$, and $R$ satisfies $\sup_{\infnorm{\blambda} \leq R}G(\blambda) > 1 + \delta$, then there exists a constant $C > 0$ such that
			\begin{equation}
				\tp\left(\widehat{G}^{\textup{ER}}(\hblambda) \leq 1+\delta\right) \lesssim \exp\{-Cn\} +\max_k \te\norma{\widehat{\tp}_{Y|X}(Y=k)-\tp_{Y|X}(Y=k)},
			\end{equation}
			where $\delta$ is an input parameter in Algorithm \ref{algo}.
	\end{enumerate}
\end{theorem}

Analyzing similarly in Remark \ref{rmk: oracle npmc-cx}, we conclude that Theorem \ref{thm: error bound alg 2} confirms the multi-class NP oracle properties of NPMC-ER. As discussed in Section \ref{subsec: npmc-er}, because $\widehat{G}^{\textup{ER}}(\blambda)$ is not necessarily concave, for technical reasons, we can only search for optimal $\blambda$ within a bounded region $\infnorm{\blambda} \leq R$ where $R > 0$ is a constant.  On the other hand, to ensure that this search region covers the true optimal $\blambda^*$ (when the NPMC problem is feasible) or is large enough to find a large $G(\blambda)$ value (when the NPMC problem is infeasible), we need to ensure that $R$ is not very small, leading to the conditions $R \geq \infnorm{\blambda^*}$ and $\sup_{\infnorm{\blambda} \leq R}G(\blambda) > 1 + \delta$ in (\rom{1}) and (\rom{2}), respectively. The empirical results are not very sensitive to the choice of $R$, and we set $R=1000$ in all numerical studies.

\subsection{Comparison of NPMC-CX and NPMC-ER from theoretical perspective}\label{subsec: comparison theory}
We now summarize the difference between the two algorithms from theoretical perspectives.
\begin{itemize}
	\item Both NPMC-CX and NPMC-ER exhibit NP oracle properties under certain conditions. 
	\item NPMC-CX assumes the function class used to estimate the posterior $\tp_{Y|X=\bx}(Y = k)$ has a vanishing Rademacher complexity, while NPMC-ER does not impose such a restriction. This distinction arises because NPMC-CX utilizes all training data simultaneously, necessitating control over model complexity for certain uniform convergence results. In contrast, NPMC-ER leverages sample splitting, creating independence that only requires pointwise convergence instead of uniform convergence, regardless of the model class considered. Further details are available in the corresponding proofs provided in supplementary materials.
\end{itemize}

\section{Numerical Experiments}\label{sec: numerical}
We demonstrate the effectiveness of NPMC-CX and NPMC-ER through a simulation example and a real data study on loan default prediction. All numerical experiments were conducted using R. Our proposed algorithms, NPMC-CX and NPMC-ER, have been implemented in the package \texttt{npcs} (\url{https://CRAN.R-project.org/package=npcs}). In the simulations, we vary the training sample size $n$ from 1000 to 9000 with an increment of 2000, while keeping the test sample size fixed at 20,000. Without specific notice, each setting in both simulations and real data studies is repeated 500 times. Due to space constraints, we provide additional numerical results and more implementation details, including the choice of tuning parameters in Section \ref{sec: additional numerical results supp} of the supplementary materials.


\subsection{Simulation}\label{subsec: simulations}


Consider a three-class independent Gaussian conditional distributions $X|Y=k \sim N(\bmu_k, \bm{I}_p)$, where $p = 5$, $\bmu_1 = (-1,2,1,1,1)^T$, $\bmu_2 = (0,1,0,1,0)^T$, $\bmu_3 = (1,1,-1,0,1)^T$ and $\bm{I}_p$ is the $p$-dimensional identity matrix. The marginal distribution of $Y$ is $\tp(Y=1) = \tp(Y=3) = 0.3$ and $\tp(Y=2) = 0.4$. 

We aim to solve the following NPMC problem:
\vspace{-0.5cm}
\begin{align}
	&\min_{\phi} \quad \tp_{X|Y=3}(\phi(X)\neq 3) \\
	&\text{s.t.} \quad \tp_{X|Y=1}(\phi(X)\neq 1) \leq 0.15,  \quad \tp_{X|Y=2}(\phi(X)\neq 2) \leq 0.3.
\end{align}
\vspace*{-1.4cm}

\begin{figure}[!h]
	\centering
	\includegraphics[width=\textwidth]{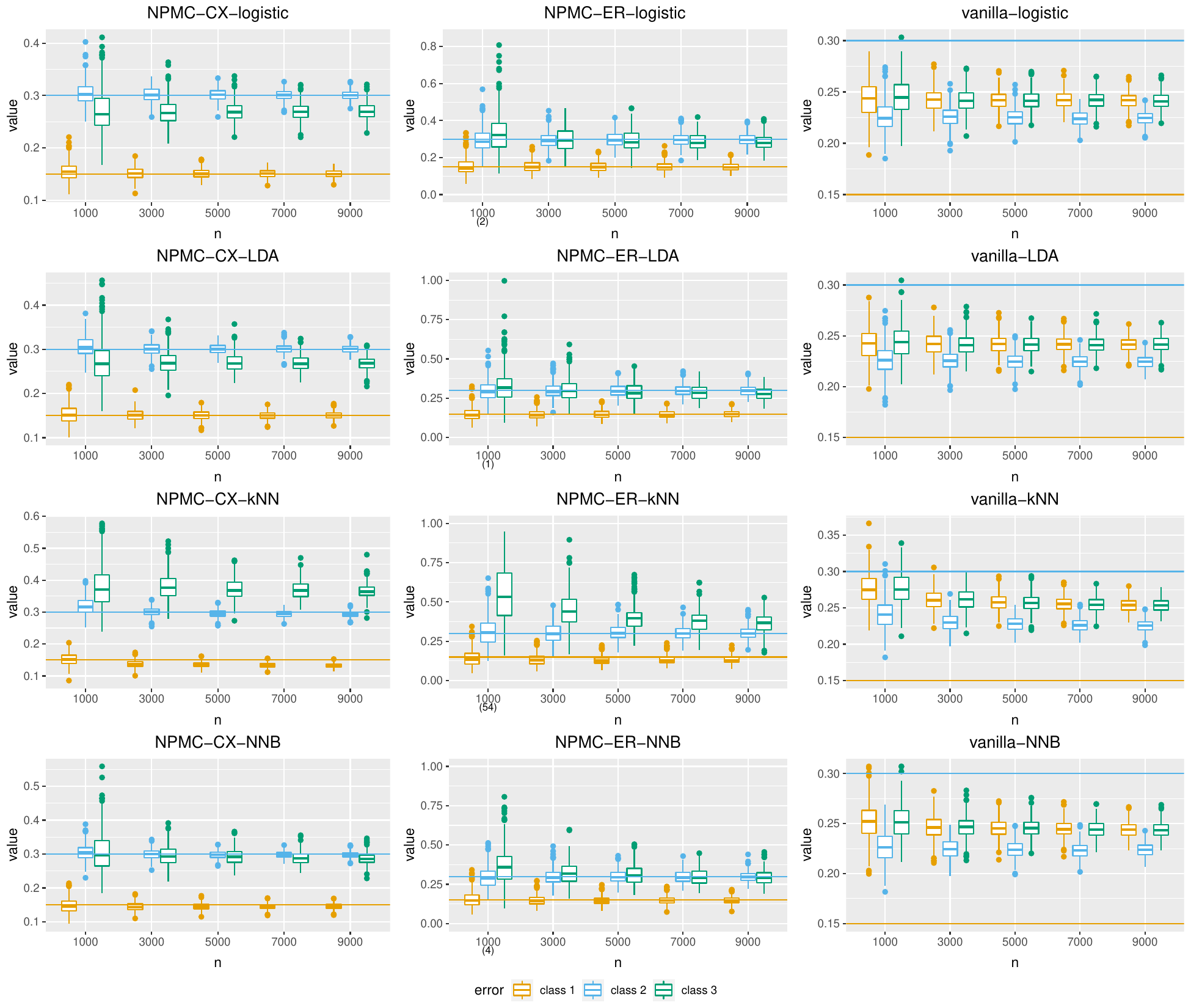}
	\caption{Per-class error rates under each classifier and training sample size setting in simulation. Horizontal lines in corresponding colors mark the target control levels. In some graphs, additional values are displayed in brackets beneath the training sample size, $n$. These values represent the number of instances where the algorithms reported infeasibility during evaluation. }
	\label{fig: model1}
\end{figure}

We run the proposed algorithms NPMC-CX and NPMC-ER with four function classes to estimate $\tp_{Y|X}$, including logistic regression, LDA, $k$NN, and non-parametric na\"ive Bayes model with Gaussian kernel. For comparison, we also fit four corresponding vanilla classifiers trained without error controls as benchmarks. Box plots show the per-class error rates under each method and training sample size setting in Figure \ref{fig: model1}.

One can see that vanilla classifiers fail to control the error of class 1 and ``over-control" the error of class 2. In contrast, NPMC-CX and NPMC-ER work very well by controlling the error rates around the target control level, which matches our theoretical results in Section \ref{sec: theory}. By comparing the error rates of class $2$ between NPMC methods and vanilla classifiers, we observe that to achieve a successful control over $\tp_{X|Y=1}(\phi(X)\neq 1)$ \footnote{To be more precise, the graphs only show the empirical error rates on the test data.} and $\tp_{X|Y=2}(\phi(X)\neq 2)$ around the corresponding levels, there is a cost in terms of the performance on class 3. When the training sample size $n$ increases, the variance of error rates for each method tends to decrease. For NPMC-CX-LDA and NPMC-CX-NNB, when $n$ is small, sometimes the algorithm outputs the infeasibility warning. For NPMC-CX-LDA, this behavior might be due to LDA's higher sample size requirements (because of the need to estimate the covariance matrix) compared to other methods like logistic regression. For NPMC-CX-NNB, this phenomenon could be caused by the improper choice of bandwidth. 

Another noteworthy observation is the higher variances of error rates on class 3 compared to the other two classes, particularly evident when $n$ is small. This phenomenon arises because the decision boundary of NP classifiers traverses the densely populated area for class 3 but not for classes 1 and 2 when stringent error controls are imposed on the latter. Consequently, even a small change in the decision boundary can lead to a relatively bigger change in the error of class 3 compared to classes 1 and 2. 

To validate the feasibility and strong duality checking algorithms induced by Corollary \ref{cor: df check} (see the algorithms in Section \ref{subsec: dk check NPMC supp} of the supplements and note that the feasibility prediction is the same as in NPMC-CX and NPMC-ER), we conducted experiments for them with NPMC-CX-logistic and NPMC-ER-$k$NN by fixing the random training data of size $n = 10^5$ and considering all choices of $(\alpha_1, \alpha_2)$ within range $[0.01,1]^2$ with a grid size $0.01$. Note that the feasibility and strong duality can be theoretically verified for any specific $(\alpha_1, \alpha_2)$ in this example. The following lemma, in conjunction with Lemma \ref{lem: feasibility toy example}, establishes the ground truth regarding strong duality and feasibility.

\begin{lemma}\label{lem: strong duality toy example}
	The strong duality in Assumption \ref{asmp: strong duality} holds for 3-class NPMC problem \eqref{eq: NPMC} with $X|Y = k \sim N(\bmu_k, \bm{I}_p)$ for $k = 1, 2, 3$, $\mathcal{A} = \{1,2\}$, and the target levels $\alpha_1$, $\alpha_2 \in [0,1]$, if and only if
	\vspace*{-0.3cm}
	\begin{equation}
		\twonorm{\bmu_1 - \bmu_2} \neq \Phi^{-1}(1-\alpha_1) + \Phi^{-1}(1-\alpha_2),
	\end{equation}
	\vspace*{-1.2cm}
	
	\noindent where $\Phi^{-1}$ is the inverse CDF function of $N(0, 1)$.
\end{lemma}

\begin{figure}[t]
	\centering
	\includegraphics[width=\textwidth]{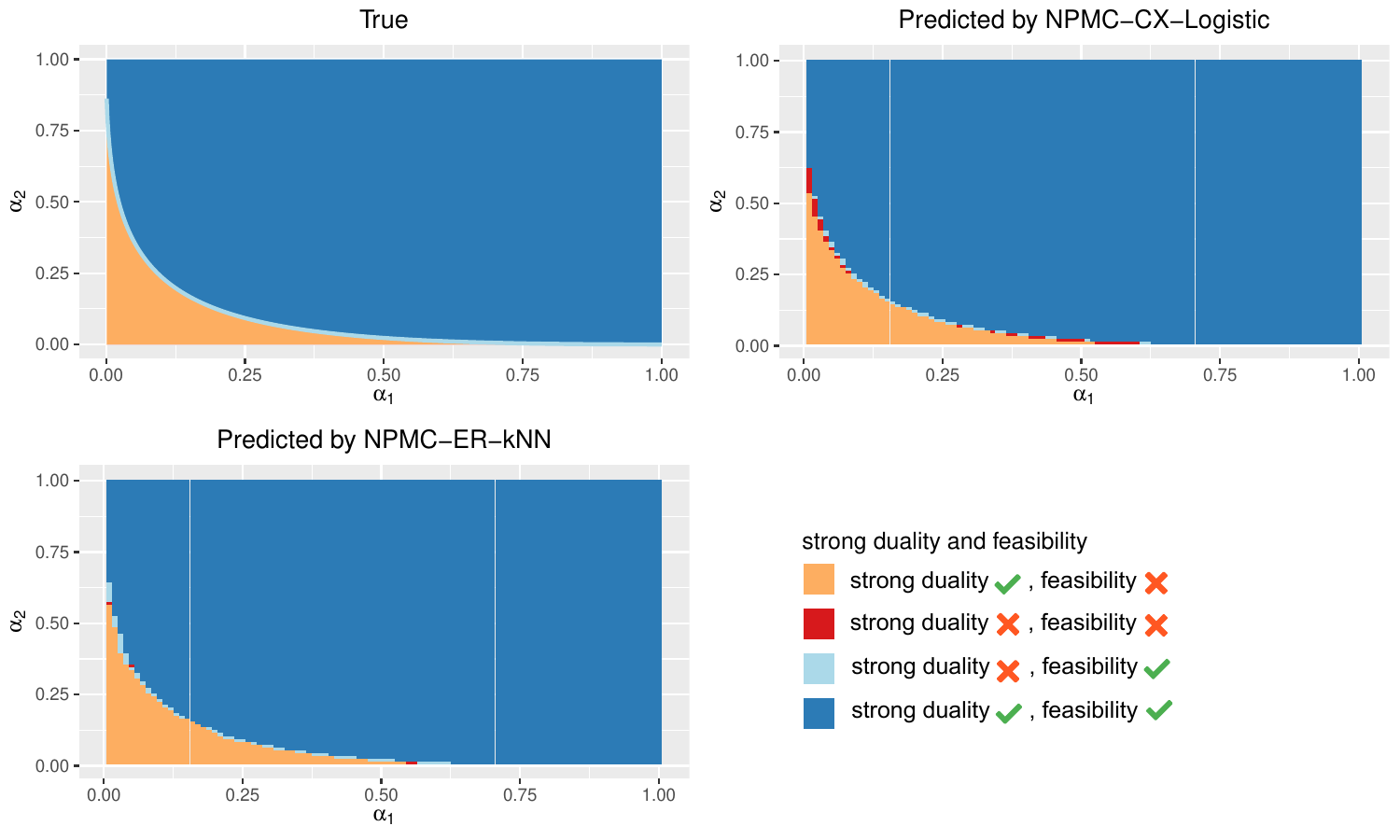}
	\caption{Strong duality and feasibility of simulation: ground truth and predicted results.}
	\label{fig: model1_true_combined}
\end{figure}

We then compared the true feasibility and strong duality with the predictions generated by our feasibility and strong duality checking algorithms in Figure \ref{fig: model1_true_combined}. It shows that our algorithms can accurately predict the feasibility and strong duality with sufficient data. Hence, practitioners can first utilize algorithms in Section \ref{subsec: dk check NPMC supp} to assess the feasibility and strong duality for various target error levels, thereby gaining insights into the problem difficulty, especially when they are unsure about the appropriate target levels for error controls. In other words, our feasibility and strong duality checking algorithms offer a prediction of the \textit{landscape} of an NPMC problem with various target levels.

\subsection{A real data study: loan default prediction}\label{subsec: real-data}
Identifying high-risk customers prone to late payments or default is paramount for banks and lending institutions in managing risk. Providing loans to high-risk customers often results in greater losses than denying loans to low-risk customers, underscoring the importance of effective risk assessment strategies. The Neyman-Pearson classification framework is particularly valuable in this context for its ability to address asymmetric errors.

LendingClub, a peer-to-peer lending company, caters to borrowers seeking personal loans ranging from \$1000 to \$40000. The LendingClub dataset (\url{https://www.kaggle.com/code/emmaruyiyang/lending-club-loan-default-prediction-eda/input}) encompasses loan data spanning from 2007 to 2015. It includes details such as loan amount, term length, current status, and borrower information like annual income and number of bankcard accounts. The objective is to predict the loan status based on these variables. The original dataset contains various labels for loan status, including ``fully paid", ``late payment" with varying durations, ``in grace period",``default", and  ``charge off". For simplicity, we categorize them into three groups: class 1 (bad status: default or charge off), class 2 (fair status: late payment but not default), and class 3 (excellent status: fully paid). Following some preprocessing steps (refer to Section \ref{subsubsec: lendingclub supp} for details), the dataset comprises 264274 observations with 25 features and 1 response variable. The sample sizes for the three classes are 45072 (17.1\%), 19265 (7.3\%), and 199937 (75.6\%), respectively. The significant class imbalance poses an additional challenge in addressing this problem. 

We would like to solve the following NPMC problem
\vspace{-0.5cm}
\begin{align}
	&\min_{\phi} \quad \tp_{X|Y=3}(\phi(X)\neq 3) \\
	&\text{s.t.} \quad \tp_{X|Y=1}(\phi(X)\neq 1) \leq \alpha_1,  \quad \tp_{X|Y=2}(\phi(X)\neq 2) \leq \alpha_2,
\end{align}
\vspace*{-1.3cm}

\noindent where $\alpha_1$ is typically chosen to be smaller than $\alpha_2$ because misclassifying observations of class 1 is more detrimental than misclassifying those of class 2.

\begin{figure}[!]
	\centering
	\includegraphics[width=\textwidth]{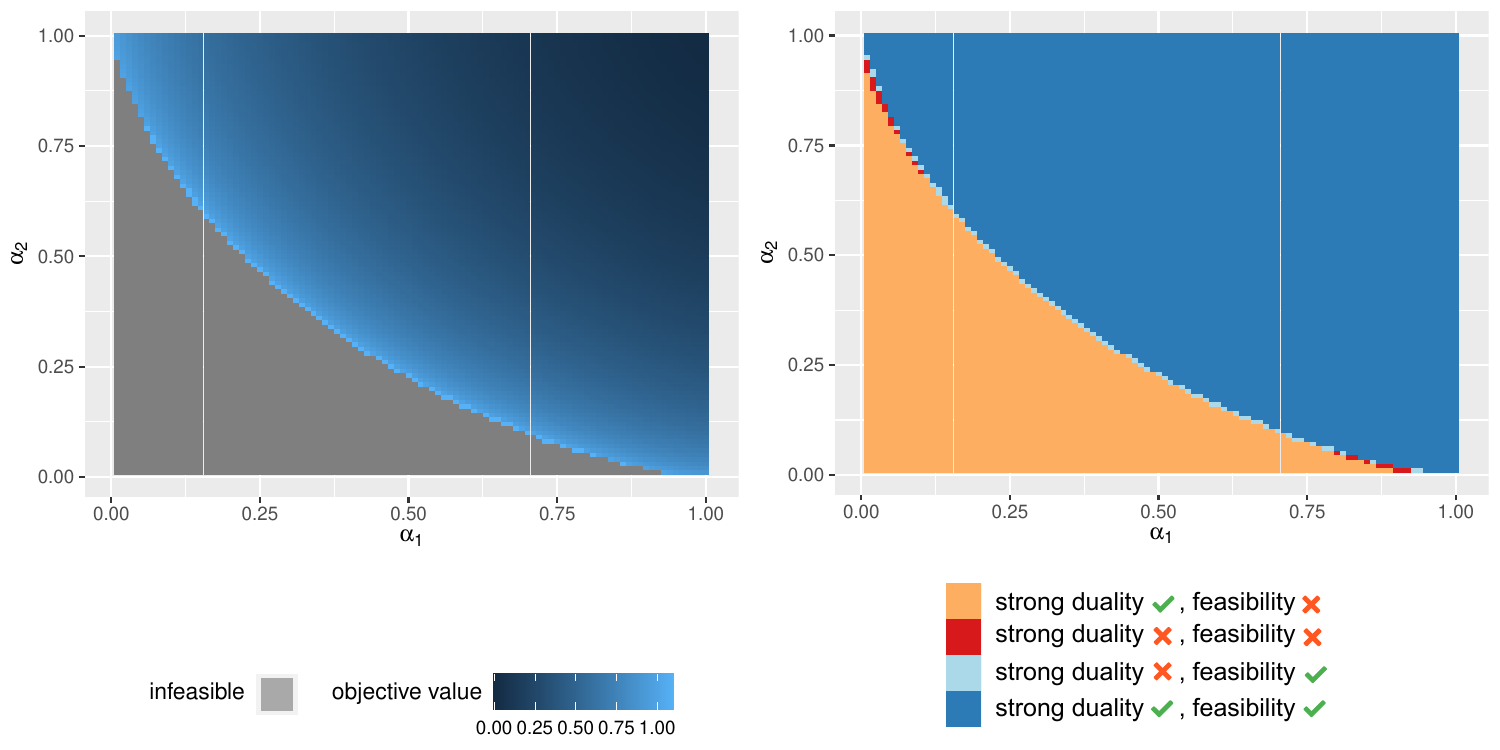}
	\caption{Strong duality and feasibility of NPMC problem for the LendingClub dataset with different target error  levels: predicted by Algorithm \ref{algo: df check CX} with NPMC-CX-logistic.}
	\label{fig: loanclub_NPMC_logistic}
\end{figure}

\begin{figure}[!h]
	\centering
	\includegraphics[width=\textwidth]{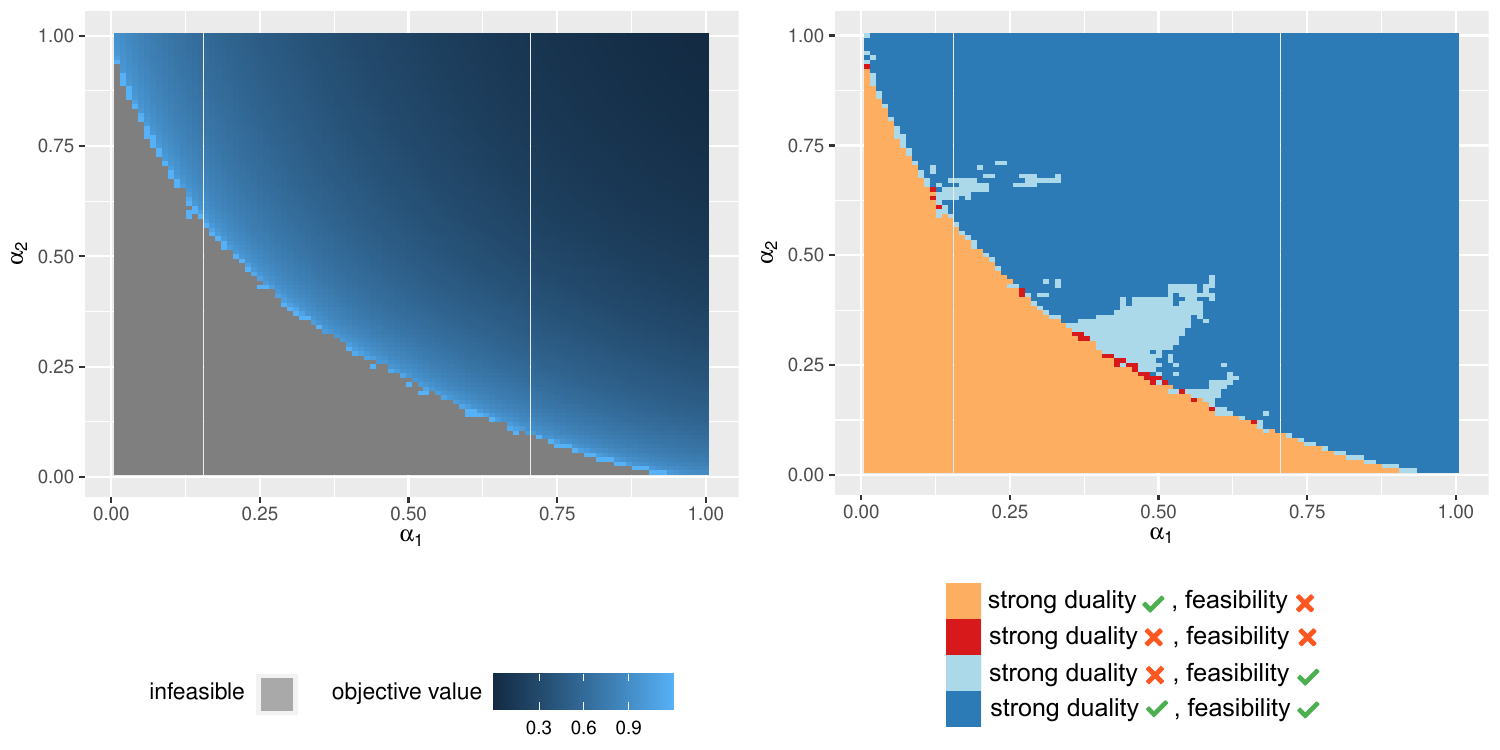}
	\caption{Strong duality and feasibility of NPMC problem for the LendingClub dataset with different target error  levels: predicted by Algorithm \ref{algo: df check ER} with NPMC-ER-RF.}
	\label{fig: loanclub_NPMC_rf}
\end{figure}

As described in Section \ref{subsec: simulations}, practitioners can experiment with various target levels $(\alpha_1, \alpha_2)$ using our feasibility and strong duality checking algorithms (Algorithms \ref{algo: df check CX} and \ref{algo: df check ER} in Section \ref{subsec: dk check NPMC supp}) to assess the problem's complexity and select the target level based on feasibility and practical considerations. We present the predicted strong duality, feasibility, and objective values using Algorithm \ref{algo: df check CX} with NPMC-CX-logistic (NPMC-CX with $\mathcal{M}$ as logistic regresion) and Algorithm \ref{algo: df check ER} with NPMC-ER-RF (NPMC-ER with $\mathcal{M}$ as random forests) on the entire dataset for different $(\alpha_1, \alpha_2) \in [0,1]^2$, in Figures \ref{fig: loanclub_NPMC_logistic} and \ref{fig: loanclub_NPMC_rf}, respectively. These figures illustrate the tradeoff between error rates for the three classes. To ensure the feasibility of the NP problem, the error thresholds $(\alpha_1, \alpha_2)$ must not be set too low. This requirement largely stems from the intrinsic complexity of the task, especially the challenge of distinguishing between classes 1 and 2. When logistic regression and random forests are trained solely on data from these classes, both methods display a binary misclassification error rate approaching 30\%, underscoring the inherent difficulty.

\begin{figure}[!h]
	\centering
	\includegraphics[width=\textwidth]{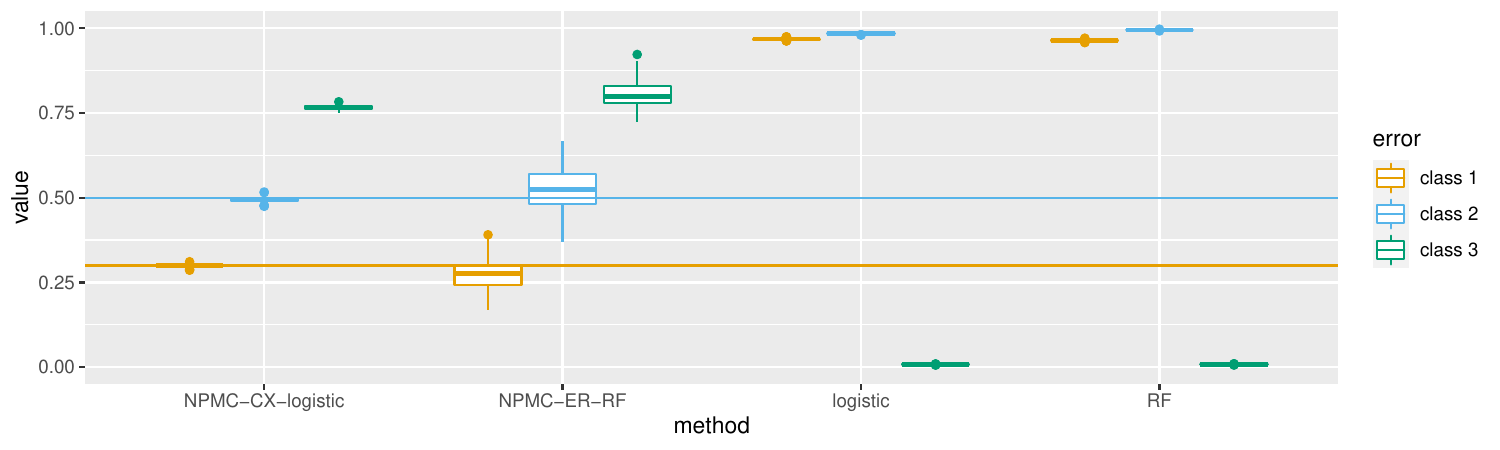}
	\caption{Per-class error rates and objective function values under each classifier for the NPMC problem on the LendingClub dataset. Horizontal lines in corresponding colors mark the target control levels.}
	\label{fig: loanclub_NPMC}
\end{figure}

Next, we fix $\alpha_1 = 0.3$ and $\alpha_2 = 0.5$, and conduct experiments with NPMC-CX-logistic and NPMC-ER-RF, alongside vanilla logistic regression and random forests as benchmarks. We randomly split the entire data into 50\% training and 50\% testing data over 500 replications. Box plots in Figure \ref{fig: loanclub_NPMC} display the per-class error rates under each classifier and across various training sample size settings. Notably, vanilla logistic regression and random forests tend to assign all observations to class 3 due to the significant imbalance in sample sizes. In contrast, NPMC-CX-logistic and NPMC-ER-RF effectively control the error rates of classes 1 and 2 around the specified target levels.

We also run similar experiments on the confusion matrix control problem outlined in Section \ref{subsec: np} and the detailed results are presented in Section \ref{subsubsec: lendingclub supp}.

\subsection{Comparison of NPMC-CX and NPMC-ER from experimental perspective}\label{subsec: comparison exp}
From the previous numerical results, we can observe that:
\begin{itemize}
	\item NPMC-CX works better under parametric models (e.g., logistic and LDA) by controlling the error rates well and achieving a lower objective function value compared to NPMC-ER, but can sometimes fail to control error rates under target levels for non-parametric models (e.g., $k$NN, RF, and SVM with RBF kernel).
	\item Compared to NPMC-CX, NPMC-ER requires a larger sample size to perform well due to sample splitting in Algorithm \ref{algo: erm}, but it is more robust to different model types.
\end{itemize}
These observations align well with our intuition from theoretical analysis (Section \ref{subsec: comparison theory}). Therefore, for practitioners, if a Rademacher class (usually parametric) is believed to be suitable for the problem at hand, we suggest using NPMC-CX. If the non-parametric model is believed to work better and sample size is not very small, we suggest using NPMC-ER.

\section{Discussions}\label{sec: discussion}

\subsection{Summary}
In this paper, we connect Neyman-Pearson multi-class classification (NPMC) problems with cost-sensitive learning (CS) problems, and propose two algorithms, NPMC-CX and NPMC-ER, to solve the NPMC problem \eqref{eq: NPMC} via CS techniques. To our knowledge, this is the first work solving NPMC problems with theoretical guarantees. We have presented some theoretical results, including conditions for strong duality and multi-class NP oracle properties for the two algorithms. Furthermore, we propose practical algorithms to verify the NPMC feasibility and strong duality, which can offer practitioners a landscape of the NPMC problem with various target error levels. Our algorithms are shown to be effective through extensive simulations and real data studies. 

Comparing NPMC-CX and NPMC-ER, we find:
\begin{itemize}
	\item Both algorithms are shown to satisfy multi-class NP properties. However, NPMC-CX necessitates a function class with a vanishing Rademacher complexity for estimating $\tp_{Y|X=\bx}(Y= k)$, while NPMC-ER has no such constraints.
	\item In practice, NPMC-CX works well for parametric models but may struggle with some non-parametric models. Due to data splitting, NPMC-ER requires a larger sample size but is more robust to diverse model types.
	\item Therefore, we suggest the practitioners go with NPMC-CX when a parametric model is favored. When the non-parametric model is believed to work better, and there is enough training data, we suggest using NPMC-ER.
\end{itemize}

Furthermore, the general confusion matrix control problem outlined in Section \ref{subsec: np} is discussed in detail in Section \ref{sec: general confusion matrix problem} of supplementary materials, and we extended our two NPMC algorithms to solve that problem. The theoretical results are also provided.

\vspace*{-0.2cm}
\subsection{Future research directions}
There are many interesting future avenues to explore. Here, we list three of them. 
\begin{enumerate}[(i)]
	\item There are many approaches to fitting a CS classifier. We use \eqref{eq: cs classifier given lambda empirical} to fit the CS classifier in our NPMC algorithms, which sometimes is called the thresholding strategy in binary CS problems \citep{dmochowski2010maximum}. It might be interesting to explore other approaches and replace \eqref{eq: cs classifier given lambda empirical} accordingly.
	\item \cite{li2020bridging} studied the methodological relationship between the binary NP paradigm and CS paradigm, and constructed a CS classifier with type-I error controls. In this paper, we focus on the multi-class NP paradigm and build a multi-class NP classifier via CS learning, which can be viewed as the inverse to \cite{li2020bridging}. Exploring the other direction in the multi-class cases would be interesting: developing multi-class CS classifiers with specific error controls.
	\item As one reviewer pointed out, the current multi-class NP oracle properties might not be strong enough in some degenerated cases where the NPMC problem can vary with $n$ and $J(\phi^*) = \smallo(1)$ or $\alpha_k = \smallo(1)$ for some $k \in \mathcal{A}$. It would be intriguing to generalize the existing multi-class NP oracle properties from $R_k(\hphi) \leq \alpha_k + \smallo_{\tp}(1)$, $\forall k\in \mathcal{A}$ and $J(\hphi) \leq J(\phi^*) + \smallo_{\tp}(1)$ to $R_k(\hphi) \leq \alpha_k + \smallo_{\tp}(\alpha_k)$, $\forall k\in \mathcal{A}$ and $J(\hphi) \leq J(\phi^*) + \smallo_{\tp}(J(\phi^*))$.
\end{enumerate}

\vspace*{-0.6cm}

\section*{Acknowledgments}

We thank the Co-Editor, the AE, and two anonymous reviewers for their insightful comments, which greatly improved a prior version of the paper. 


\spacingset{0.88} 

{\small
\bibliography{reference_np.bib}}
\bibliographystyle{apalike}
\spacingset{1.5} 

\newpage
\setcounter{page}{1}
\setcounter{section}{0}
\renewcommand{\thesection}{S.\arabic{section}}
\renewcommand{\theequation}{\thesection.\arabic{equation}}
\begin{center}
{\large\bf Supplementary Materials of ``Neyman-Pearson Multi-class Classification via Cost-sensitive Learning"}
\end{center}
\renewcommand{\cftsecnumwidth}{30pt}
\renewcommand{\cftsubsecnumwidth}{35pt}
\renewcommand{\cftsubsubsecnumwidth}{50pt}
\tableofcontents
\addtocontents{toc}{\protect\setcounter{tocdepth}{5}}

\section{Additional Details of NPMC Problem}\label{sec: additional NPMC supp}
In this section, we delve deeper into the NPMC problem, offering additional insights to complement the discussions in Section \ref{sec: theory}. In Section \ref{subsec: dk check NPMC supp}, we continue discussing checking strong duality and feasibility in practice, building upon the theoretical development in Section \ref{subsec: strong duality}. Two algorithms motivated by Corollary \ref{cor: df check} and the associated theory are presented. Beyond the NP oracle properties, we introduce the concept of strong consistency for NP algorithms in Section \ref{subsec: strong consistency supp}. By slightly enhancing the existing assumptions, we attain strong consistency for NPMC-CX. In Section \ref{subsec: discussions on assumptions}, we will validate most of these assumptions in a logistic regression problem, where the logistic regression model serves as the function class $\mathcal{M}$ to estimate the posterior $\tp_{Y|X}$ in both algorithms. Finally, in Section \ref{subsec: discussion np ht}, we further explore the underpinnings of the NPMC problem \eqref{eq: NPMC} and the connection to other problems. This discussion encompasses various aspects, including the interplay between hypothesis testing and NPMC, feasibility considerations, and the role of randomization.

\subsection{Checking strong duality and feasibility}\label{subsec: dk check NPMC supp}
In Section \ref{subsec: strong duality}, we outlined sufficient and necessary conditions for strong duality (Theorem \ref{thm: strong duality}) along with a powerful corollary (Corollary \ref{cor: df check}) that facilitates the design of algorithms to verify strong duality and feasibility for NPMC problems in practice. The fundamental idea, as discussed, is to replace $\blambda^*$ and $G(\blambda^*)$ in Corollary \ref{cor: df check} with $\hblambda$ and $\widehat{G}^{\textup{CX}}(\hblambda)$ from NPMC-CX or $\hblambda$ and $\widehat{G}^{\textup{ER}}(\hblambda)$ from NPMC-ER, leading to Algorithms \ref{algo: df check CX} and \ref{algo: df check ER}.



\begin{algorithm}[!h]
\setstretch{1.2}
\caption{Test strong duality and feasibility with NPMC-CX}
\label{algo: df check CX}
\KwIn{training data $\{(\bm{x}_i, y_i)\}_{i = 1}^n$, target upper bounds of errors $\balpha$, the weighting vector of objective function $\bw$, a function class $\mathcal{M}$ to estimate $\tp_{Y|X}$, a small constant $\delta \in (0, 1)$}
\KwOut{strong duality indicator $\hat{s}$ and feasibility indicator $\hat{f}$}
A finite $\hblambda$ and a corresponding CS classifier $\hphi$, or $\perp$ (we obtain a sequence $\{\hblambda^{(m)}\}_{m=1}^{\infty}$ with $\infnorm{\hblambda^{(m)}} \rightarrow +\infty$ and $\widehat{G}^{\textup{CX}}(\hblambda^{(m)}) \rightarrow +\infty$\footnotemark) $\leftarrow$ Steps 1, 3 of NPMC-CX with $\mathcal{M}$ \\
\uIf{we obtain finite $\hblambda$ and a corresponding CS classifier $\hphi$}{
	Check if $\widehat{R}_k(\hphi) \leq \alpha_k(1+\delta)$ for all $k \in \mathcal{A}$ and $\widehat{G}^{\textup{CX}}(\hblambda) \leq 1+\delta$, where $\widehat{R}_k(\hphi)$ is the empirical classification error calculated on $\{(\bm{x}_i, y_i)\}_{i = 1}^n$ \\
	\uIf{Yes}{
		Set $\hat{s} = 1$, $\hat{f} = 1$
	}\uElseIf{$\widehat{G}^{\textup{CX}}(\hblambda) \leq 1+\delta$}{
		Set $\hat{s} = 0$, $\hat{f} = 1$
	}\Else{Set $\hat{s} = 0$, $\hat{f} = 0$}
}\ElseIf{$\perp$}{
	Set $\hat{s} = 1$, $\hat{f} = 0$
}
Output $\hat{s}$ and $\hat{f}$
\end{algorithm}
\footnotetext{Note that $\widehat{G}^{\textup{CX}}(\blambda)$ is piecewise-linear in $\blambda$ with finite non-differentiable points, therefore $\sup_{\blambda \subseteq \mathbb{R}^{|\mathcal{A}|}_+}\widehat{G}^{\textup{CX}}(\blambda)$ is either finite and achievable at some finite $\hblambda$ or infinite and approached by a sequence $\{\hblambda^{(m)}\}_{m=1}^{\infty}$ with $\infnorm{\hblambda^{(m)}} \rightarrow +\infty$.}

\begin{algorithm}[!h]
\setstretch{1.2}
\caption{Test strong duality and feasibility with NPMC-ER}
\label{algo: df check ER}
\KwIn{training data $\{(\bm{x}_i, y_i)\}_{i = 1}^n$, target upper bounds of errors $\balpha$, the weighting vector of objective function $\bw$, a function class $\mathcal{M}$ to estimate $\tp_{Y|X}$, a small constant $\delta \in (0, 1)$, a search range $R > 0$, a treshold $R_G > 1+\delta$}
\KwOut{strong duality indicator $\hat{s}$ and feasibility indicator $\hat{f}$}
$\hblambda$, $\hphi \leftarrow$ Steps 1-3 of NPMC-ER with $\mathcal{M}$ and search range $R$\\
\uIf{we obtain a finite $\hblambda$ and a corresponding CS classifier $\hphi$}{
	Check if $\widehat{R}_k(\hphi) \leq \alpha_k(1+\delta)$ for all $k \in \mathcal{A}$ and $\widehat{G}^{\textup{ER}}(\hblambda) \leq 1+\delta$, where $\widehat{R}_k(\hphi)$ is the empirical classification error calculated on the splitted data set $\mathcal{D}_1$ in NPMC-ER\\
	\uIf{Yes}{
		Set $\hat{s} = 1$, $\hat{f} = 1$
	}\uElseIf{$\widehat{G}^{\textup{ER}}(\hblambda) \leq 1+\delta$}{
		Set $\hat{s} = 0$, $\hat{f} = 1$
	}\uElseIf{$\widehat{G}^{\textup{ER}}(\hblambda) > R_G$}{
		Set $\hat{s} = 1$, $\hat{f} = 0$
	}\Else{
		Set $\hat{s} = 0$, $\hat{f} = 0$
	}
}
Output $\hat{s}$ and $\hat{f}$
\end{algorithm}

We have the following result for Algorithms \ref{algo: df check CX} and \ref{algo: df check ER}, which assures that Algorithms \ref{algo: df check CX} and \ref{algo: df check ER} can accurately predict strong duality and feasibility with high probability,
 
\begin{theorem}\label{thm: dk check}
	Suppose $\{X|Y=k\}_{k=1}^K$ are continuous random variables (i.e. have Lebesgue density). Suppose the following conditions hold:
	\begin{itemize}
		\item For (\rom{1}) and (\rom{2}): Assumptions \ref{asmp: posterior distance}-\ref{asmp: marginal finite sample} (for Algorithm \ref{algo: df check CX}) or Assumptions \ref{asmp: posterior distance}, \ref{asmp: second order info}, \ref{asmp: marginal finite sample} (for Algorithm \ref{algo: df check ER}) hold, and $\delta \gtrsim [C_{\textup{Rad}}(n) + \max_k \te\norm{\widehat{\tp}_{Y|X}(Y=k)-\tp_{Y|X}(Y=k)}]^{\bar{\gamma}/2}$ (for Algorithm \ref{algo: df check CX}) or $\delta \gtrsim n^{-\bar{\gamma}/4}$ (for Algorithm \ref{algo: df check ER}).
		\item For (\rom{3}): Assumptions \ref{asmp: posterior distance}, \ref{asmp: rademacher} (for Algorithm \ref{algo: df check CX}) or Assumption \ref{asmp: posterior distance} (for Algorithm \ref{algo: df check ER}) hold, and $\delta \gtrsim [C_{\textup{Rad}}(n) + \max_k \te\norm{\widehat{\tp}_{Y|X}(Y=k)-\tp_{Y|X}(Y=k)}]^{1/2}$ (for Algorithm \ref{algo: df check CX}) or $\delta \gtrsim n^{-1/4}$ (for Algorithm \ref{algo: df check ER}).
		\item For (\rom{4}): Assumptions \ref{asmp: posterior distance}, \ref{asmp: second order info}, \ref{asmp: rademacher} (for Algorithm \ref{algo: df check CX}) or Assumptions \ref{asmp: posterior distance}, \ref{asmp: second order info} (for Algorithm \ref{algo: df check ER}) hold, and $\delta \gtrsim [C_{\textup{Rad}}(n) + \max_k \te\norm{\widehat{\tp}_{Y|X}(Y=k)-\tp_{Y|X}(Y=k)}]^{1/2}$ (for Algorithm \ref{algo: df check CX}) or $\delta \gtrsim n^{-1/4}$ (for Algorithm \ref{algo: df check ER}).
	\end{itemize}
	Then we have the following results for Algorithms \ref{algo: df check CX} and \ref{algo: df check ER}.
	\begin{enumerate}[(i)]
		\item If the NPMC problem is feasible and strong duality holds, then
			\begin{equation}
				\tp(\hat{f} = 1, \hat{s} = 1) \geq 1-C\exp\{-C'n\delta^{4/\bar{\gamma}}\} -C\delta^{\frac{-2\vee (1+\bar{\gamma})}{\bar{\gamma}}}\max_k \te|\widehat{\tp}_{Y|X}(Y=k) - \tp_{Y|X}(Y=k)|.
			\end{equation}
		\item Suppose the NPMC problem is feasible and strong duality fails.
			\begin{enumerate}
				\item For Algorithm \ref{algo: df check CX}, if $\blambda^*$ is finite and $R_k(\phi^*_{\blambda^*}) \leq \alpha_k(1+\delta)$, then with probability at least $1-C\exp\{-C'n\delta^{4/\bar{\gamma}}\} -C\delta^{\frac{-2\vee (1+\bar{\gamma})}{\bar{\gamma}}}\max_k \te|\widehat{\tp}_{Y|X}(Y=k) - \tp_{Y|X}(Y=k)|$, we will obtain $\hblambda$ and $\hphi$ from Step 1, and $R_k(\hphi) \leq \alpha_k(1+\frac{3}{2}\delta)$ for all $k \in \mathcal{A}$;
				\item For Algorithm \ref{algo: df check ER}, if $\infnorm{\blambda^*}\leq R$ and $R_k(\phi^*_{\blambda^*}) \leq \alpha_k(1+\delta)$, then with probability at least $1-C\exp\{-C'n\delta^{4/\bar{\gamma}}\} -C\delta^{\frac{-2\vee (1+\bar{\gamma})}{\bar{\gamma}}}\max_k \te|\widehat{\tp}_{Y|X}(Y=k) - \tp_{Y|X}(Y=k)|$, we will obtain $\hblambda$ and $\hphi$ from Step 1, and $R_k(\hphi) \leq \alpha_k(1+\frac{3}{2}\delta)$ for all $k \in \mathcal{A}$;
				\item If $\blambda^*$ is finite and $R_k(\phi^*_{\blambda^*}) > \alpha_k(1+\frac{3}{2}\delta)$, then
					\begin{equation}
						\tp(\hat{f} = 1, \hat{s} = 0) \geq 1-C\exp\{-C'n\delta^{4/\bar{\gamma}}\} -C\delta^{\frac{-2\vee (1+\bar{\gamma})}{\bar{\gamma}}}\max_k \te|\widehat{\tp}_{Y|X}(Y=k) - \tp_{Y|X}(Y=k)|.
					\end{equation} 
			\end{enumerate}
		\item If NPMC problem is infeasible and strong duality holds, then
			\begin{equation}
				\tp(\hat{f} = 0, \hat{s} = 1) \geq 1-C\exp\{-C'n\} -C\max_k \te|\widehat{\tp}_{Y|X}(Y=k) - \tp_{Y|X}(Y=k)|.
			\end{equation}
		\item If NPMC problem is infeasible and strong duality fails, $\blambda^*$ is finite, and $G(\phi^*_{\blambda^*}) > 1+\frac{3}{2}\delta$, then
			\begin{equation}
				\tp(\hat{f} = 0, \hat{s} = 0) \geq 1-C\exp\{-C'n\delta^{2}\} - C\delta^{-1}\max_{k}\te |\widehat{\tp}_{Y|X}(Y = k) - \tp_{Y|X}(Y = k)|.
			\end{equation}
	\end{enumerate}
\end{theorem}

\begin{figure}
	\centering
	\includegraphics[width=\textwidth]{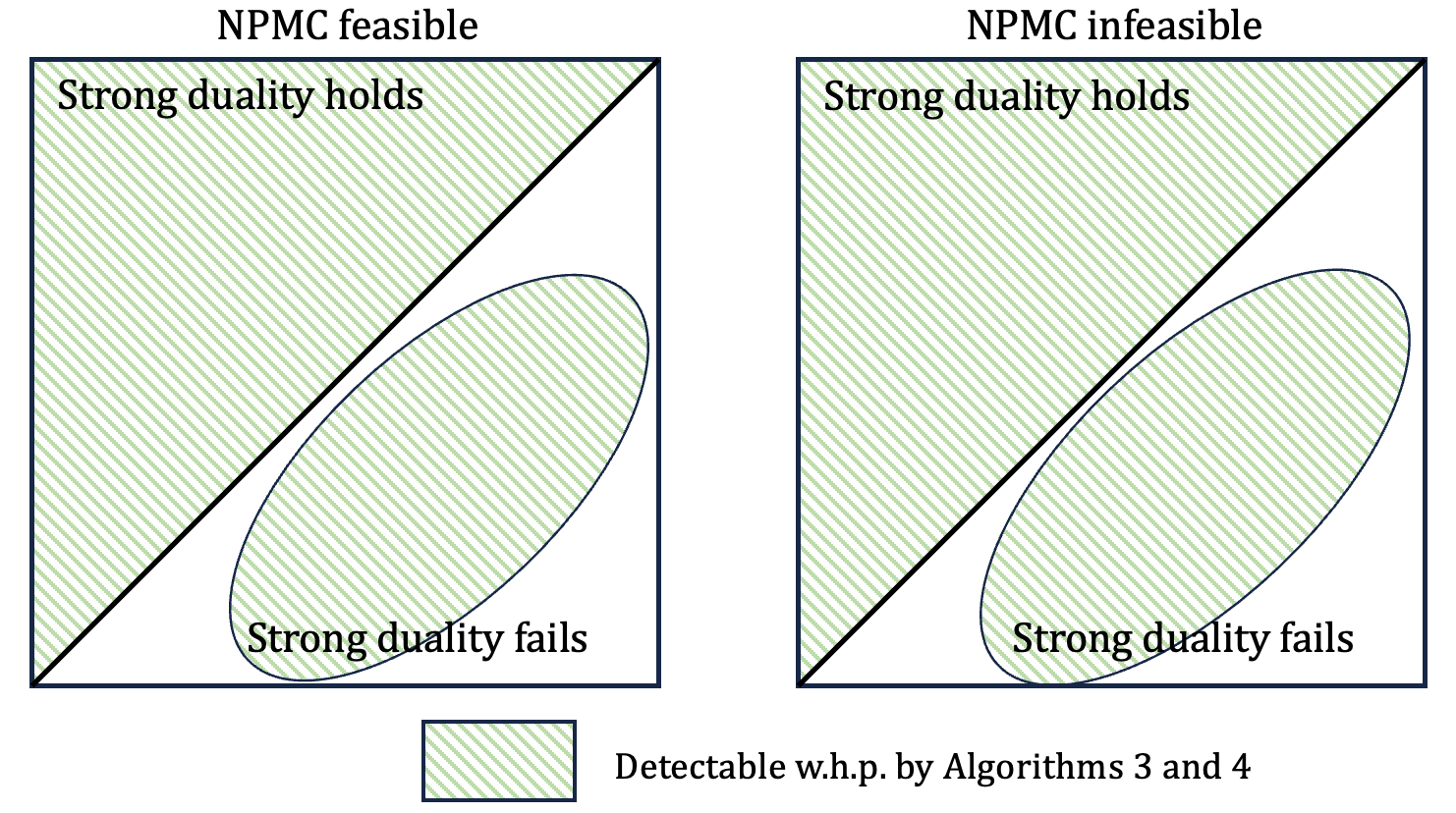}
	\caption{The detectable regions (for strong duality and feasibility) of Algorithms \ref{algo: df check CX} and \ref{algo: df check ER}. The green shaded area represents the regime where Algorithms \ref{algo: df check CX} and \ref{algo: df check ER} can correctly predict strong duality and feasibility with high probability (w.h.p.). }
	\label{fig: duality_feasibility_region}
\end{figure}

Theorem \ref{thm: dk check} guarantees that Algorithms \ref{algo: df check CX} and \ref{algo: df check ER} can correctly predict the strong duality and feasibility of NPMC problems with high probability, when strong duality holds. When strong duality fails, additional conditions are required to maintain similar high-probability guarantees. Surprisingly, according to the points (\rom{2}).(a) and (\rom{2}).(b), when the NPMC problem is feasible, even if strong duality fails, feasible classifiers can still be derived from NPMC-CX and NPMC-ER under certain circumstances. Therefore, in practice, we recommend the practitioners focus solely on checking the feasibility using Algorithms \ref{algo: df check CX} and \ref{algo: df check ER} (or equivalently, using the feasibility check in NPMC-CX and NPMC-ER presented in Algorithms \ref{algo} and \ref{algo: erm}). Figure \ref{fig: duality_feasibility_region} illustrates the regime where Algorithms \ref{algo: df check CX} and \ref{algo: df check ER} can correctly predict strong duality and feasibility with high probability.

Furthermore, the following result demonstrates that in instances where strong duality and feasibility are upheld, the empirical optimal value of the objective function can effectively estimate the true optimal value of the objective function in the dual CS problem. This provides a means to estimate the optimal objective value for the primal NPMC problem, when strong duality holds.

\begin{theorem}\label{thm: obj prediction dk check}
Under the same assumptions required by Theorem \ref{thm: dk check}, we have the following results for Algorithms \ref{algo: df check CX} and \ref{algo: df check ER}. To distinguish $\hblambda$ from NPMC-CX and NPMC-ER, we use different notations $\hblambda^{\textup{CX}}$ and $\hblambda^{\textup{ER}}$. 

If the NPMC problem is feasible and strong duality holds, then 
		\begin{align}
			&\tp\bigg(\bigg|\widehat{G}^{\textup{CX}}(\hblambda^{\textup{CX}}) - \sup_{\blambda \in \mathbb{R}^{|\mathcal{A}|}_+} G(\blambda) \bigg| \leq \tau_1\bigg) \\
			&\quad\geq 1- C\exp\{-Cn\tau_1^4\} - C\tau_1^{-2}\max_k \te\norma{\widehat{\tp}_{Y|X}(Y=k)-\tp_{Y|X}(Y=k)}, \\
			&\tp\bigg(\bigg|\widehat{G}^{\textup{ER}}(\hblambda^{\textup{ER}}) - \sup_{\blambda \in \mathbb{R}^{|\mathcal{A}|}_+} G(\blambda) \bigg| \leq \tau_2\bigg) \\
			&\quad\geq 1-C\exp\{-Cn\tau_2^4\} - C\tau_2^{-2}\max_k \te\norma{\widehat{\tp}_{Y|X}(Y=k)-\tp_{Y|X}(Y=k)},
		\end{align}
		when $1\geq \tau_1 \gtrsim \big[C_{\textup{Rad}}(n) + \max_k \te\norm{\widehat{\tp}_{Y|X}(Y=k)-\tp_{Y|X}(Y=k)}\big]^{1/2}$ and $1\geq \tau_2 \gtrsim n^{-1/4}$.
\end{theorem}

Note that when the strong duality in Assumption \ref{asmp: strong duality} holds, we have $\sup_{\blambda \in \mathbb{R}^{|\mathcal{A}|}_+} G(\blambda) = \min_{\phi \in \mathfrak{C}}J(\phi)$, therefore $\widehat{G}^{\textup{CX}}(\hblambda^{\textup{CX}})$ and $\widehat{G}^{\textup{ER}}(\hblambda^{\textup{ER}})$ can serve as reliable estimates for the objective function of the optimal NPMC classifier. This elucidates how the predicted objective values are calculated in Figures \ref{fig: loanclub_NPMC_logistic} and \ref{fig: loanclub_NPMC_rf}.

\subsection{Strong consistency}\label{subsec: strong consistency supp}
Besides the oracle properties introduced in Section \ref{subsec: np oracle properties}, we can also consider the following so-called strong consistency as another ideal property of NPMC algorithms. The definition is motivated by the strong consistency proposed for binary NP problems in \citeapp{scott2005neyman}.

\noindent\textbf{Strong consistency for the NPMC problem}:\vspace{-0.2cm}
\begin{enumerate}[(i)]
	\item If the NPMC problem is feasible and has an optimal solution $\phi^*$, then the algorithm outputs a solution $\hphi$ such that $\lim_{n \rightarrow \infty}R_k(\hphi)$ and $\lim_{n \rightarrow \infty}J(\hphi)$ exist $\forall k \in \mathcal{A}$, and
		\begin{enumerate}
			\item $\lim_{n \rightarrow \infty}R_k(\hphi) \leq \alpha_k$ a.s., $\forall k\in \mathcal{A}$;
			\item $\lim_{n \rightarrow \infty}J(\hphi) = J(\phi^*)$ a.s..
		\end{enumerate}
	\item Denote the event that the algorithm indicates infeasibility of NPMC problem given $\{(\bx_i, y_i)\}_{i=1}^n$ as $\mathcal{G}_n$. If the NPMC problem is infeasible, then $\tp(\liminf_{n \rightarrow \infty}\mathcal{G}_n) = 1$.
\end{enumerate}

Besides the conditions required for NP oracle properties, we require the function class used to estimate $\tp_{Y|X}$ to be indexed by a parameter $\bbeta \in \mathcal{B}$, where $\mathcal{B}$ is a compact set in a finite-dimensional metric space. Then imposing a stronger almost sure version of Assumption \ref{asmp: posterior distance} and another continuity condition (Assumption \ref{asmp: error rate continuity}) can imply strong consistency for NPMC-CX.

\begin{manualassumption}{2'}\label{asmp: posterior distance as}
	$\lim_{n \rightarrow \infty} \widehat{\tp}_{Y|X=\bx}(Y = k) = \tp_{Y|X=\bx}(Y = k)$ a.s. (w.r.t. the training data $\{(\bx_i, y_i)\}_{i=1}^n$) for almost everywhere $\bx$ (w.r.t. the distribution of $X$), for all $k$.
\end{manualassumption}

\begin{manualassumption}{4'}\label{asmp: error rate continuity}
	For almost all $\bx$ (w.r.t. the distribution of $X$), the estimated conditional probability $\widehat{\tp}_{Y|X=\bx}(Y=k;\bbeta)$ is a continuous function of the parameter $\bbeta$.
\end{manualassumption}

\begin{theorem}[Strong consistency of NPMC-CX]\label{thm: consistency}
	NPMC-CX satisfies strong consistency in the following senses.
	\begin{enumerate}[(i)]
		\item When the NPMC problem \eqref{eq: NPMC} is feasible, if Assumptions \ref{asmp: strong duality}, \ref{asmp: posterior distance as}, \ref{asmp: second order info} and \ref{asmp: error rate continuity} hold, then there exists a solution $\phi^*$, such that $\lim_{n\rightarrow \infty}R_k(\hphi) = R_k(\phi^*)$ a.s. for all $k$'s. 
		\item When the NPMC problem \eqref{eq: NPMC} is infeasible, if Assumptions \ref{asmp: strong duality}, \ref{asmp: posterior distance as} and \ref{asmp: error rate continuity} hold, then for any $M > 0$, $\liminf_{n \rightarrow \infty}\widehat{G}^{\textup{CX}}(\hblambda) > M$ a.s..
	\end{enumerate}
\end{theorem}

\subsection{Discussions on assumptions}\label{subsec: discussions on assumptions}
In Section \ref{sec: theory}, we impose a series of assumptions to show the NP oracle properties and strong consistency. Among these conditions, Assumption \ref{asmp: strong duality} is central and necessary to make the whole argument work. In general, since the original NPMC problem is not necessarily convex, it is challenging to demonstrate the strong duality. Theorem \ref{thm: strong duality} connects the strong duality with the feasibility of solutions to the CS problem under the NPMC problem, making the strong duality condition more explicit and transparent. Assumption \ref{asmp: posterior distance} requires the estimate $\widehat{\tp}_{Y|X}$ to be close to the true conditional probability $\tp_{Y|X}$, which is often trivial to hold when the estimator is constructed with the knowledge of the true model. Assumption \ref{asmp: rademacher} imposes restrictions on the Rademacher complexities of the function class $\mathcal{M}$ used to estimate $\tp_{Y|X}$, which can be verified for most parametric models and certain non-parametric models. Assumption \ref{asmp: marginal finite sample} is a margin condition and can be verified when the distribution $\tp_{X|Y = k}$ is nice. Regarding the assumptions for the strong consistency, Assumption \ref{asmp: error rate continuity} requires the continuity of conditional probability estimator w.r.t. the coefficient and can be easily verified for many parametric models.

Among these assumptions, Assumption \ref{asmp: strong duality} is generally hard to check. Nevertheless, thanks to Theorem \ref{thm: dk check}, we can assess the strong duality and feasibility in practice through Algorithms \ref{algo: df check CX} and \ref{algo: df check ER} in practice. Assumptions \ref{asmp: posterior distance}, \ref{asmp: posterior distance as}, \ref{asmp: second order info}, \ref{asmp: rademacher}, \ref{asmp: error rate continuity}, and \ref{asmp: marginal finite sample} can be checked given the estimated model and the underlying true model. Next, we verify them using the multinomial logistic regression model as an example.

Suppose the true conditional distribution of $Y$ given $X=\bx$ is $\tp_{Y|X}(Y=k) = \frac{\exp((\bbeta_k^*)^T \bx)}{\sum_{j=1}^K\exp((\bbeta_j^*)^T \bx)}$, where $k = 1, \ldots, K$, $\bbeta_k^* \in \mathbb{R}^p$ and $\bbeta_K^* = \bm{0}$. And we estimate it by $\widehat{\tp}_{Y|X}(Y=k) = \frac{\exp(\hbbeta_k^T \bx)}{\sum_{j=1}^K\exp(\hbbeta_j^T \bx)}$. Denote $\hbbeta = (\hbbeta_1, \ldots, \hbbeta_{K-1})$, which is the maximum likelihood estimator (MLE). In addition, suppose $X$ has bounded and continuously differentiable density function $f_X$ in $\mathbb{R}^p$, i.e. $f_X'$ is continuous and $\infnorm{f_X} < \infty$. Also assume $\te X_j^2$ exists for all $j = 1:p$, where $X_j$ is the $j$-th coordinate of $X$. 
\begin{itemize}
	\item First let's check Assumption \ref{asmp: posterior distance} and \ref{asmp: posterior distance as}. By similar arguments in \citeapp{wald1949note}, we can prove the MLE $\hbbeta$ is strongly consistent to $\bbeta$, i.e., $\lim_{n \rightarrow \infty} \hbbeta = \bbeta^*$ a.s., which verifies Assumption \ref{asmp: posterior distance as}. Then for any $\bx \in \mathbb{R}^p$, $\lim_{n\rightarrow \infty}\widehat{\tp}_{Y|X=\bx}(Y=k) = \tp_{Y|X=\bx}(Y=k)$ a.s.. Then by dominated convergence theorem, $\lim_{n\rightarrow \infty} \te_X\norm{\widehat{\tp}_{Y|X}(Y=k) - \tp_{Y|X}(Y=k)} = 0$ a.s., where the expectation $\te_X$ is w.r.t. $X$. This implies $\te_X\norm{\widehat{\tp}_{Y|X}(Y=k) - \tp_{Y|X}(Y=k)} \convp 0$. Then for any $\epsilon > 0$, let $\epsilon' = \epsilon/2 = 2\delta'$, such that $\tp(\te_X\norm{\widehat{\tp}_{Y|X}(Y=k) - \tp_{Y|X}(Y=k)} > \epsilon') \leq \delta'$ when $n > N(\epsilon)$. Therefore,
		\begin{equation}
			\te\left[\te_X\norm{\widehat{\tp}_{Y|X}(Y=k) = \tp_{Y|X}(Y=k)}\right] \leq \epsilon' + 2\tp\left(\te_X\norm{\widehat{\tp}_{Y|X}(Y=k) = \tp_{Y|X}(Y=k)} > \epsilon'\right) \leq \epsilon,
		\end{equation}
		when $n > N(\epsilon)$, which is equivalent to $\lim_{n\rightarrow \infty}\te[\te_X\norm{\widehat{\tp}_{Y|X}(Y=k) = \tp_{Y|X}(Y=k)}] = 0$. This verifies Assumption \ref{asmp: posterior distance}.
	\item Next let's verify the first part of Assumption \ref{asmp: second order info}, i.e. $G(\blambda)$ is continuously twice differentiable. Denote the conditional density of $X|Y=k$ as $f_{X|Y=k}$, then by Bayes rule, $f_{X|Y=k}(\bx) = \tp_{Y|X=\bx}(Y=k)f_X(\bx)/\pi_k^*$. According to \eqref{eq: f eq 1}, it suffices to show $\tp_{X|Y=k}(\phi^*_{\blambda}(X) = k)$ is continuously twice differentiable w.r.t. $\blambda$ at $\blambda^*$. In fact, we can show the twice continuous differentiability at any $\blambda$ with all $\lambda_k > 0$. To see this, consider $\tilde{\bbeta}_j = - \bbeta_j^* + \bbeta_k^*$ for $j \in \{1, \ldots, K\} \backslash \{k\}$, and we construct $\{\tilde{\bbeta}_j\}_{j = k, K+1,\ldots, p}$ to be linearly independent of $\{\tilde{\bbeta}_j\}_{j \in \{1, \ldots, K\} \backslash \{k\}}$. Let $\tilde{Z} = (\tilde{Z}_1, \ldots, \tilde{Z}_p)^T$, where $\tilde{Z} = BX$, $X = (X_1, \ldots, X_p)^T$, and $B = [\tilde{\bbeta}_1, \ldots, \tilde{\bbeta}_{p}]^T \in \mathbb{R}^{p \times p}$. By the construction procedure of $B$, $B$ is invertible. Through the linear transformation formula of densities, we know that the density of $\tilde{Z}$ is $f_{\tilde{Z}}(\bm{z}) = |B|^{-1}\cdot f_{X|Y=k}(B^{-1}\bm{z}) = \tp_{Y|X=B^{-1}\bm{z}}(Y=k)f_X(B^{-1}\bm{z})/\pi_k^*$, which is continuously differentiable w.r.t. $\bm{z} \in \mathbb{R}^p$. Denote $Z =  (Z_1,\ldots, Z_{K-1})^T = (\tilde{Z}_1, \ldots, \tilde{Z}_{k-1}, \tilde{Z}_{k+1}, \ldots, \tilde{Z}_K)^T \in \mathbb{R}^{K-1}$, which has the density $f_{Z}(\bm{z})$. By dominated convergence theorem, $f_{Z}(\bm{z})$ is continuously differentiable w.r.t. $\bm{z} \in \mathbb{R}^{K-1}$. Therefore,
		\begin{align}
			&\tp_{X|Y=k}(\phi^*_{\blambda}(X) = k) \\
			&= \tp_{X|Y=k}\left(c_k(\lambda_k, \pi^*_k)\cdot \tp_{Y|X=\bx}(Y=k) > \max_{j \neq k}\left[c_j(\lambda_j, \pi^*_j)\cdot \tp_{Y|X=\bx}(Y=j)\right]\right) \\
			&= \tp_{X|Y=k}\left(c_k(\lambda_k, \pi^*_k) > \max_{j \neq k}\left[c_j(\lambda_j, \pi^*_j)\cdot \exp\{\tilde{\bbeta}_j^TX\}\right]\right) \\
			&= \tp_{X|Y=k}\left(\log c_k(\lambda_k, \pi^*_k) > \max_{j \neq k}\left[\log c_j(\lambda_j, \pi^*_j) + \tilde{\bbeta}_j^TX\right]\right) \\
			&= \tp_{X|Y=k}\left(\bigcap_{j \neq k}\left\{ \tilde{\bbeta}_j^TX < \log c_k(\lambda_k, \pi^*_k) - \log c_j(\lambda_j, \pi^*_j)\right\}\right) \\
			&= \tp\left(Z_1 < z_{1}, \ldots, Z_{K-1} < z_{k-1} \right),
		\end{align}
		where $z_j(\lambda_k, \lambda_j) = \log c_k(\lambda_k, \pi^*_k) - \log c_j(\lambda_j, \pi^*_j)$ when $j < k$, and $z_j = \log c_k(\lambda_k, \pi^*_k) - \log c_{j+1}(\lambda_{j+1}, \allowbreak \pi^*_{j+1})$ when $j \geq k$. Next we will show $\frac{\partial^2 \tp\left(Z_1 < z_{1}, \ldots, Z_{K-1} < z_{k-1} \right)}{\partial \lambda_{j_1}\partial \lambda_{j_2}}$ exists and is continuous for any $j_1$ and $j_2$. For simplicity, we consider the case $k \geq 3$ and $j_1 = 1$, $j_2 = 2$. By straightforward calculations,
		\begin{align}
			&\frac{\partial^2 \tp\left(Z_1 < z_{1}, \ldots, Z_{K-1} < z_{k-1} \right)}{\partial \lambda_{1}\partial \lambda_{2}} \\
			&= \int_{-\infty}^{z_3(\lambda_k, \lambda_3)}\dots \int_{-\infty}^{z_{K-1}(\lambda_k, \lambda_{K-1})}f_{Z}(z_1(\lambda_k, \lambda_1), z_2(\lambda_k, \lambda_2), u_3, \ldots, u_{K-1}) du_3 \dots du_{K-1} \\
			&\quad \cdot  \frac{\partial z_1(\lambda_k, \lambda_1)}{\partial \lambda_1} \cdot  \frac{\partial z_2(\lambda_k, \lambda_2)}{\partial \lambda_2} \\
			&= \int_{-\infty}^{z_3(\lambda_k, \lambda_3)}\dots \int_{-\infty}^{z_{K-1}(\lambda_k, \lambda_{K-1})}f_{Z}(z_1(\lambda_k, \lambda_1), z_2(\lambda_k, \lambda_2), u_3, \ldots, u_{K-1}) du_3 \dots du_{K-1} \\
			&\quad \cdot [c_1(\lambda_1, \pi^*_1)\pi^*_1]^{-1}\cdot [c_1(\lambda_2, \pi^*_2)\pi^*_2]^{-1},
		\end{align}
		which exists and is continuous as long as $\lambda_1 > 0$ and $\lambda_2 > 0$ (avoiding the case that $c_1(\lambda_1, \pi^*_1) = 0$ or $c_2(\lambda_2, \pi^*_2) = 0$). Similarly, we can show that $\frac{\partial^2 \tp\left(Z_1 < z_{1}, \ldots, Z_{K-1} < z_{k-1} \right)}{\partial \lambda_{j_1}\partial \lambda_{j_2}}$ exists and is continuous for any $j_1$ and $j_2$, as long as $\lambda_j > 0$ for all $j$. Thus we proved the second-order continuously differentiability of $G(\blambda)$. Besides, by Proposition \ref{prop: concave G CX}, $G(\blambda)$ is concave, therefore we  know that $\nabla^2 G(\blambda) \preceq 0$. However, it is hard in general to show that $\nabla^2 G(\blambda) \prec 0$.  
	\item Next we check Assumption \ref{asmp: rademacher}. Recall that the class of estimated conditional probability $\mathcal{M} = \{\{\widehat{\tp}_{Y|X=\bx}(Y=k;\bbeta)\}_{k=1}^K: \bbeta = (\bbeta_1, \ldots, \bbeta_{K-1})^T \in \mathcal{B} \subseteq (\mathbb{R}^p)^{\otimes (K-1)}\}$ has the Rademacher complexity 
		\begin{equation}
			C_{\textup{Rad}}(n) \coloneqq \max_{k=1:K}\te_{\bm{\epsilon}}\te_{\{\bx_i\}_{i=1}^n} \sup_{\bbeta \in \mathcal{B}}\bigg|\frac{1}{n}\sum_{i=1}^n \epsilon_i\widehat{\tp}_{Y|X = \bx_i}(Y=k; \bbeta)\bigg|,
		\end{equation}
		where $\mathcal{B}$ is a compact subset of $(\mathbb{R}^p)^{\otimes (K-1)}$. Note that for logistic regression models, $\tp_{Y|X=\bx}(Y=k; \bbeta)$ is a Lipschitz function of $\{\bbeta^T_k\bx\}_{k=1}^{K-1} \in \mathbb{R}^{K-1}$ with a constant Lipschitz parameter w.r.t. $\ell_2$-metric in $\mathbb{R}^{K-1}$. Then, by applying Corollary 1 (a vector contraction inequality for Rademacher complexities) in \citeapp{maurer2016vector}, we immediately have
		\begin{align}
			\te_{\bm{\epsilon}}\te_{\{\bx_i\}_{i=1}^n} \sup_{\bbeta \in \mathcal{B}}\bigg|\frac{1}{n}\sum_{i=1}^n \epsilon_i\tp_{Y|X = \bx_i}(Y=k; \bbeta)\bigg| &\lesssim \sum_{k=1}^{K-1}\te_{\bm{\epsilon}}\te_{\{\bx_i\}_{i=1}^n} \sup_{\twonorm{\bbeta_k} \leq C'} \bigg|\frac{1}{n}\sum_{i=1}^n \epsilon_i(\bbeta_k^T\bx_i)\bigg| \\
			&\lesssim \sum_{k=1}^{K-1}\te_{\{\bx_i\}_{i=1}^n} \frac{1}{n}\sqrt{\sum_{i=1}^n \twonorm{\bx_i}^2} \\
			&\lesssim n^{-1/2},
		\end{align}
		where the last two inequalities are due to Jensen's inequality. Hence $C_{\textup{Rad}}(n) \lesssim n^{-1/2} \rightarrow 0$ as $n \rightarrow \infty$.
	\item Assumption \ref{asmp: error rate continuity} is trivial to hold by the format of $\widehat{\tp}_{Y|X}$. 
	\item Finally, let's verify Assumption \ref{asmp: marginal finite sample}. Without loss of generality, suppose $c_k^* = c_k(\blambda^*, \bpi^*) > 0$ for all $k$'s and $\underline{c} = (\min_{k}c_k^*)^{-1}$. And we only check $\tp_{X|Y=K}(\norm{\varphi_K(X)} \leq t) \lesssim t^{\bar{\gamma}}$ when $t$ is smaller than some constant $C \in (0, 1)$ and $\bar{\gamma} > 0$. $\tp_{X|Y=k}(\norm{\varphi_k(X)} \leq t) \lesssim t^{\bar{\gamma}}$ can be similarly discussed. Especially the simplest way is to change the reference level in the multinomial logistic regression model, as we did above, to verify Assumption \ref{asmp: second order info}. Note that
		\begin{align}
			&\norm{\varphi_K(X)} \leq t  \\
			&\Longleftrightarrow \norma{c_K^* - \max_{j \leq K-1}\{c_j^*e^{(\bbeta_j^*)^TX}\}} \leq t + t\sum_{j \leq K-1}e^{(\bbeta_j^*)^TX}\leq t+t(K-1) \underline{c}\max_{j \leq K-1}\{c_j^*e^{(\bbeta_j^*)^TX}\} \\
			&\Longleftrightarrow \frac{c_K^* -t}{1+t\underline{c}(K-1)} \leq \max_{j \leq K-1}\{c_j^*e^{(\bbeta_j^*)^TX}\} \leq \frac{c_K^* +t}{1-t\underline{c}(K-1)}.
		\end{align}
		Suppose $t < (\min_k c_k^*) \wedge (\underline{c}(K-1))^{-1}$. Denote the density of $(\bbeta^*_j)^TX$ as $\tilde{f}_j$. It is bounded by some constant $M > 0$ on $\mathbb{R}$ due to the boundedness of the density of $X$. Then, the marginal probability
		\begin{align}
			\tp(\norm{\varphi_K(X)} \leq t) &\leq \tp\left(\frac{c_K^* -t}{1+t\underline{c}(K-1)} \leq \max_{j \leq K-1}\{c_j^*e^{(\bbeta_j^*)^TX}\} \leq \frac{c_K^* +t}{1-t\underline{c}(K-1)}\right) \\
			&\leq \sum_{j=1}^{K-1}\tp\left(\frac{c_K^* -t}{1+t\underline{c}(K-1)} \leq c_j^*e^{(\bbeta_j^*)^TX} \leq \frac{c_K^* +t}{1-t\underline{c}(K-1)}\right) \\
			&= \sum_{j=1}^{K-1}\tp\left(\log \left(\frac{c_K^* -t}{c_j^*[1+t\underline{c}(K-1)]}\right) \leq (\bbeta_j^*)^TX \leq \log \left(\frac{c_K^* +t}{c_j^*[1-t\underline{c}(K-1)]}\right)\right) \\
			&\leq \sum_{j=1}^{K-1} \tilde{f}_j(\xi_{j, t})\left[\log \left(\frac{c_K^* +t}{c_j[1-t\underline{c}(K-1)]}\right) - \log \left(\frac{c_K^* -t}{c_j^*[1+t\underline{c}(K-1)]}\right)\right]\\
			&\leq (K-1)MC'\norma{\frac{c_K^* +t}{c_j^*[1-t\underline{c}(K-1)]}- \frac{c_K^* -t}{c_j^*[1+t\underline{c}(K-1)]}} \\
			&\leq Ct,
		\end{align}
		where $C$ and $C'$ are some positive constants and $\xi_{j, t}$ is some constant falling between $\log \left(\frac{c_k^* -t}{c_j^*[1+t\underline{c}(K-1)]}\right)$ and $\log \left(\frac{c_k^* +t}{c_j^*[1-t\underline{c}(K-1)]}\right)$. Therefore, Assumption \ref{asmp: marginal finite sample} holds with $\bar{\gamma} = 1$.
\end{itemize}

Assumption \ref{asmp: rademacher} can also be verified with some non-parametric classes. For example, when $\mathcal{M} = \{\{\widehat{\tp}_{Y|X=\bx}(y = k;\bbeta)\}: \widehat{\tp}_{Y|X=\bx}(y = k;\bbeta) \textup{ is } L\text{-Lipschitz in }\bx, \bbeta \in \mathcal{B}\}$ with $L > 0$ and $\tp_X$ has a bounded support in $\mathbb{R}^p$, then by the analysis in Example 13.10 of \citeapp{wainwright2019high}, we can verify that Assumption \ref{asmp: rademacher} holds.

The estimation error $\max_{k}\te|\widehat{\tp}_{Y|X}(Y = k) - \tp_{Y|X}(Y = k)|$ is a crucial part of the convergence rates of NPMC-CX and NPMC-ER. In the logistic regression example above, we verified that this estimation error indeed vanishes as $n \rightarrow \infty$, which verifies Assumption \ref{asmp: posterior distance}. Moreover, standard non-asymptotic analysis on M-estimators shows that the estimation error of $\bbeta^*$ satisfies $\max_{k}\te \twonorm{\hbbeta^{(k)}-\bbeta^{(k)*}} \lesssim n^{-1/2}$, which implies $\max_{k}\te|\widehat{\tp}_{Y|X}(Y = k) - \tp_{Y|X}(Y = k)| \lesssim \max_{k}\te\twonorm{\hbbeta^{(k)}-\bbeta^{(k)*}} \lesssim n^{-1/2} \rightarrow 0$ by Lipschizness of logistic functions and the bounded domain of $X$. We can also verify this assumption for some non-parametric classes. For example, consider kernel regression with box kernels $K(x) = \mathds{1}(|x| \leq h)$ and bandwidth $h \asymp n^{-1/(d+2)}$. By standard results in classical non-parametrics (e.g., see \citealpapp{tsybakov2009introduction}), we have $\max_{k}\te|\widehat{\tp}_{Y|X}(Y = k) - \tp_{Y|X}(Y = k)|\asymp n^{-1/(d+2)} \rightarrow 0$.

\subsection{Discussions on NPMC problem, hypothesis testing, and randomization}\label{subsec: discussion np ht}
In Section \ref{subsec: np}, we have connected the NPMC problem \eqref{eq: NPMC} to the distributional hypothesis testing problem with a composite null hypothesis consisting of finite arguments and listed their similarities and differences. Here, we would like to provide some additional insights. In the hypothesis testing problem, suppose that we have collected some data $X_n = \{x_i\}_{i=1}^n \sim$ some distribution $\tp$ and we would like to test 
\begin{equation}
	H_0: \tp \in \{\tp^{(k)}\}_{k=1}^K \, \text{ v.s. } H_1: \tp = \tp^{K+1}.
\end{equation}
We want to maximize the statistical power of the test while controlling the type-I error in every case under a target level $\alpha$. This can be formulated as an optimization problem over a \textit{deterministic} testing function $\varphi: X_n \rightarrow \{0,1\}$, where $\varphi(X_n) = 0$ or $1$ means accepting or rejecting $H_0$, respectively. For simplicity, assume $\{\tp^{(k)}\}_{k=1}^{K+1}$ has a family of corresponding densities $\{p_k\}_{k=1}^{K+1}$ under a base measure $\mu$ (e.g. Lebesgue measure). Then, the hypothesis testing problem is
\begin{align}
	&\max_{\varphi} \int \varphi(X_n)p_{K+1}(X_n)d \mu \\
	&\text{s.t.} \max_{k = 1:K}\int \varphi(X_n)p_k(X_n)d \mu \leq \alpha. \label{eq: HT}
\end{align}
In general, directly solving \eqref{eq: HT} is challenging. People usually connect it to the dual problem, where we try to find the \textit{least favorable distribution} (LFD, \citealpapp{lehmann1986testing}) $\{q_k\}_{k=1}^K$ on $\{p_k\}_{k=1}^{K}$ satisfying $q_k \geq 0$ and $\sum_{k=1}^K q_k = 1$ by solving
\begin{align}
	&\min_{\{q_k\}_{k=1}^K}\max_{\varphi} \int \varphi(X_n)p_{K+1}(X_n)d \mu \\
	&\text{s.t.} \int \bigg[\sum_{k=1}^Kq_kp_k(X_n)\bigg]\varphi(X_n)d \mu \leq \alpha. \label{eq: HT lfd}
\end{align}
In problem \eqref{eq: HT lfd}, given any $\{q_k\}_{k=1}^K$, the best testing function $\varphi$ can be obtained through the NP lemma. Comparing the CS classification problem \eqref{eq: dual of NPMC} with the dual of hypothesis testing problem \eqref{eq: HT lfd}, the $\lambda_k$'s play a similar role as the $\{q_k\}_{k=1}^K$ does. The underlying ideas are quite similar, i.e., using strong duality to connect the primal and dual problems, then solving the dual, which leads to a primal optimal solution. Nevertheless, as we mentioned earlier in Section \ref{subsec: np}, the two problems have some intrinsic differences. Moreover, in the hypothesis testing problem, since $\{p_k\}_{k=1}^{K+1}$ is known, in many cases, we can first guess the LFD and then verify it. However, in the NP problem, this is impossible since we do not have access to the distribution of $X$ given $Y=k$. 

Note that in both the NPMC problem \eqref{eq: NPMC} and the hypothesis testing problem \eqref{eq: HT lfd}, we are considering \textit{deterministic} classifiers and testing functions. The family of deterministic classifiers or testing functions works well when the strong duality holds. But in some cases, especially when the distribution of $X$ given $Y=k$ (for the NPMC problem) and the distribution of $X_n$ given $\theta=\theta_k$ (for the hypothesis testing problem) are not absolutely continuous, the strong duality could break under the family of deterministic classifiers or testing functions. In this case, for the hypothesis testing problem, \textit{randomization} has been shown to be a powerful method to solve the issue. For example, if we consider an enriched family of \textit{randomized} testing functions $\{\varphi: \varphi = \varphi_1 \text{ with probability } \omega \text{ and } \varphi = \varphi_2 \text{ probability } (1-\omega), 0\leq \omega \leq 1, \varphi_1,\varphi_2 \text{ map } X_n \text{ to } \{0, 1\}\}$, then the strong duality will hold again, which motivates the randomization part in the NP lemma. For the NPMC problem \eqref{eq: NPMC}, for simplicity, most of our theoretical results assume the strong duality (Assumption \ref{asmp: strong duality}) holds. In this situation, randomization does not help. However, it may help when the strong duality breaks, in which case an enlarged family of randomized classifiers may fix the strong duality. For simplicity, we do not discuss this case in the current paper and leave it to future studies.

Different from the hypothesis testing problem \eqref{eq: HT}, which is essentially a binary problem and always feasibile, the feasibility of the NPMC problem \eqref{eq: NPMC} is not always guaranteed. However, if the strong duality holds (which can be characterized by Theorem \ref{thm: strong duality}), then an unbounded dual problem \eqref{eq: dual of NPMC} is \textit{equivalent} to an infeasible primal problem \eqref{eq: NPMC}. If the strong duality does not hold, then by weak duality, an unbounded dual problem \eqref{eq: dual of NPMC} \textit{implies} an infeasible primal problem \eqref{eq: NPMC}.

\section{Extension to Confusion Matrix Control Problem}\label{sec: general confusion matrix problem}
In Section \ref{subsec: np}, we proposed two extensions of the binary NP problem. Previously, we have comprehensively studied the first one. In this section, we consider the other extension that leads to an entry-wise confusion matrix control problem. We will show how our previous algorithms and theory can be extended to this new setting. 

For any classifier $\phi$, we denote the component of confusion matrix at $k$-th row and $r$-th column as $R_{kr}(\phi) = \tp_{X|Y=k}(\phi(X) = r)$, where $r$, $k = 1, \ldots, K$. We may abuse the notations used in the previous sections, and the readers shall keep in mind that we are discussing a different version of the NP problem in this section.

We are interested in the following \textit{generalized} Neyman-Pearson \textit{multi-class} classification (GNPMC) problem:
\begin{align}
	&\min_{\phi} \quad J(\phi) = \sum_{k=1}^K\sum_{r \neq k} w_{kr} \tp_{X|Y=k}(\phi(X)= r) \\
	&\text{s.t.} \quad \tp_{X|Y=k}(\phi(X)= r) \leq \alpha_{kr}, \quad (k,r) \in \mathcal{A}, \label{eq: G-NPMC}
\end{align}
where $\phi: \mathcal{X} \rightarrow \{1, \ldots, K\}$ is a classifier, $\alpha_{kr} \in (0, 1)$, $w_{kr} \geq 0$, and $\sum_{k\neq r}w_{kr} = 1$, and $\mathcal{A} \subseteq (\{1, \ldots, K\} \times \{1, \ldots, K\})\backslash \{(k, k): 1\leq k \leq K\}$. The NPMC problem \eqref{eq: NPMC} we defined in Section \ref{sec: introduction} can be viewed as a simplified version of problem \eqref{eq: G-NPMC}. Compared to the NPMC problem, GNPMC problem allows more sophisticated error controls.

We want to connect \eqref{eq: G-NPMC} to the following cost-sensitive (CS) multi-class classification problem:
\begin{equation}\label{eq: general G cs problem}
	\min_{\phi} \quad \textup{Cost}(\phi) = \sum_{k=1}^K\sum_{r \neq k} \pi_k^* c_{kr} \tp_{X|Y=k}(\phi(X)= r),
\end{equation}
where $\phi: \mathcal{X} \rightarrow \{1, \ldots, K\}$ and $c_{kr} \geq 0$. 

Similar to Lemma \ref{lemma: cs oracle classifier}, we can define an optimal classifier of problem \eqref{eq: general G cs problem} from the costs and conditional probabilities $\{\tp_{Y|X=\bx}(Y=k)\}_{k=1}^K$.

\begin{lemma}\label{lemma: cs oracle classifier confusion matrix}
	Define classifier $\bar{\phi}^*: \bx \mapsto \argmin_{r=1:K} \{\sum_{k\neq r}c_{kr}\tp_{Y|X=\bx}(Y = k)\}$. Then $\bar{\phi}^*$ is one of the optimal classifiers of \eqref{eq: general G cs problem} in the following sense: for any classifier $\phi$, $\textup{Cost}(\bar{\phi}^*) \leq \textup{Cost}(\phi)$.
\end{lemma}

With Lemma \ref{lemma: cs oracle classifier confusion matrix} in hand, we can successfully extend our algorithms NPMC-CX and NPMC-ER to the confusion matrix control problem. Imposing similar assumptions as in the simplified case discussed in Section \ref{sec: theory}, we can prove that NPMC-CX satisfies the multi-class NP oracle properties and strong consistency, and NPMC-ER satisfies the multi-class NP oracle properties, under the generalized framework. Before getting into the details, we first extend the multi-class NP oracle properties and strong consistency described in Section  \ref{subsec: np oracle properties} for the GNPMC problem \eqref{eq: G-NPMC}.

\vspace{0.2cm}
\noindent\textbf{Multi-class NP oracle properties for the GNPMC problem}:\vspace{-0.2cm}
\begin{enumerate}[(i)]
	\item If the GNPMC problem is feasible and has an optimal solution $\phi^*$, then the algorithm outputs a solution $\hphi$ which satisfies
		\begin{enumerate}
			\item $R_{kr}(\hphi) \leq \alpha_{kr} + \mathcal{O}_{\tp}(\epsilon(n))$, $\forall (k,r)\in \mathcal{A}$;
			\item $J(\hphi) \leq J(\phi^*) + \mathcal{O}_{\tp}(\epsilon_J(n))$,
		\end{enumerate}
		where $\epsilon(n)$ and $\epsilon_J(n) \rightarrow 0$ as $n \rightarrow \infty$.
	\item Denote the event that the algorithm indicates infeasibility of GNPMC problem given $\{(\bx_i, y_i)\}_{i=1}^n$ as $\mathcal{G}_n$. If the GNPMC problem is infeasible, then $\tp(\mathcal{G}_n) \rightarrow 1$, as $n \rightarrow \infty$.
\end{enumerate}

\noindent\textbf{Strong consistency for the GNPMC problem}:\vspace{-0.2cm}
\begin{enumerate}[(i)]
	\item If the GNPMC problem is feasible and has an optimal solution $\phi^*$, then the algorithm outputs a solution $\hphi$ such that $\lim_{n \rightarrow \infty}R_{kr}(\hphi)$ and $\lim_{n \rightarrow \infty}J(\hphi)$ exist $\forall (k,r)\in \mathcal{A}$, and
		\begin{enumerate}
			\item $\lim_{n \rightarrow \infty}R_{kr}(\hphi) \leq \alpha_{kr}$ a.s., $\forall (k,r)\in \mathcal{A}$;
			\item $\lim_{n \rightarrow \infty}J(\hphi) = J(\phi^*)$ a.s..
		\end{enumerate}
	\item Denote the event that the algorithm indicates infeasibility of GNPMC problem given $\{(\bx_i, y_i)\}_{i=1}^n$ as $\mathcal{G}_n$. If the NP problem is infeasible, then $\tp(\liminf_{n \rightarrow \infty}\mathcal{G}_n) = 1$.
\end{enumerate}

Since the intuition and most parts of the derivation for the GNPMC problem \eqref{eq: G-NPMC} are similar to those for the NPMC problem \eqref{eq: NPMC}, we present the algorithms and the associated theory directly without deriving from sketches.
\subsection{Two algorithms: GNPMC-CX and GNPMC-ER}
For problem \eqref{eq: G-NPMC}, consider its Lagrangian form as
\begin{align}
	L(\blambda, \phi) &= \sum_{(k,r) \notin \mathcal{A}}w_{kr}\tp_{X|Y=k}(\phi(X)= r) + \sum_{(k,r) \in \mathcal{A}}(w_{kr} + \lambda_{kr})\tp_{X|Y=k}(\phi(X)= r) -\sum_{(k,r) \in \mathcal{A}}\lambda_{kr}\alpha_{kr} \\
	&= \sum_{r=1}^K \sum_{k \neq r}c_{kr}(\blambda, \bpi^*)\tp_{X,Y}(\phi(X)= r, Y=k) -\sum_{(k,r) \in \mathcal{A}}\lambda_{kr}\alpha_{kr} \\
	&= \te_X \bigg[\sum_{k \neq \phi(X)}c_{k\phi(X)}(\blambda, \bpi^*)\tp_{Y|X}(Y=k)\bigg] -\sum_{(k,r) \in \mathcal{A}}\lambda_{kr}\alpha_{kr},\label{eq: f eq 1 gnpmc}
\end{align}
where $\blambda = \{\lambda_{kr}\}_{(k,r) \in \mathcal{A}}$ and 
\begin{equation}
	c_{kr}(\blambda, \bpi^*) = \begin{cases}
		w_{kr}/\pi_k^*, \quad &(k,r) \notin \mathcal{A};\\
		(w_{kr}+\lambda_{kr})/\pi_k^*, \quad &(k,r) \in \mathcal{A}.
	\end{cases}
\end{equation}

Therefore, we can define the dual problem as $\max_{\blambda \in \mathbb{R}_+^{|\mathcal{A}|}}G(\blambda)$, where $G(\blambda) = \min_{\phi}L(\blambda, \phi) = L(\blambda, \phi^*)$ and $\phi^*_{\blambda}=\argmin_{r=1:K}\{\sum_{k \neq r}c_{kr}(\blambda, \bpi^*)\tp_{Y|X=\bx}(Y=k)\}$ by Lemma \ref{lemma: cs oracle classifier confusion matrix}.
Define 
\begin{equation}
	c_{kr}(\blambda, \hbpi) = \begin{cases}
		w_{kr}/\hpi_k, \quad &(k,r) \notin \mathcal{A};\\
		(w_{kr}+\lambda_{kr})/\hpi_k, \quad &(k,r) \in \mathcal{A}.
	\end{cases}
\end{equation}
Then we can extend the NPMC-CX and NPMC-ER to the algorithms GNPMC-CX and GNPMC-ER for the GNPMC problem \eqref{eq: G-NPMC} in a straightforward way. 

Similar to NPMC-CX, GNPMC-CX estimates $L(\blambda, \phi)$, $G(\blambda)$, and $\phi^*_{\blambda}$ by
\begin{align}
	\hL^{\textup{CX}}(\blambda, \phi) &= \frac{1}{n}\sum_{i=1}^n \sum_{k \neq \phi(\bx_i)} c_{k\phi(\bx_i)}(\blambda, \hbpi)\widehat{\tp}_{Y|X=\bx_i}(Y = k) -\sum_{(k,r) \in \mathcal{A}}\lambda_{kr}\alpha_{kr}, \\ \label{eq: F lambda hat GNPMC}
	\widehat{G}^{\textup{CX}}(\blambda) &\coloneqq \widehat{G}^{\textup{CX}}(\blambda; \widehat{\tp}_{Y|X}, \hbpi) = \min_{\phi}\hL^{\textup{CX}}(\blambda, \phi) = \hL^{\textup{CX}}(\blambda, \hphi_{\blambda}),\\
	\hphi_{\blambda}: \bx &\mapsto \argmin_{r=1:K}\bigg\{\sum_{k \neq r}c_{kr}(\blambda, \hbpi)\widehat{\tp}_{Y|X=\bx}(Y=k)\bigg\},
\end{align}
respectively, where $\widehat{\tp}_{Y|X}$ is the estimated conditional probability, and $\widehat{\tp}_{Y|X}$ can be obtained by fitting different models on the data.  

For GNPMC-ER, assume the training sample size to be $2n$. Similar to NPMC-ER, we divide the training data randomly into two parts of size $n$. For simplicity, denote them as $\mathcal{D}_1 = \{(\bx_i, y_i)\}_{i=1}^n = \{\{(\bm{x}_i^{(k)}, y_i^{(k)})\}_{i = 1}^{n_k}\}_{k=1}^K$ \footnote{$\{(\bm{x}_i^{(k)}, y_i^{(k)})\}_{i = 1}^{n_k}$ are the observations from class $k$ in $\mathcal{D}_1$.} and $\mathcal{D}_2 = \{(\bx_i, y_i)\}_{i=n+1}^{2n}$. Similar to NPMC-ER, GNPMC-ER estimates $L(\blambda, \phi)$, $G(\blambda)$, and $\phi^*_{\blambda}$ by
\begin{align}
	\hL^{\textup{ER}}(\blambda, \phi) &= \sum_{(k,r) \notin \mathcal{A}}w_{kr}\cdot \frac{1}{n_k}\sum_{i=1}^{n_k}\mathds{1}(\phi(\bx_i^{(k)})=r) + \sum_{(k,r) \in \mathcal{A}}(w_{kr} + \lambda_{kr})\cdot \frac{1}{n_k}\sum_{i=1}^{n_k}\mathds{1}(\phi(\bx_i^{(k)})=r) \\
	&\quad -\sum_{(k,r) \in \mathcal{A}}\lambda_{kr}\alpha_{kr}, \\
	\widehat{G}^{\textup{ER}}(\blambda) &\coloneqq \widehat{G}^{\textup{ER}}(\blambda; \widehat{\tp}_{Y|X}, \hbpi) = \hL^{\textup{ER}}(\blambda, \hphi_{\blambda}), \\
	\hphi_{\blambda}: \bx &\mapsto \argmin_{r=1:K}\bigg\{\sum_{k \neq r}c_{kr}(\blambda, \hbpi)\widehat{\tp}_{Y|X=\bx}(Y=k)\bigg\},
\end{align}
respectively. Note that $\hL^{\textup{ER}}$ is calculated on $\mathcal{D}_1$, while $\widehat{\tp}_{Y|X}$ and $\hbpi$ (hence $\hphi_{\blambda}$) are calculated on $\mathcal{D}_2$.

Details of the two algorithms are presented in Algorithms \ref{algo: gnpmc-cx} and \ref{algo: gnpmc-er}, respectively.

\begin{algorithm}[!h]
\caption{GNPMC-CX}
\label{algo: gnpmc-cx}
\KwIn{training data $\{(\bm{x}_i, y_i)\}_{i = 1}^n$, target upper bounds of errors $\balpha$, the weighting vector of objective function $\bw$, a function class $\mathcal{M}$ to estimate $\tp_{Y|X}$, a small constant $\delta > 0$}
\KwOut{the fitted classifier $\hphi$ or report the GNPMC problem as infeasible}
$\widehat{\tp}_{Y|X}, \hbpi \leftarrow$ the estimates of $\tp_{Y|X}$ (through $\mathcal{M}$) and $\bpi^*$ on training data $\{(\bm{x}_i, y_i)\}_{i = 1}^n$\\
\uIf{$\sup_{\blambda \in \mathbb{R}_{+}^{|\mathcal{A}|}} \widehat{G}^{\textup{CX}}(\blambda; \widehat{\tp}_{Y|X}, \hbpi) \leq 1+\delta$}{
$\hblambda \leftarrow \argmax_{\blambda \in \mathbb{R}_{+}^{|\mathcal{A}|}} \widehat{G}^{\textup{CX}}(\blambda; \widehat{\tp}_{Y|X}, \hbpi)$\\
Report the NP problem as feasible and output the solution $\hphi(\bx) = \argmin_{r=1:K} \{\sum_{k \neq r}c_{kr}(\blambda, \hbpi)\widehat{\tp}_{Y|X=\bx}(Y = k)\}$\\
}
\Else{Report the GNPMC problem as infeasible}
\end{algorithm}

\begin{algorithm}[!h]
\caption{GNPMC-ER}
\label{algo: gnpmc-er}
\KwIn{training data $\{(\bm{x}_i, y_i)\}_{i = 1}^{2n}$, target upper bound of errors $\balpha$, the weighting vector of objective function $\bw$, a search range $R > 0$, a function class $\mathcal{M}$ to estimate $\tp_{Y|X}$, a small constant $\delta > 0$}
\KwOut{the fitted classifier $\hphi$ or report the GNPMC problem as infeasible}
Randomly divide the whole training data (and reindex them) into $\mathcal{D}_1 \bigcup \mathcal{D}_2 = \{(\bm{x}_i, y_i)\}_{i = 1}^{n} \bigcup \{(\bm{x}_i, y_i)\}_{i = n+1}^{2n}$\\
$\widehat{\tp}_{Y|X}, \hbpi \leftarrow$ the estimates of $\tp_{Y|X}$ (through $\mathcal{M}$) and $\bpi^*$ on $\mathcal{D}_{2} = \{(\bm{x}_i, y_i)\}_{i = n+1}^{2n}$\\
$\hblambda \leftarrow \argmax_{\blambda \in \mathbb{R}_{+}^{|\mathcal{A}|}, \infnorm{\blambda} \leq R} \widehat{G}^{\textup{ER}}(\blambda; \widehat{\tp}_{Y|X}, \hbpi)$, where $\widehat{G}^{\textup{ER}}$ is estimated on $\mathcal{D}_1 = \{(\bm{x}_i, y_i)\}_{i = 1}^{n}$\\
\uIf{$\widehat{G}^{\textup{ER}}(\hblambda) \leq 1+ \delta$}{
Report the GNPMC problem as feasible and output the solution $\hphi_{\blambda}(\bx) = \argmin_{r=1:K} \{\sum_{k \neq r}c_{kr}(\blambda, \hbpi)\widehat{\tp}_{Y|X=\bx}(Y = k)\}$\\
}
\Else{Report the GNPMC problem as infeasible}
\end{algorithm}

\subsection{Theory on GNPMC-CX and GNPMC-ER}
In this section, we extend the theoretical analysis in Section \ref{sec: theory} for the NPMC problem \eqref{eq: NPMC} to the case of GNPMC problem \eqref{eq: G-NPMC}. Some assumptions we made in Section \ref{sec: theory} (for example, Assumptions \ref{asmp: posterior distance} and \ref{asmp: rademacher}) are still necessary for the GNPMC problem \eqref{eq: G-NPMC}. The others, like Assumptions \ref{asmp: strong duality}, \ref{asmp: second order info}, and \ref{asmp: marginal finite sample}, may be subject to slight changes for the GNPMC problem.

\subsubsection{Checking the strong duality and feasibility}
Similar to the NPMC case, strong duality is the bridge between the GNPMC problem \eqref{eq: G-NPMC} and the cost-sensitive learning problem \eqref{eq: general G cs problem}.
\begin{assumption}[Strong duality for the GNPMC problem]\label{asmp: strong duality gnpmc}
	Suppose it holds that
	\begin{equation}
		\inf_{\phi \in \mathfrak{C}}J(\phi) = \sup_{\blambda \in \mathbb{R}_{+}^{|\mathcal{A}|}}G(\blambda),
	\end{equation}
	where $\mathfrak{C}$ includes all feasible classifiers for the GNPMC problem \eqref{eq: G-NPMC}. If $\mathfrak{C} \neq \emptyset$, the infimum over $\phi \in \mathfrak{C}$ is achivable, and the supremum over $\blambda \in \mathbb{R}_{+}^{|\mathcal{A}|}$ can be achieved at a finite $\blambda$.
\end{assumption}


We have a sufficient and necessary characterization for strong duality as below.
\begin{theorem}[Sufficient and necessary conditions for GNPMC strong duality]\label{thm: strong duality gnpmc}
 Suppose $\{X|Y=k\}_{k=1}^K$ are continuous random variables (i.e., have Lebesgue density). 
\begin{enumerate}[(i)]
	\item When the GNPMC problem \eqref{eq: G-NPMC} is feasible, the strong duality holds if and only if there exists $\blambda = \{\lambda_{kr}\}_{(k,r) \in \mathcal{A}}$ such that $\phi_{\blambda}^*$ is feasible for the NP problem, i.e., $\tp_{X|Y=k}(\phi_{\blambda}^*(X) = r) \leq \alpha_{kr}$ for all $(k,r) \in \mathcal{A}$.
	\item Suppose $\tp_{Y|X=\bx}(Y=r) \geq a > 0$ for a.s. $\bx$ (w.r.t. the distribution of $X$) for all $r \in \{r: (k,r)\in \mathcal{A}\}$. When the GNPMC problem \eqref{eq: G-NPMC} is infeasible, the strong duality holds (i.e., $\sup_{\blambda \in \mathbb{R}_{+}^{|\mathcal{A}|}}G(\blambda)$ is unbounded from above) if and only if for an arbitrary $\blambda \in \mathbb{R}^{|\mathcal{A}|}_+$, $\phi_{\blambda}^*$ is infeasible for GNPMC problem \eqref{eq: G-NPMC}, i.e., $\exists$ at least one pair of $(k,r) \in \mathcal{A}$ such that $\tp_{X|Y=k}(\phi_{\blambda}^*(X) = r) > \alpha_{kr}$.
\end{enumerate}
\end{theorem}

Similar to Corollary \ref{cor: df check}, we have the following corollary which connects the strong duality and feasibility of GNPMC problem \eqref{eq: G-NPMC} to $\argmax_{\blambda \in \mathbb{R}_{+}^{|\mathcal{A}|}}G(\blambda)$ and $G(\blambda^*)$. The proof is almost the same as the proof of Corollary \ref{cor: df check}, so we omit it.

\begin{corollary}\label{cor: df check gnpmc}
	Suppose $\{X|Y=k\}_{k=1}^K$ are continuous random variables (i.e. have Lebesgue density).  We have the following equivalence.
	\begin{enumerate}[(i)]
		\item GNPMC problem is feasibile, strong duality holds $\Leftrightarrow$ $\exists$ a finite $\blambda^* \in \argmax_{\blambda \in \mathbb{R}_{+}^{|\mathcal{A}|}}G(\blambda)$ and $\phi^*_{\blambda^*}$ is feasible;
		\item GNPMC problem is infeasibile, strong duality holds $\Leftrightarrow$ $\exists$ an infinite $\blambda^*$ and $G(\blambda^*) = +\infty$;
		\item \begin{itemize}
				\item GNPMC problem is feasible, strong duality fails $\Rightarrow$ For any $\blambda^* \in \argmax_{\blambda \in \mathbb{R}_{+}^{|\mathcal{A}|}}G(\blambda)$, $\lambda^*$ is infinite \footnote{When we say infinite $\blambda^*$, we refer to a sequence $\{(\blambda^*)^{(m)}\}_{m=1}^{\infty}$ s.t. $\infnorm{(\blambda^*)^{(m)}}\rightarrow +\infty$, $\lim_{m \rightarrow \infty} G((\blambda^*)^{(m)}) = \sup_{\blambda \in \mathbb{R}_+^{|\mathcal{A}|}}G(\blambda)$ exists and is denoted as $G(\blambda^*)$.} or $\blambda^*$ is finite but $\phi^*_{\blambda^*}$ is infeasible, and 	$G(\blambda^*) \leq 1$;
				\item For any $\blambda^* \in \argmax_{\blambda \in \mathbb{R}_{+}^{|\mathcal{A}|}}G(\blambda)$, $\lambda^*$ is infinite or $\blambda^*$ is finite but $\phi^*_{\blambda^*}$ is infeasible, and $G(\blambda^*) \leq 1 \Rightarrow$ strong duality fails, GNPMC problem can be either feasible or infeasible;
			 \end{itemize}
		\item \begin{itemize}
 				\item GNPMC problem is infeasible, strong duality fails $\Rightarrow$ For any $\blambda^* \in \argmax_{\blambda \in \mathbb{R}_{+}^{|\mathcal{A}|}}G(\blambda)$, $\blambda^*$ is infinite or $\blambda^*$ is finite but $\phi^*_{\blambda^*}$ is infeasible, and $G(\blambda^*) < +\infty$.
 				\item For any $\blambda^* \in \argmax_{\blambda \in \mathbb{R}_{+}^{|\mathcal{A}|}}G(\blambda)$, $\blambda^*$ is infinite or $\blambda^*$ is finite but $\phi^*_{\blambda^*}$ is infeasible, and $1 < G(\blambda^*) < +\infty \Rightarrow$ strong duality fails, GNPMC problem is infeasible
 			  \end{itemize}
	\end{enumerate}
\end{corollary}

We can replace $\blambda^*$ and $G(\blambda^*)$ in Corollary \eqref{cor: df check gnpmc} with the their estimates $\hblambda$ and $\widehat{G}(\hblambda)$ from GNPMC-CX or GNPMC-ER and obtain two algorithms to check strong duality and feasibility in practice. We summarize them in Algorithms \ref{algo: df check CX gnpmc} and \ref{algo: df check ER gnpmc}, which are GNPMC counterparts of Algorithms \ref{algo: df check CX} and \ref{algo: df check ER}.

\begin{algorithm}[!h]
\setstretch{1.2}
\caption{Test strong duality and feasibility with GNPMC-CX}
\label{algo: df check CX gnpmc}
\KwIn{training data $\{(\bm{x}_i, y_i)\}_{i = 1}^n$, target upper bounds of errors $\balpha$, the weighting vector of objective function $\bw$, a function class $\mathcal{M}$ to estimate $\tp_{Y|X}$, a small constant $\delta \in (0, 1)$}
\KwOut{strong duality indicator $\hat{s}$ and feasibility indicator $\hat{f}$}
A finite $\hblambda$ and a corresponding CS classifier $\hphi$, or $\perp$ (we obtain a sequence $\{\hblambda^{(m)}\}_{m=1}^{\infty}$ with $\infnorm{\hblambda^{(m)}} \rightarrow +\infty$ and $\widehat{G}^{\textup{CX}}(\hblambda^{(m)}) \rightarrow +\infty$\footnotemark) $\leftarrow$ Steps 1, 3 of GNPMC-CX with $\mathcal{M}$ \\
\uIf{we obtain finite $\hblambda$ and a corresponding CS classifier $\hphi$}{
	Check if $\widehat{R}_{kr}(\hphi) \leq \alpha_{kr}(1+\delta)$ for all $(k, r) \in \mathcal{A}$ and $\widehat{G}^{\textup{CX}}(\hblambda) \leq 1+\delta$, where $\widehat{R}_{kr}(\hphi)$ is the empirical classification error calculated on $\{(\bm{x}_i, y_i)\}_{i = 1}^n$ \\
	\uIf{Yes}{
		Set $\hat{s} = 1$, $\hat{f} = 1$
	}\uElseIf{$\widehat{G}^{\textup{CX}}(\hblambda) \leq 1+\delta$}{
		Set $\hat{s} = 0$, $\hat{f} = 1$
	}\Else{Set $\hat{s} = 0$, $\hat{f} = 0$}
}\ElseIf{$\perp$}{
	Set $\hat{s} = 1$, $\hat{f} = 0$
}
Output $\hat{s}$ and $\hat{f}$
\end{algorithm}
\footnotetext{Note that $\widehat{G}^{\textup{CX}}(\blambda)$ is piecewise-linear in $\blambda$ with finite non-differentiable points, therefore $\sup_{\blambda \subseteq \mathbb{R}^{|\mathcal{A}|}_+}\widehat{G}^{\textup{CX}}(\blambda)$ is either finite and achievable at some finite $\hblambda$ or infinite and approached by a sequence $\{\hblambda^{(m)}\}_{m=1}^{\infty}$ with $\infnorm{\hblambda^{(m)}} \rightarrow +\infty$.}

\begin{algorithm}[!h]
\setstretch{1.2}
\caption{Test strong duality and feasibility with GNPMC-ER}
\label{algo: df check ER gnpmc}
\KwIn{training data $\{(\bm{x}_i, y_i)\}_{i = 1}^n$, target upper bounds of errors $\balpha$, the weighting vector of objective function $\bw$, a function class $\mathcal{M}$ to estimate $\tp_{Y|X}$, a small constant $\delta \in (0, 1)$, a search range $R > 0$, a threshold $R_G > 1+\delta$}
\KwOut{strong duality indicator $\hat{s}$ and feasibility indicator $\hat{f}$}
$\hblambda$, $\hphi \leftarrow$ Steps 1-3 of GNPMC-ER with $\mathcal{M}$ and search range $R$\\
\uIf{we obtain a finite $\hblambda$ and a corresponding CS classifier $\hphi$}{
	Check if $\widehat{R}_{kr}(\hphi) \leq \alpha_{kr}(1+\delta)$ for all $(k, r) \in \mathcal{A}$ and $\widehat{G}^{\textup{ER}}(\hblambda) \leq 1+\delta$, where $\widehat{R}_{kr}(\hphi)$ is the empirical classification error calculated on the splitted data set $\mathcal{D}_1$ in NPMC-ER\\
	\uIf{Yes}{
		Set $\hat{s} = 1$, $\hat{f} = 1$
	}\uElseIf{$\widehat{G}^{\textup{ER}}(\hblambda) \leq 1+\delta$}{
		Set $\hat{s} = 0$, $\hat{f} = 1$
	}\uElseIf{$\widehat{G}^{\textup{ER}}(\hblambda) > R_G$}{
		Set $\hat{s} = 1$, $\hat{f} = 0$
	}\Else{
		Set $\hat{s} = 0$, $\hat{f} = 0$
	}
}
Output $\hat{s}$ and $\hat{f}$
\end{algorithm}

We have the following result, which serves as a theoretical guarantee of Algorithms \ref{algo: df check CX gnpmc} and \ref{algo: df check ER gnpmc}. The proof is the same as the proof of Theorem \ref{thm: dk check}, so we omit it.
 
\begin{theorem}\label{thm: dk check gnpmc}
	Suppose $\{X|Y=k\}_{k=1}^K$ are continuous random variables (i.e. have Lebesgue density). Suppose Assumptions \ref{asmp: posterior distance}-\ref{asmp: marginal finite sample} (for Algorithm \ref{algo: df check CX}) or Assumptions \ref{asmp: posterior distance}, \ref{asmp: second order info}, \ref{asmp: marginal finite sample} (for Algorithm \ref{algo: df check ER}) hold. Then, we have the following results for Algorithms \ref{algo: df check CX} and \ref{algo: df check ER}.
	\begin{enumerate}[(i)]
		\item If the NPMC problem is feasible and strong duality holds, then
			\begin{equation}
				\tp(\hat{f} = 1, \hat{s} = 1) \geq 1-C\exp\{-C'n\delta^{4/\bar{\gamma}}\} -C\delta^{\frac{-2\vee (1+\bar{\gamma})}{\bar{\gamma}}}\max_k \te|\widehat{\tp}_{Y|X}(Y=k) - \tp_{Y|X}(Y=k)|.
			\end{equation}
		\item Suppose the NPMC problem is feasible and strong duality fails.
			\begin{enumerate}
				\item For Algorithm \ref{algo: df check CX}, if $\blambda^*$ is finite and $R_{kr}(\phi^*_{\blambda^*}) \leq \alpha_{kr}(1+\delta)$, then with probability at least $1-C\exp\{-C'n\delta^{4/\bar{\gamma}}\} -C\delta^{\frac{-2\vee (1+\bar{\gamma})}{\bar{\gamma}}}\max_k \te|\widehat{\tp}_{Y|X}(Y=k) - \tp_{Y|X}(Y=k)|$, we will obtain $\hblambda$ and $\hphi$ from Step 1, and $R_{kr}(\hphi) \leq \alpha_{kr}(1+\frac{3}{2}\delta)$ for all $(k, r) \in \mathcal{A}$;
				\item For Algorithm \ref{algo: df check ER}, if $\infnorm{\blambda^*}\leq R$ and $R_{kr}(\phi^*_{\blambda^*}) \leq \alpha_{kr}(1+\delta)$, then with probability at least $1-C\exp\{-C'n\delta^{4/\bar{\gamma}}\} -C\delta^{\frac{-2\vee (1+\bar{\gamma})}{\bar{\gamma}}}\max_k \te|\widehat{\tp}_{Y|X}(Y=k) - \tp_{Y|X}(Y=k)|$, we will obtain $\hblambda$ and $\hphi$ from Step 1, and $R_{kr}(\hphi) \leq \alpha_{kr}(1+\frac{3}{2}\delta)$ for all $(k, r) \in \mathcal{A}$;
				\item If $\blambda^*$ is finite and $R_k(\phi^*_{\blambda^*}) > \alpha_k(1+\frac{3}{2}\delta)$, then
					\begin{equation}
						\tp(\hat{f} = 1, \hat{s} = 0) \geq 1-C\exp\{-C'n\delta^{4/\bar{\gamma}}\} -C\delta^{\frac{-2\vee (1+\bar{\gamma})}{\bar{\gamma}}}\max_k \te|\widehat{\tp}_{Y|X}(Y=k) - \tp_{Y|X}(Y=k)|.
					\end{equation} 
			\end{enumerate}
		\item If the NPMC problem is infeasible and strong duality holds, then
			\begin{equation}
				\tp(\hat{f} = 0, \hat{s} = 1) \geq 1-C\exp\{-C'n\} -C\max_k \te|\widehat{\tp}_{Y|X}(Y=k) - \tp_{Y|X}(Y=k)|.
			\end{equation}
		\item If the NPMC problem is infeasible and strong duality fails, $\blambda^*$ is finite, and $G(\phi^*_{\blambda^*}) > 1+\frac{3}{2}\delta$, then
			\begin{equation}
				\tp(\hat{f} = 0, \hat{s} = 0) \geq 1-C\exp\{-C'n\delta^{2}\} - C\delta^{-1}\max_{k}\te |\widehat{\tp}_{Y|X}(Y = k) - \tp_{Y|X}(Y = k)|.
			\end{equation}
	\end{enumerate}
\end{theorem}

\subsubsection{Analysis on GNPMC-CX}

The following two assumptions are adapted from Assumptions \ref{asmp: second order info} and \ref{asmp: marginal finite sample} for the GNPMC problem.
\begin{assumption}[Strict concavity]\label{asmp: second order info gnpmc}
	$G(\blambda)$ is continuously twice-differentiable at $\blambda^*$ and $\nabla^2 G(\blambda^*) \prec 0$, where $\blambda^* = \argmax G(\blambda)$ is unique.
\end{assumption}

\begin{assumption}[Margin condition]\label{asmp: marginal finite sample gnpmc}
	Denote the function characterizing the decision boundary of class $r$ as $\varphi_r(\bx) = \sum_{k\neq r}c_{kr}(\blambda^*, \bpi^*)\tp_{Y|X=\bx}(Y=k) -\min_{j \neq r}\{\sum_{k\neq j} c_{kj}(\blambda^*, \bpi^*) \tp_{Y|X=\bx}(Y=k)\}$, where $\blambda^* = \argmax G(\blambda)$ is unique. It holds
	\begin{equation}
		\max_{k \neq r \in [K]}\tp_{X|Y=k}(\norm{\varphi_k(X)} \leq t) \lesssim t^{\bar{\gamma}},
	\end{equation}
	with some $\bar{\gamma} > 0$ and a non-negative $t$ smaller than some constant $C \in (0, 1)$.
\end{assumption}

With these conditions, we can show that the NP oracle properties hold for GNPMC-CX.
\begin{theorem}[Multi-class NP oracle properties of GNPMC-CX]\label{thm: error rate bound gnpmc}
GNPMC-CX satisfies the multi-class NP oracle properties in the following senses.
	\begin{enumerate}[(i)]
		\item When the GNPMC problem \eqref{eq: G-NPMC} is feasible, if Assumptions \ref{asmp: posterior distance}, \ref{asmp: rademacher}, \ref{asmp: strong duality gnpmc}, \ref{asmp: second order info gnpmc}, and \ref{asmp: marginal finite sample gnpmc} hold, then there exist a solution $\phi^*$ and a constant $C >0$ such that
			\begin{align*}
				& \max_{k,r: k\neq r}\tp(\norm{R_{kr}(\hphi) - R_{kr}(\phi^*)} > \tau) \\\lesssim &  \exp\{-Cn\tau^{4/\bar{\gamma}}\} + \tau^{-\frac{2\vee (1+\bar{\gamma})}{\bar{\gamma}}}\max_k \te\norma{\widehat{\tp}_{Y|X}(Y=k)-\tp_{Y|X}(Y=k)}, 
			\end{align*}
		when $1 \geq \tau \gtrsim \big[C_{\textup{Rad}}(n) + \max_k \te\norm{\widehat{\tp}_{Y|X}(Y=k)-\tp_{Y|X}(Y=k)}\big]^{\bar{\gamma}/2}$.
		\item When the GNPMC problem \eqref{eq: G-NPMC} is infeasible, if Assumptions \ref{asmp: posterior distance}, \ref{asmp: rademacher}, and \ref{asmp: strong duality gnpmc} hold, then there exists a constant $C > 0$ such that
			\begin{equation}
				\tp\left(\norm{\widehat{G}^{\textup{CX}}(\hblambda)} \leq 1+\delta\right) \lesssim \exp\{-Cn\} +\max_k \te\norma{\widehat{\tp}_{Y|X}(Y=k)-\tp_{Y|X}(Y=k)}.
			\end{equation}
	\end{enumerate}
\end{theorem}

\begin{remark}\label{rmk: oracle gnpmc-cx}
	Similar to Remark \ref{rmk: oracle npmc-cx}, when the GNPMC problem \eqref{eq: G-NPMC} is feasible,
	\begin{align}
		R_{kr}(\hphi) -  \alpha_{kr} &\leq R_{kr}(\hphi) -  R_{kr}(\phi^*) \leq \mathcal{O}_{\tp}(\epsilon(n)), \quad \forall (k,r) \in \mathcal{A}, \\
		J(\hphi) - J(\phi^*) &\leq \mathcal{O}_{\tp}(\epsilon(n)),
	\end{align}
	where $\epsilon(n) = n^{-\bar{\gamma}/4} + \left(\max_k \te\norm{\widehat{\tp}_{Y|X}(Y=k)-\tp_{Y|X}(Y=k)}\right)^{-\bar{\gamma}/(2\vee (1+\bar{\gamma}))} + \sqrt{C_{\textup{Rad}}(n)} \rightarrow 0$.
	Hence, Theorem \ref{thm: error rate bound gnpmc} verifies that GNPMC-CX satisfies the multi-class NP oracle properties. 
\end{remark}


Under certain conditions, we can also show that strong consistency holds for GNPMC-CX. We require the function class used to estimate $\tp_{Y|X}$ to be indexed by a parameter $\bbeta \in \mathcal{B}$, where $\mathcal{B}$ is a compact set in a finite-dimensional metric space.

\begin{theorem}[Strong consistency of GNPMC-CX]\label{thm: consistency gnpmc}
	GNPMC-CX satisfies strong consistency in the following senses.
	\begin{enumerate}[(i)]
		\item When the GNPMC problem \eqref{eq: G-NPMC} is feasible, if Assumptions \ref{asmp: posterior distance as}, \ref{asmp: error rate continuity}, \ref{asmp: strong duality gnpmc}, \ref{asmp: second order info gnpmc} hold, then there exists a solution $\phi^*$, such that $\lim_{n\rightarrow \infty}R_{kr}(\hphi) = R_{kr}(\phi^*)$ a.s. for all $k$ and $r \in \{1,\ldots,K\}$.
		\item When the GNPMC problem \eqref{eq: G-NPMC} is infeasible, if Assumptions  \ref{asmp: posterior distance as}, \ref{asmp: error rate continuity}, and \ref{asmp: strong duality gnpmc} hold, then for any $M > 0$, $\liminf_{n \rightarrow \infty}\widehat{G}^{\textup{CX}}(\hblambda) > M$ a.s..
	\end{enumerate}
\end{theorem}

\subsubsection{Analysis on GNPMC-ER}
Under certain conditions, we can verify the NP oracle properties for GNPMC-ER. Similar to NPMC-ER, GNPMC-ER does not require a Rademacher class (Assumption \ref{asmp: error rate continuity}) to satisfy NP oracle properties.
\begin{theorem}[Multi-class NP oracle properties of GNPMC-ER]\label{thm: error bound alg 2 gnpmc}
	GNPMC-ER satisfies the multi-class NP oracle properties in the following senses.
	\begin{enumerate}[(i)]
		\item When the GNPMC problem \eqref{eq: G-NPMC} is feasible, if Assumptions \ref{asmp: posterior distance}, \ref{asmp: strong duality gnpmc}, \ref{asmp: second order info gnpmc}, and \ref{asmp: marginal finite sample gnpmc} hold, and $R \geq \infnorm{\blambda^*}$ with $\blambda^* = \argmax_{\blambda \in \mathbb{R}^{|\mathcal{A}|}_+} G(\blambda)$, then there exist a solution $\phi^*$ and some constants $C, C' > 0$ such that
			\begin{align}
				&\max_{k,r: k\neq r}\tp(\norm{R_{kr}(\hphi) - R_{kr}(\phi^*)} > \tau) \\  &\lesssim   \exp\{-Cn\tau^{4/\bar{\gamma}}\} + \tau^{-\frac{2\vee (1+\bar{\gamma})}{\bar{\gamma}}}\max_k \te\norma{\widehat{\tp}_{Y|X}(Y=k)-\tp_{Y|X}(Y=k)},
			\end{align}
		when $1 \geq \tau \gtrsim n^{-\bar{\gamma}/4}$.
		\item When the GNPMC problem \eqref{eq: G-NPMC} is infeasible, if Assumptions \ref{asmp: posterior distance} and \ref{asmp: strong duality gnpmc} hold, and $R$ satisfies $\sup_{\infnorm{\blambda} \leq R}G(\blambda) > 1 + \delta$, then there exists a constant $C > 0$ such that
			\begin{equation}
				\tp\left(\norm{\widehat{G}^{\textup{ER}}(\hblambda)} \leq 1\right) \lesssim \exp\{-Cn\} +\max_k \te\norma{\widehat{\tp}_{Y|X}(Y=k)-\tp_{Y|X}(Y=k)}.
			\end{equation}
	\end{enumerate}
\end{theorem}

Analyzing in the same way as in Remark \ref{rmk: oracle gnpmc-cx}, we know that Theorem \ref{thm: error bound alg 2 gnpmc} verifies multi-class NP oracle properties of GNPMC-ER.

\section{Extension to Worst-case Objective Functions}\label{sec: NPMC-max supp}
In Section \ref{subsec: np}, we delineated the NPMC problem as follows: 
\begin{align}
	&\min_{\phi} \quad J(\phi) = \sum_{k=1}^K w_k \tp_{X|Y=k}(\phi(X)\neq k) \\
	&\text{s.t.} \quad \tp_{X|Y=k}(\phi(X)\neq k) \leq \alpha_k, \quad k \in \mathcal{A}, \label{eq: NPMC recap}
\end{align}
where $\phi: \mathcal{X} \rightarrow \{1, \ldots, K\}$ is a classifier, $\alpha_k \in [0, 1)$, $w_k \geq 0$ and $\mathcal{A} \subseteq \{1, \ldots, K\}$. In certain scenarios, rather than minimizing the weighted misclassification errors across different classes as the objective function, we may want to minimize the worst-case error rate among classes not in $\mathcal{A}$ for the sake of fairness. This motive leads us to consider the following NPMC-max problem:
\begin{align}
	&\min_{\phi} \quad \max_{k \notin \mathcal{A}} \tp_{X|Y=k}(\phi(X)\neq k) \\
	&\text{s.t.} \quad \tp_{X|Y=k}(\phi(X)\neq k) \leq \alpha_k, \quad k \in \mathcal{A}, \label{eq: NPMC max}
\end{align}
with $\alpha_k \in [0, 1)$, $w_k \geq 0$ and $\mathcal{A} \subseteq \{1, \ldots, K\}$. 

In the following, we will elucidate how the heuristics for the NPMC problem \eqref{eq: NPMC recap} in Section \ref{sec: method} can be extended to address the NPMC-max problem \eqref{eq: NPMC max}. 

First, for convenience, let us rewrite problem \eqref{eq: NPMC max} as
\begin{align}
	&\min_{\phi, t \in [0, 1]} \quad t  \\
	&\text{s.t.} \quad \tp_{X|Y=k}(\phi(X)\neq k) \leq t, \quad\,\,\,\, k \notin \mathcal{A}\\
	&\quad\quad\,\, \tp_{X|Y=k}(\phi(X)\neq k) \leq \alpha_k, \quad k \in \mathcal{A}. \label{eq: NPMC max rewritten}
\end{align}
Then we have the Lagrangian function
\begin{align}
	L(\blambda, \phi, t) &= t + \sum_{k \notin \mathcal{A}}\lambda_k[1-\tp_{X|Y=k}(\phi(X) = k) - t] + \sum_{k \in \mathcal{A}}\lambda_k[1-\tp_{X|Y=k}(\phi(X) = k) - \alpha_k]\\
	&= -\sum_{k=1}^K \lambda_k \tp_{X|Y=k}(\phi(X) = k) + \bigg(1-\sum_{k \notin \mathcal{A}}\lambda_k\bigg)t + \sum_{k \notin \mathcal{A}}\lambda_k + \sum_{k \in \mathcal{A}}\lambda_k(1-\alpha_k).
\end{align}
The dual problem of \eqref{eq: NPMC max rewritten} is
\begin{equation}
	\max_{\blambda \in \mathbb{R}_+^{|\mathcal{A}|}} \min_{\phi, t \in [0, 1]} L(\blambda, \phi, t).
\end{equation}
Note that the first term in $F_{\blambda}(\phi, t)$ can be reformulated as 
\begin{align}
	\sum_{k=1}^K \lambda_k \tp_{X|Y=k}(\phi(X) = k) &= \sum_{k=1}^K \frac{\lambda_k}{\pi_k^*}\cdot \pi_k^* \tp_{X|Y=k}(\phi(X) = k) \\
	&= \te_X\bigg[\frac{\lambda_{\phi(X)}}{\pi_{\phi(X)}}\tp_{Y=k|X}(Y = \phi(X)) \bigg].
\end{align}
Then given any $\blambda = \{\lambda_k\}_{k=1}^K$ and $t \in [0, 1]$, we have the following CS classifier minimizing the cost function $F_{\blambda}(\phi, t)$:
\begin{equation}\label{eq: cs classifier npmc-max}
	\phi^*_{\blambda}: \bx \mapsto \argmax_{k=1:K}\bigg\{\frac{\lambda_k}{\pi_k^*}\tp_{Y|X = \bx}(Y = k)\bigg\} \in \argmin_{\phi}L(\blambda, \phi, t).
\end{equation}
Given $\blambda$ and $\phi^*_{\blambda}$, we have
\begin{equation}
	L(\blambda, \phi_{\blambda}^*, t) = \underbrace{-\sum_{k=1}^K \lambda_k \tp_{X|Y=k}(\phi(X) = k) + \sum_{k \notin \mathcal{A}}\lambda_k + \sum_{k \in \mathcal{A}}\lambda_k(1-\alpha_k)}_{\textup{unrelated to }t} + \bigg(1-\sum_{k \notin \mathcal{A}}\lambda_k\bigg)t.
\end{equation}
Therefore,
\begin{equation}
	t_{\blambda}^* = \begin{cases}
		1, \quad &\textup{if } 1-\sum_{k \notin \mathcal{A}}\lambda_k < 0; \\
		0, \quad &\textup{if } 1-\sum_{k \notin \mathcal{A}}\lambda_k > 0; \\
		\textup{free in }[0, 1], \quad &\textup{if } 1-\sum_{k \notin \mathcal{A}}\lambda_k = 0;
	\end{cases} \quad \in \argmin_{t \in [0, 1]} L(\blambda, \phi_{\blambda}^*, t). 
\end{equation}
Hence we can define
\begin{align}
	G(\blambda) &\coloneqq L(\blambda, \phi_{\blambda}^*, t^*_{\blambda}) \\
	&= \begin{cases}
		-\sum_{k=1}^K \lambda_k \tp_{X|Y = k}(\phi^*_{\blambda}(X) = k) + \sum_{k \in \mathcal{A}} \lambda_k(1-\alpha_k) + 1, \hspace{1.45cm} \textup{if } 1-\sum_{k \notin \mathcal{A}}\lambda_k < 0; \\
		-\sum_{k=1}^K \lambda_k \tp_{X|Y = k}(\phi^*_{\blambda}(X) = k) + \sum_{k \notin \mathcal{A}} \lambda_k + \sum_{k \in \mathcal{A}} \lambda_k(1-\alpha_k), \,\, \textup{if } 1-\sum_{k \notin \mathcal{A}}\lambda_k \geq 0.
	\end{cases} \label{eq: G npmc-max}
\end{align}
Similar to the strong duality defined in Assumption \ref{asmp: strong duality} for NPMC problem \eqref{eq: NPMC}, we can define the following strong duality for NPMC-max problem.

\begin{assumption}[Strong duality for NPMC-max problem]\label{asmp: strong duality npmc-max}
	It holds that
	\begin{equation}
		\inf_{\phi \in \mathfrak{C}}J(\phi) = \sup_{\blambda \in \mathbb{R}_{+}^{|\mathcal{A}|}}G(\blambda),
	\end{equation}
	where $\mathfrak{C}$ includes all feasible classifiers for the NPMC-max problem \eqref{eq: NPMC max}. If $\mathfrak{C} \neq \emptyset$, the infimum over $\phi \in \mathfrak{C}$ is achievable, and the supremum over $\blambda \in \mathbb{R}_{+}^{K}$ can be achieved at a finite $\blambda$.
\end{assumption}

Similar to the analysis for NPMC problem \eqref{eq: NPMC}, if NPMC-max problem \eqref{eq: NPMC max} (or equivalently, \eqref{eq: NPMC max rewritten}) is feasible and strong duality holds, then on the population level, we can find $\blambda$ that maximizes $G(\blambda)$. Subsequently, we can substitute it into \eqref{eq: cs classifier npmc-max} to obtain the CS classifier, which also serves as the solution to the NPMC-max problem. Conversely, if the NPMC-max problem is infeasible and strong duality holds, then we must have $\sup_{\blambda \in \mathbb{R}_+^K}G(\blambda) = +\infty$.

Building upon these heuristic insights, algorithms akin to NPMC-CX and NPMC-ER can be proposed, and their theoretical foundations can be explored. Given the extensive coverage in this paper, we defer the detailed exposition to future research endeavors. We hope the idea of utilizing duality between NP problems and CS problems can be helpful for researchers working on NP problems and pave the way for a novel series of NP methodologies.

Before concluding this section, it is worth mentioning that the formula of $G(\blambda)$ can be further simplified. The current formulation \eqref{eq: G npmc-max} contains two pieces, and it might be hard to find its maximizer. However, it can be shown that it suffices to seek the maximizer under the constraint $\sum_{k \notin \mathcal{A}}\lambda_k = 1$. Under this constraint, $G(\blambda)$ simplifies to $-\sum_{k=1}^K \lambda_k \tp_{X|Y = k}(\phi^*_{\blambda}(X) = k) + \sum_{k \notin \mathcal{A}} \lambda_k + \sum_{k \in \mathcal{A}} \lambda_k(1-\alpha_k)$.

\begin{lemma}\label{lem: npmc-max}
	Suppose Assumption \ref{asmp: strong duality npmc-max} holds. 
	\begin{enumerate}[(i)]
		\item If NPMC-max problem is feasible, then there exists optimal $\blambda^* = \{\blambda_k^*\}_{k=1}^K \in \argmax_{\blambda \in \mathbb{R}_+^K}G(\blambda)$ satisfying $\sum_{k \notin \mathcal{A}}\lambda_k = 1$.
		\item If NPMC-max problem is infeasible, then there exists a sequence $\{\blambda^{(m)}\}_{m=1}^{\infty}$ with $\blambda^{(m)} = \{\blambda_k^{(m)}\}_{k=1}^K \in \mathbb{R}_+^K$ and $\sum_{k \notin \mathcal{A}} \lambda_k^{(m)} = 1$ such that $G(\blambda^{(m)}) \rightarrow +\infty$ as $m \rightarrow \infty$.
	\end{enumerate}
\end{lemma}

\section{Additional Numerical Results}\label{sec: additional numerical results supp}

\subsection{More implementation details}\label{subsec: implementation supp}

The optimization procedure in step 3 of Algorithms \ref{algo} and \ref{algo: erm} to find $\hblambda$ is implemented through the function \texttt{hjkb} in the package \texttt{dfoptim}, which solves derivative-free optimization problems by Hooke-Jeeves algorithm \citepapp{hooke1961direct, kelley1999iterative}. For the error control tolerance parameter $\delta$ in NPMC-CX and NPMC-ER, we set $\delta = 0.1$ for NPMC-CX and $\delta = 0.2$ for NPMC-ER, respectively. Regarding the algorithms for checking the strong duality and feasibility, we take $\delta = 0.1$ in Algorithm \ref{algo: df check CX} and \ref{algo: df check CX gnpmc}, while we take $\delta = 0.2$ and $R_G = 10$ in Algorithm \ref{algo: df check ER} and \ref{algo: df check ER gnpmc}. A larger $\delta$ is used for ER methods by considering the loss of sample-size efficiency by data splitting. We set the search range $R = 1000$ in NPMC-ER, GNPMC-ER, Algorithm \ref{algo: df check ER}, and Algorithm \ref{algo: df check ER gnpmc}.

Various packages are used to fit different classification methods and produce estimates for posterior probability $\tp_{Y|X}$. These methods include logistic regression (logistic, package \texttt{nnet}), linear discriminant analysis (LDA, package \texttt{MASS}), $k$-nearest neighbors ($k$NN, package \texttt{caret}), non-parametric na\"ive Bayes classifier with Gaussian kernel (NNB, package \texttt{naivebayes}), support vector machines with RBF kernel (SVM, package \texttt{e1071}), and random forest (RF, package \texttt{randomForest}), where the corresponding abbreviations and packages are indicated in the parentheses. For $k$NN, the number of nearest neighbors is set to $k = \lfloor \sqrt{n/K} \rfloor$, where $n$ is the training sample size, and $K$ is the number of classes. For NNB, the kernel bandwidth is selected based on Silverman's rule of thumb \citepapp{silverman2018density}. All parameters for SVM are set to default values as suggested in package \texttt{e1071}. For instance, the RBF kernel has the form $\exp\{-\gamma|u-v|\}$ with $\gamma = 1/p$ where $p$ is the number of variables in the data. Moreover, the constant of the regularization term in the Lagrange formulation is set to be $1$. More details can be found in \citeapp{david1071}. For RF, all parameters are set to default values as suggested in package \texttt{randomForest}. For example, the number of trees is set to be $500$, and the number of variables randomly sampled as candidates at each split is set to be $\lfloor \sqrt{p} \rfloor$ where $p$ is the number of variables in the data. More details are available in \citeapp{liaw2002classification}.
 
\subsection{Simulations}
This section collects simulation results in addition to the results in Section \ref{subsec: simulations} of the main text (denoted as Case 1). Case 2 is a 4-class NPMC problem. Case 3 is a variant setting of Case 1 where the marginal probabilities of three classes are very imbalanced. Case 4 follows the same setting as Case 1 but tackles a GNPMC problem with confusion matrix control.

\subsubsection{Case 1}\label{subsubsec: case 1 simulation supp}
In this subsection, we provide additional details for the simulation case 1. 

As discussed in Section \ref{subsec: np}, \citeapp{landgrebe2005neyman} proposed a general empirical apporach to solve the NPMC problem based on the multi-class ROC estimation. The method looks for the cost-sensitive learning classifiers with different weights and aims to identify the feasible cost-sensitive learning classifier that minimizes the objective function. The main idea of their method aligns with ours for the NPMC problem. However, they rely on grid search across all classes to determine the optimal weight vector, which becomes computationally impractical for more than four classes.  Additionally, their method uses only one weight for each class, which distinguishes our method from theirs in the GNPMC problem. As demonstrated in Section \ref{subsubsec: case 1 simulation gnpmc supp}, their method performs notably poorer than ours when there are multiple constraints for a row of the confusion matrix in the GNPMC problem.

\begin{figure}[!h]
	\centering
	\includegraphics[width=\textwidth]{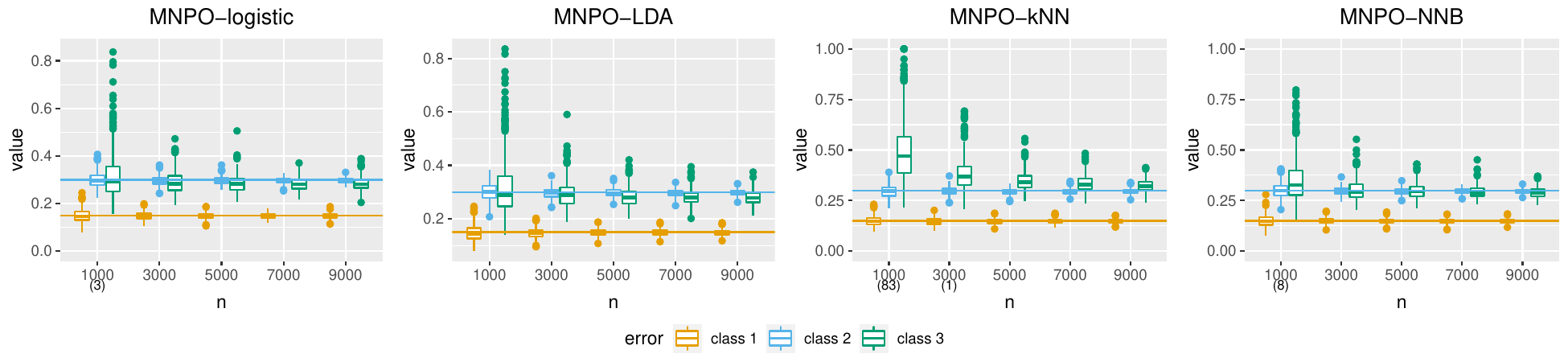}
	\caption{Per-class error rates under each classifier and training sample size setting in simulation case 1 for MNPO. Horizontal lines in corresponding colors mark the target control levels. For some graphs, there are additional numbers with brackets under the training sample size $n$, which indicates the number of simulations when the  algorithm reports infeasibility. }
	\label{fig: model1_mnpo}
\end{figure}

For comparison purposes, we denote their method as MNPO (Multiclass Neyman-Pearson Optimization, Algorithm 1 in \citealpapp{landgrebe2005neyman}) and plot the error rates of MNPO for different function classes to estimate $\tp_{Y|X}$ in Case 1 in Figure \ref{fig: model1_mnpo}. Since their method also requires sample splitting, we adopt the splitting ratio 0.5 the same as in NPMC-ER. Moreover, we set the grid search accuracy for MNPO to 0.01. Because MNPO shares similar conceptual underpinnings with our methods, the performance of MNPO, NPMC-CX, and NPMC-ER exhibits notable similarities. However, NPMC-CX has less variance when the sample size $n$ is small because it does not use sample splitting.

\begin{figure}[!h]
	\centering
	\includegraphics[width=\textwidth]{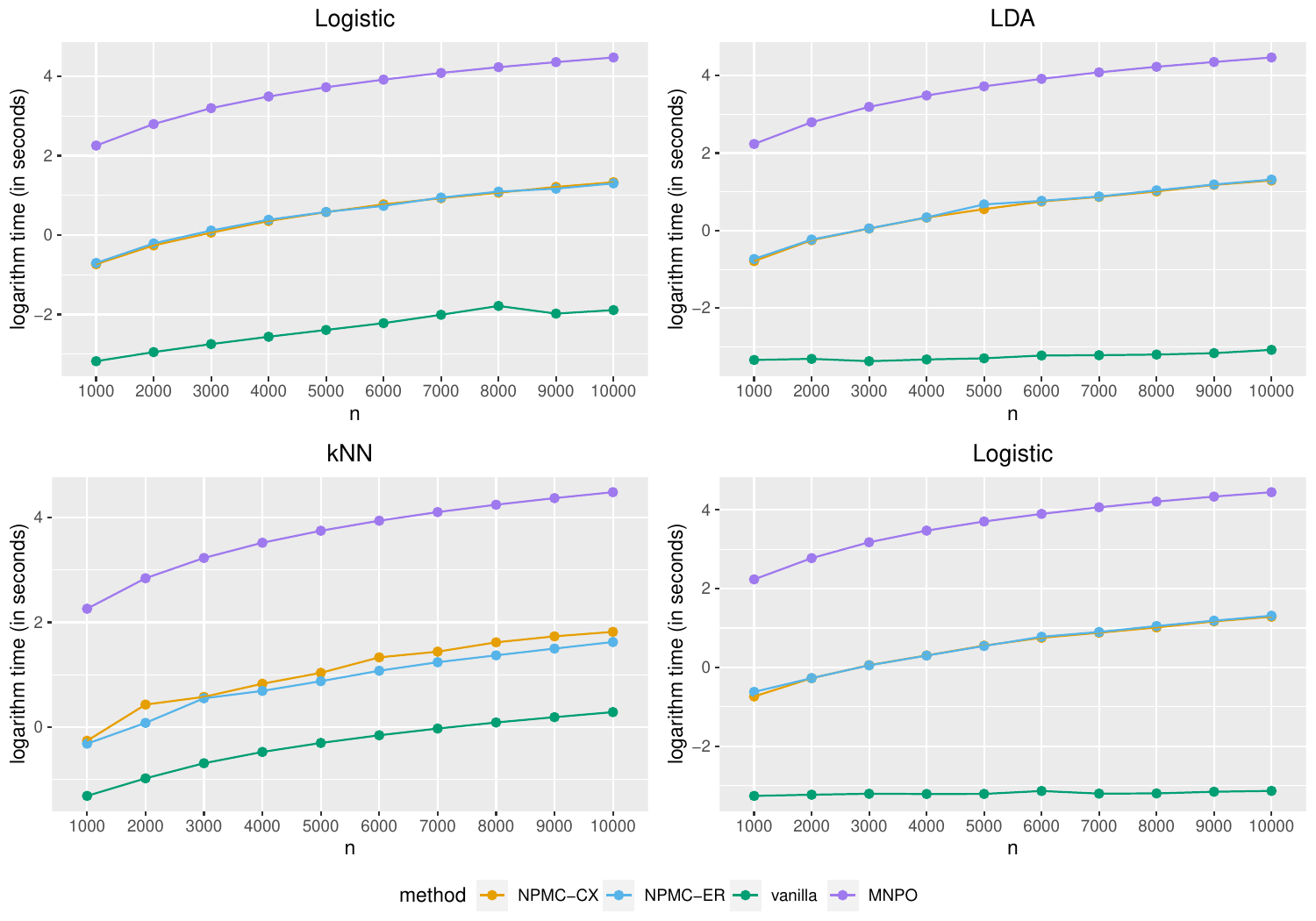}
	\caption{Logarithm (base: $e$) of the average computational time (in seconds) for different methods with different approaches to estimate $\tp_{Y|X}$. The average is taken over 500 replications for each sample size from 1000 to 10000 with increments of 1000.}
	\label{fig: mode1_time}
\end{figure}

Additionally, we depict the average computational time of each method for various sample sizes in Figure \ref{fig: mode1_time}, where it is evident that NPMC-CX and NPMC-ER are approximately 20 times faster than MNPO in this example.

\subsubsection{Case 2}\label{subsubsec: case 2 simulation supp}
In the first example, all five variables are independent Gaussian; therefore, four classifiers can estimate the posterior accurately. In this example, we consider a four-class correlated Gaussian conditional distribution, where $X|Y = k \sim N(\bm{\nu}_k, \bm{\Sigma})$ for $ k = 1, \ldots, 4$. And $\bm{\nu}_1 = (1,-2,0,-1,1)^T$, $\bm{\nu}_2 = (-1,1,-2,-1,1)^T$, $\bm{\nu}_3 = (2,0,-1,1,-1)$, $\bm{\nu}_4 = (1,0,1,2,-2)^T$, $\bm{\Sigma} = (0.1^{\mathds{1}(i \neq j)})_{p \times p}$, $p = 5$. The marginal distribution of $Y$ is $\tp(Y = 1) = 0.1$, $\tp(Y = 2) = 0.2$, $\tp(Y = 3) = 0.3$, and $\tp(Y = 4) = 0.4$.

\begin{figure}[!h]
	\centering
	\includegraphics[width=\textwidth]{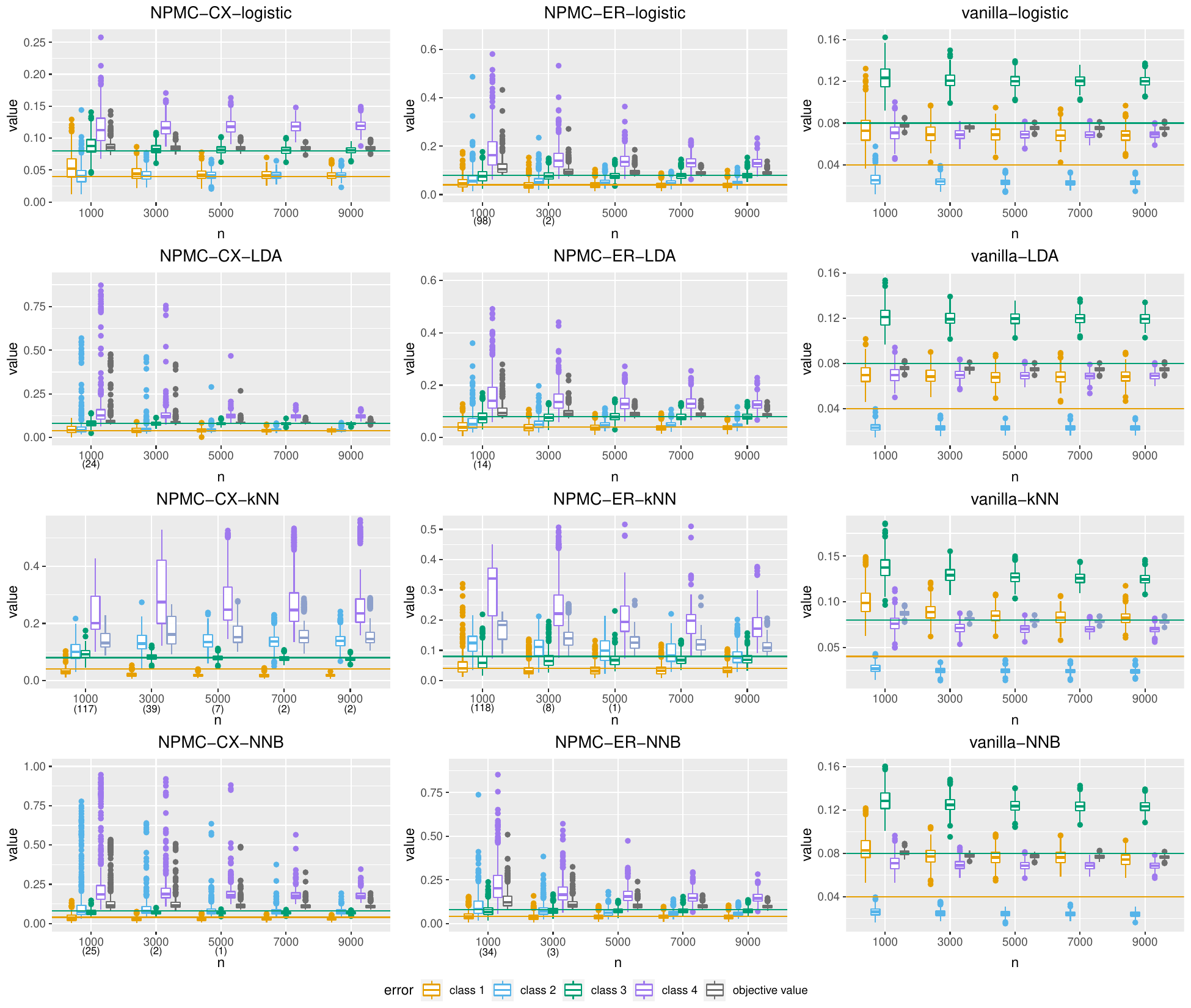}
	\caption{Per-class error rates and objective function values under each classifier and training sample size setting in simulation case 2. Horizontal lines in corresponding colors mark the target control levels. For some graphs, there are additional numbers with brackets under the training sample size $n$, which indicates the number of simulations that the algorithm reports infeasibility. }
	\label{fig: model2}
\end{figure}

The goal is to solve the following NPMC problem
\begin{align}
	&\min_{\phi} \quad \sum_{k=1}^4w_k\tp_{X|Y=1}(\phi(X)\neq 1)  \\
	&\text{s.t.} \quad \tp_{X|Y=1}(\phi(X)\neq 1) \leq 0.04,  \quad \tp_{X|Y=3}(\phi(X)\neq 3) \leq 0.08,
\end{align}
where $w_k = \tp(Y = k)$. Note that the objective function here includes errors of all four classes and is actually equal to the overall misclassification error rate $\tp(\phi(X) \neq Y)$.

Like in case 1, we study NPMC-CX, NPMC-ER, and vanilla classifiers with different function classes: logistic regression, LDA, $k$NN, and non-parametric na\"ive Bayes models with Gaussian kernels. The results are summarized in Figure \ref{fig: model2}. It can be observed that all four vanilla classifiers failed to control the error rates around the target levels. At the same time, NPMC-CX and NPMC-ER perform much better and successfully controlled $\tp_{X|Y=1}(\phi(X)\neq 1)$ and $\tp_{X|Y=3}(\phi(X)\neq 3)$ around $0.04$ and $0.08$, respectively. When the training sample size $n$ increases, the variances of error rates for each method tend to shrink. When $n$ is small, except for the NPMC-CX-logistic, the other NPMC methods sometimes lead to infeasible results. An interesting phenomenon here is that although the variables are not independent, NPMC-CX-NNB and NPMC-ER-NNB still work well in controlling the error rates. Besides, NPMC-CX-$k$NN seems to be over-conservative by strictly controlling $\tp_{X|Y=1}(\phi(X)\neq 1)$ under level $0.04$ when $n$ is large.

\subsubsection{Case 3: a variant of Case 1 with imbalanced sample sizes}
In this subsection, we investigate a variant of Case 1 characterized by imbalanced sample sizes across three classes to assess the robustness of NPMC-CX and NPMC-ER against sample size imbalances. Such imbalances are common, particularly in many real-world problems, as misclassifying observations from the rare class typically has greater consequences.

\begin{figure}[!h]
	\centering
	\includegraphics[width=\textwidth]{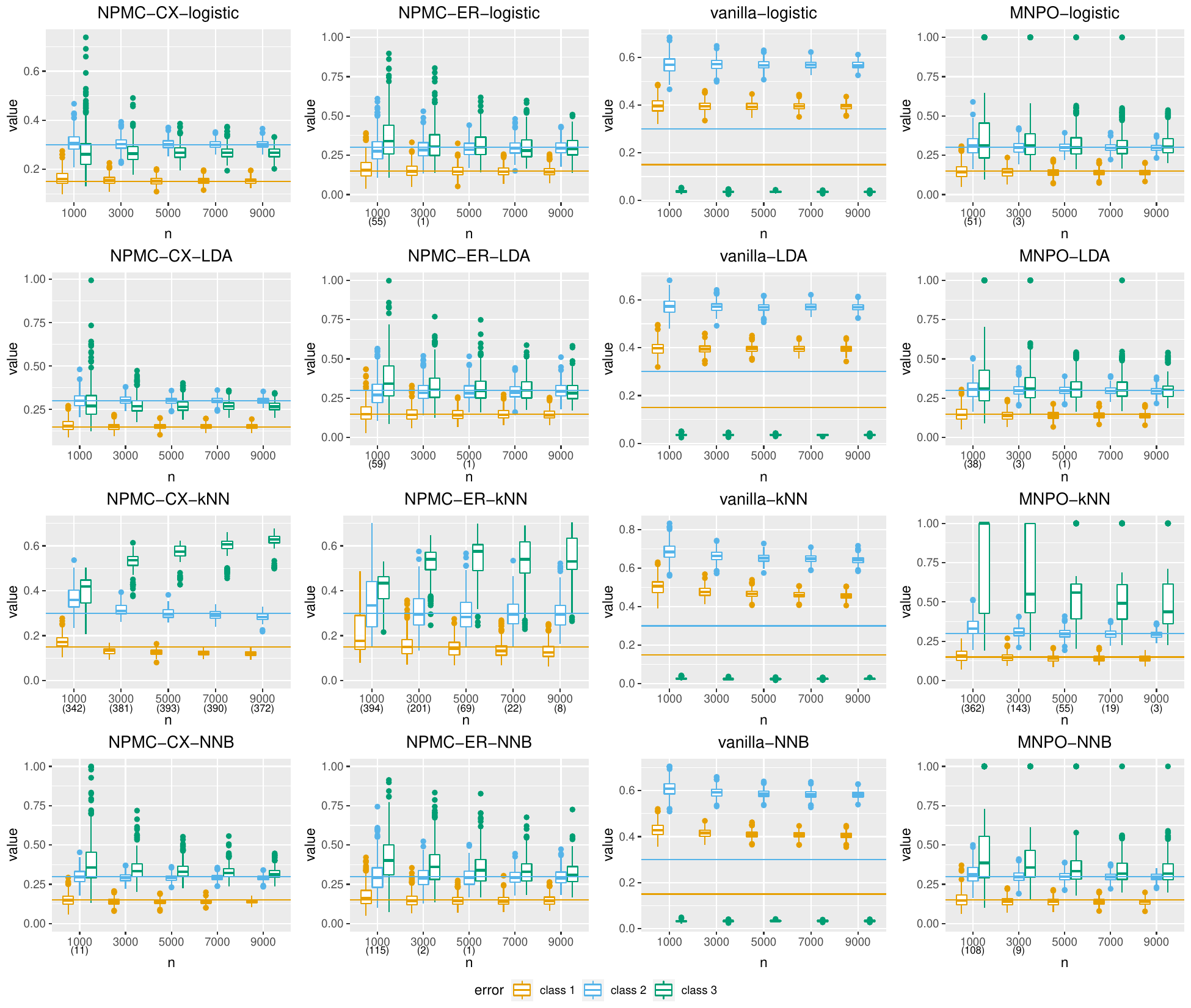}
	\caption{Per-class error rates under each classifier and training sample size setting in simulation case 3. Horizontal lines in corresponding colors mark the target control levels. For some graphs, there are additional numbers with brackets under the training sample size $n$, which indicates the number of cases that algorithms report infeasibility. }
	\label{fig: model4}
\end{figure}

We maintain the same 3-class NPMC problem in Case 1,  but alter the marginal distribution of the three classes to $\tp(Y=1) = \tp(Y=2) = 0.1$, $\tp(Y=3) = 0.8$. The performance of different methods is visualized in Figure \ref{fig: model4}. It is evident that, except for NPMC-CX-$k$NN, which produces numerous infeasibility reports (potentially due to the suboptimal choice of $k$), all other NPMC-CX and NPMC-ER methods perform well by controlling the error rates around the respective target levels. This underscores the robustness of our methods to data imbalance.  In contrast, compared to Figure \ref{fig: model1}, we can see that the data imbalance significantly damages the performance of vanilla methods on classes 1 and 2. Furthermore, MNPO methods \citepapp{landgrebe2005neyman} produce classifiers with an error rate of 1 on class 3 in some situations, which is not observed for NPMC-CX and NPMC-ER.

\subsubsection{Case 4: a variant of Case 1 with a GNPMC problem}\label{subsubsec: case 1 simulation gnpmc supp}
In our final simulation example, we explore a variant of Case 1 involving a GNPMC problem. Specifically, we consider the following GNPMC problem with the same joint distribution setting of $(X, Y)$ as in Case 1:
\begin{align}
	&\min_{\phi} \quad J(\phi) = \sum_{k=1}^3\sum_{r \neq k} w_{kr} \tp_{X|Y=k}(\phi(X)= r) \\
	&\text{s.t.} \quad \tp_{X|Y=1}(\phi(X)= 2) \leq 0.05, \quad \tp_{X|Y=3}(\phi(X)= 1) \leq 0.01, \quad \tp_{X|Y=3}(\phi(X)= 2) \leq 0.1,
\end{align}
where $w_{12} = w_{13} = 0.2$, $w_{21} = w_{23} = 0.4$, and $w_{31} = w_{32} = 0.3$. Notably,  with this specific weight setting, the objective function $J(\phi)$ corresponds to the overall misclassification error rate $\tp(\phi(X) \neq Y)$.
 
\begin{figure}[!h]
	\centering
	\includegraphics[width=\textwidth]{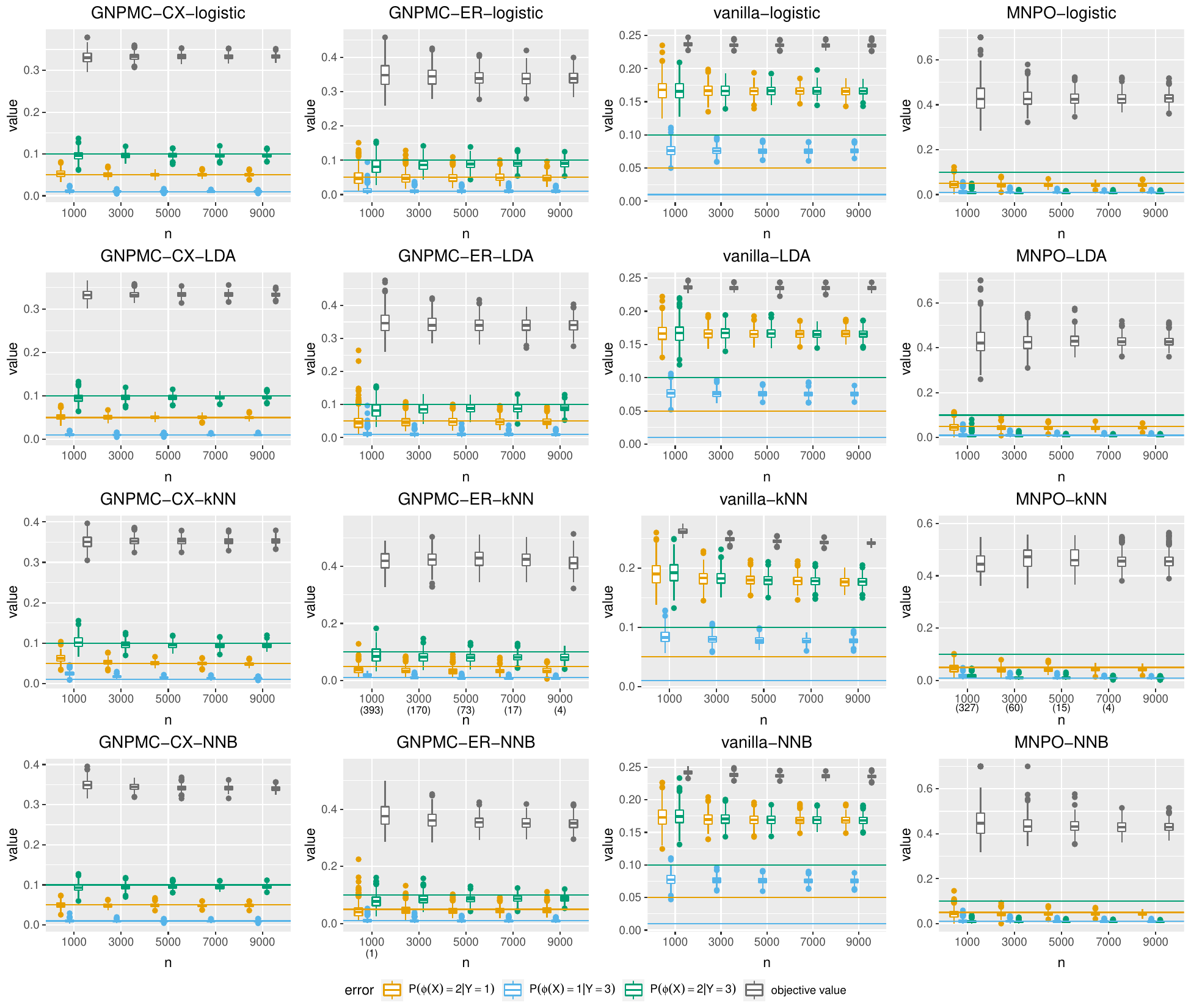}
	\caption{Per-class error rates and objective function values under each classifier and training sample size setting in simulation case 4. Horizontal lines in corresponding colors mark the target control levels. For some graphs, there are additional numbers with brackets under the training sample size $n$, which indicates the number of cases that algorithms report infeasibility. }
	\label{fig: model3}
\end{figure}

We study GNPMC-CX, GNPMC-ER, and vanilla classifiers equipped with different function classes to estimate $\tp_{Y|X}$: logistic regression, LDA, $k$NN, and non-parametric na\"ive Bayes models with Gaussian kernels. The results are summarized in Figure \ref{fig: model3}.

From the results, it can be seen that all GNPMC-CX and GNPMC-ER methods can successfully control the error rates around the target levels, except for GNPMC-CX-$k$NN with $n = 1000$, which demonstrates the power of our GNPMC methods. In contrast, all vanilla classifiers fail to control the error rate. We also include the method MNPO (Algorithm 1 in \citealpapp{landgrebe2005neyman}) into comparison. Similar to the setting in Case 1, we set the sample spitting ratio to be 0.5 and the grid accuracy to be 0.01 for MNPO. Compared to GNPMC methods, MNPO tends to over-control the error rate $\tp(\phi(X) \neq 2|Y=3)$, leading to an inflation of the objective value. This occurs because MNPO assigns only one weight to each class during classification. In this example, there is a much more stringent constraint on $\tp_{X|Y=3}(\phi(X)= 1)$, which forces MNPO to assign a large weight to class 3 to satisfy this constraint, and such a large weight is an overkill for $\tp_{X|Y=3}(\phi(X)= 2)$. In contrast, GNPMC methods utilize different Lagrangian multipliers $\lambda_{31}$ and $\lambda_{32}$ for different  constraints $\tp_{X|Y=3}(\phi(X)= 1)$ and $\tp_{X|Y=3}(\phi(X)= 2)$. A strict control on $\tp_{X|Y=3}(\phi(X)= 1)$ can only make the GNPMC methods choose a large $\lambda_{31}$, instead of large $\lambda_{32}$, which resolves this issue.  This observation aligns with our theoretical findings, indicating that GNPMC methods satisfy GNPMC oracle properties under strong duality and mild conditions, showcasing their optimality compared to other approaches, such as MNPO.

\subsection{Real data studies}

\subsubsection{LendingClub dataset}\label{subsubsec: lendingclub supp}
In this subsection, we delve deeper into the LendingClub dataset, providing additional details beyond what was covered in Section \ref{subsec: real-data}. We also extend our analysis to the GNPMC problem on this dataset and provide more experimental results. 

First, we want to introduce more details about LendingClub dataset in addition to the information we provided in Section \ref{subsec: real-data}. The original dataset contains 887379 observations and 74 features, including the loan status as the response variable. Initially, we excluded observations labeled as ``Current" or ``Issued", indicating ongoing or just-started loan processes, as they are not pertinent to our analysis. The remaining classes were grouped into three categories, as discussed in Section \ref{subsec: real-data}. Next, due to varying data availability across states, we aggregated the ``state" variable into a new ``region" variable, which has five levels: ``West", ``SouthWest", ``SouthEast", ``MidWest", and ``NorthEast`".  Similarly, to address the lack of data in certain categories, we merged some levels of the categorical variable ``purpose" into four levels: ``credit\_card", ``debt\_consolidation", ``other", and ``purchase". Subsequently, features and observations containing over 10\% missing data were removed.  The remaining missing values were then imputed using the median or mode of each feature, depending on the variable type (median for numerical features and mode for categorical features). Additionally, variables collected after loan issuance, such as ``last\_pymnt\_amnt" (last total payment amount received), ``total\_pymnt"(payments received to date for total amount funded), and ``recoveries" (post charge off gross recovery), which contain information related to loan status, were removed to focus on predicting loan status at issuance. After preprocessing, the dataset consists of 264274 observations and 26 features, including the  loan status as the response variable, A detailed description of the features can be found at \url{https://www.kaggle.com/datasets/adarshsng/lending-club-loan-data-csv/data} and \url{http://rstudio-pubs-static.s3.amazonaws.com/290261_676d9bb194ae4c9882f599e7c0a808f2.html}. 

We already explored the NPMC problem on this dataset in Section \ref{subsec: real-data}. We now want to extend our analysis to the GNPMC problem. As previously discussed, misclassifying observations across different classes can have varying impacts on lenders. For instance, extending a loan to an individual who ultimately defaults can result in a complete loss of funds, representing a significant risk for the lender. Conversely, declining a loan to an individual who would have repaid the full amount on time only results in the loss of potential interest income. The GNPMC framework in \eqref{eq: G-NPMC} allows for more sophisticated constraints on misclassification errors compared to the NPMC framework \eqref{eq: NPMC}. For example, misclassifying an applicant from class 1 as class 2 (or vice versa) may not have substantial consequences. However, misclassifying applicants from either class 1 or 2 as class 3 can potentially lead to significant issues. Motivated by this insight, we formulate the following GNPMC problem: 
\begin{align}
	&\min_{\phi} \quad \sum_{k=1}^3\sum_{r \neq k} w_{kr} \tp_{X|Y=k}(\phi(X)= r) \\
	&\text{s.t.} \quad \tp_{X|Y=1}(\phi(X) = 3) \leq \alpha_{13},  \quad \tp_{X|Y=2}(\phi(X) = 3) \leq \alpha_{23},
\end{align}
where $w_{13} = w_{23} = 0.1706$, $w_{21} = w_{23} = 0.0729$, and $w_{31} = w_{32} = 0.7565$. These numbers 0.1706, 0.0729, and 0.7565 represent the marginal probabilities of three classes, ensuring that the objective function corresponds to the overall misclassification error rate $\tp(\phi(X) \neq Y)$. 

\begin{figure}[!h]
	\centering
	\includegraphics[width=\textwidth]{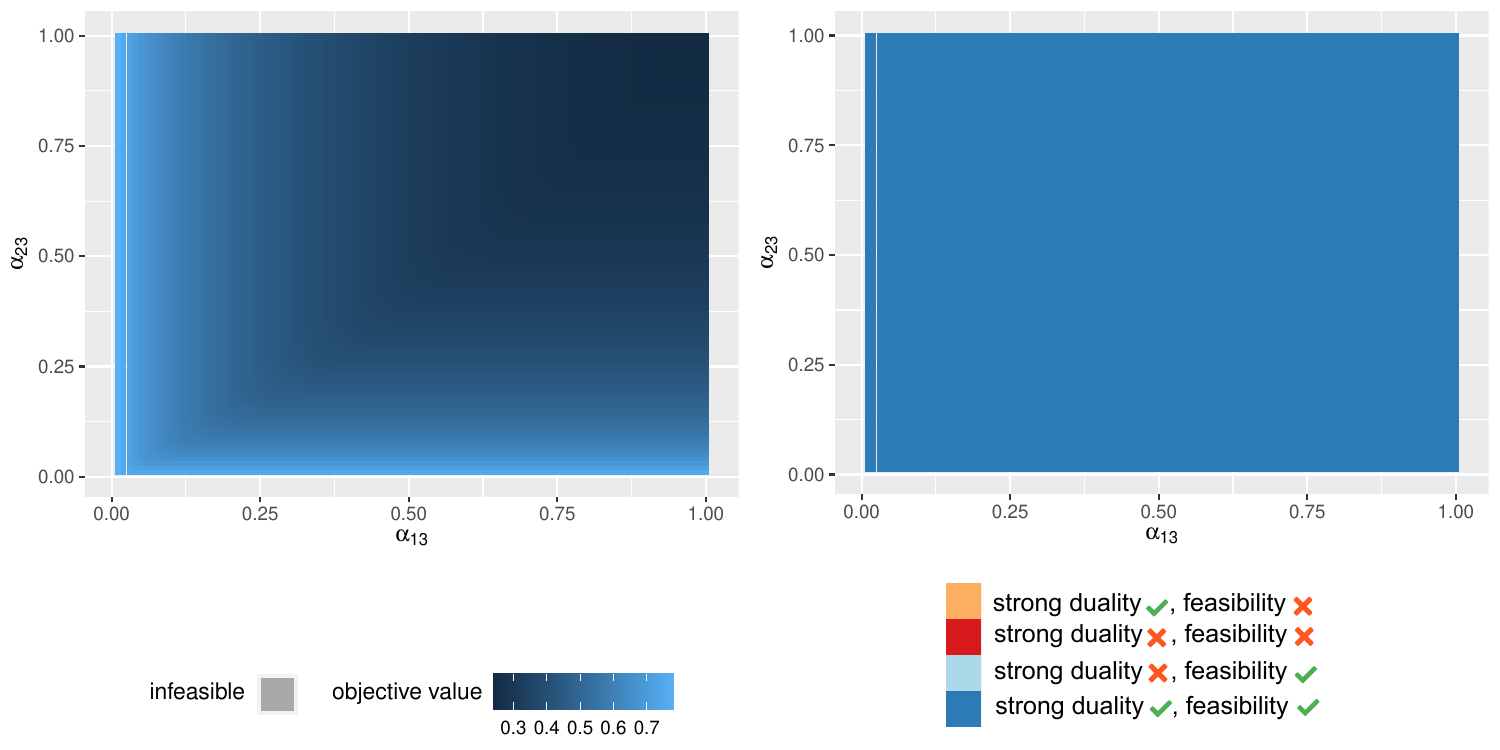}
	\caption{Strong daulity and feasibility of GNPMC problem on the LendingClub dataset with different target error levels: predicted by Algorithm \ref{algo: df check CX gnpmc} with GNPMC-CX-logistic.}
	\label{fig: loanclub_GNPMC_logistic}
\end{figure}

\begin{figure}[!h]
	\centering
	\includegraphics[width=\textwidth]{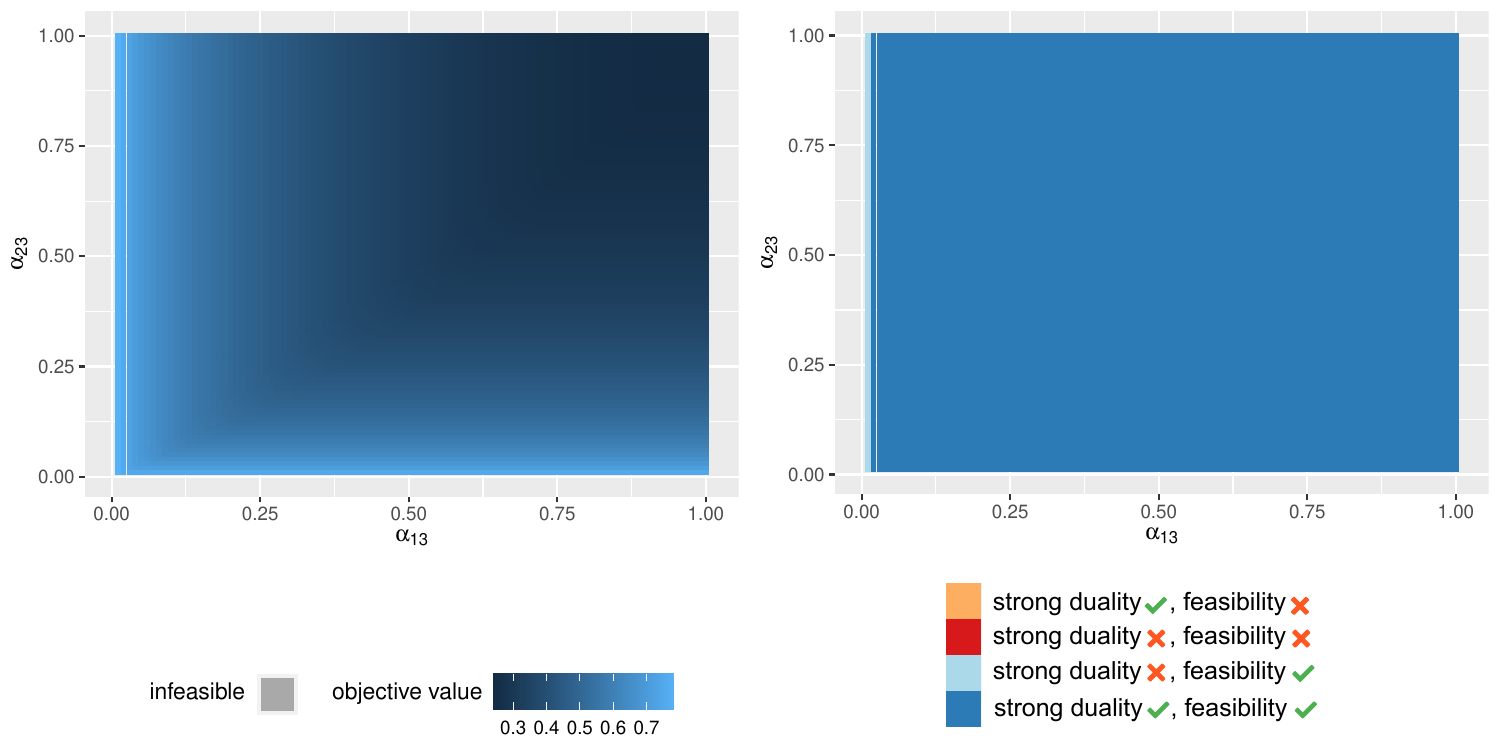}
	\caption{Strong duality and feasibility of GNPMC problem on the LendingClub dataset with different target error  levels: predicted by Algorithm \ref{algo: df check ER gnpmc} with GNPMC-ER-RF.}
	\label{fig: loanclub_GNPMC_rf}
\end{figure}

Similar to our discussion in Sections \ref{subsec: simulations} and \ref{subsec: real-data}, the practitioners can experiment with different target levels $(\alpha_{13}, \alpha_{23})$ using our feasibility and strong duality checking algorithms (Algorithms \ref{algo: df check CX gnpmc} and \ref{algo: df check ER gnpmc}) to assess problem difficulty and determine the target level based on feasibility and practical considerations. We show the strong duality, feasibility, and objective values predicted by Algorithm \ref{algo: df check CX gnpmc} with GNPMC-CX-logistic (GNPMC-CX with $\mathcal{M}$ as logistic regresion) and Algorithm \ref{algo: df check ER gnpmc} with GNPMC-ER-RF (GNPMC-ER with $\mathcal{M}$ as random forests) on the entire data set with different $(\alpha_{13}, \alpha_{23}) \in [0.01, 1]^2$ with a grid precision of 0.01, in Figures \ref{fig: loanclub_GNPMC_logistic} and \ref{fig: loanclub_GNPMC_rf}, respectively. In the contrast to Figures \ref{fig: loanclub_NPMC_logistic} and \ref{fig: loanclub_GNPMC_rf}, we do not observe a tradeoff between controlling $\tp_{X|Y=1}(\phi(X)\neq 3)$ and $\tp_{X|Y=2}(\phi(X)\neq 3)$. Instead, we see a correlative relationship. For example, fixing a specific $\alpha_{13}$, increasing $\alpha_{23}$ from $0$ to the level around $\alpha_{13}$ will decrease the predicted objective value. However, the predicted objective value remains unchanged when increasing $\alpha_{23}$ from the level around $\alpha_{13}$ to 1. This phenomenon indicates that classes 1 and 2 behave very similarly, which is reasonable because both represent loans that cannot be repaid on time with different default levels. Furthermore, the problem is predicted to be feasible for all choices $(\alpha_1, \alpha_2)$ we tried.

Next, we set $\alpha_{13} = 0.1$ and $\alpha_{23} = 0.15$, and run GNPMC-CX-logistic and GNPMC-ER-RF with vanilla logistic regression and random forests as benchmarks. We randomly split the entire data into 50\% training and 50\% testing data in 500 replications. Box plots  in Figure \ref{fig: loanclub_NPMC} illustrate the error rates $\tp_{X|Y=1}(\phi(X) = 3)$ and $\tp_{X|Y=2}(\phi(X) = 3)$ under each classifier and training sample size setting. We observe that vanilla logistic regression and random forests tend to classify all observations into class 3 due to the high imbalance of sample sizes. In contrast, GNPMC-CX-logistic and GNPMC-ER-RF effectively control the error $\tp_{X|Y=1}(\phi(X) = 3)$ around 0.1. Additionally, the error $\tp_{X|Y=2}(\phi(X) = 3)$ is lower than the target 0.15. This serves as further evidence of the correlative relationship between two errors and the similarity between classes 1 and 2.

\begin{figure}[!h]
	\centering
	\includegraphics[width=\textwidth]{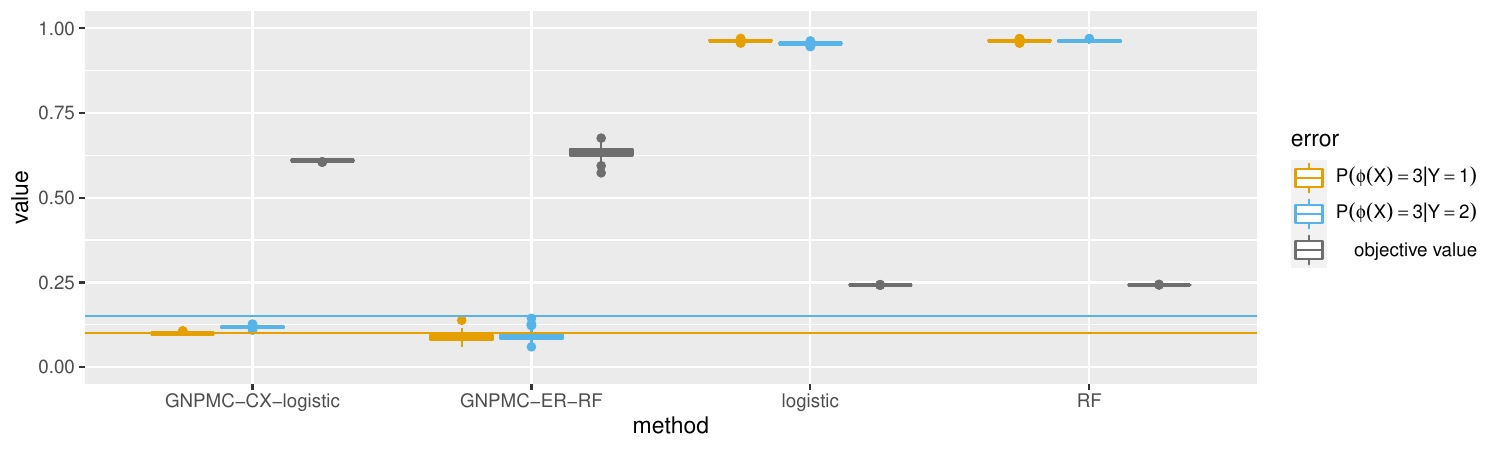}
	\caption{Error rates to be controlled and objective function values under each classifier for the GNPMC problem on the LendingClub dataset. Horizontal lines in corresponding colors mark the target control levels.}
	\label{fig: loanclub_GNPMC}
\end{figure}

\subsubsection{Dry bean dataset}\label{subsubsec: beans}
This dataset comes from the transformed images of 13,611 grains of 7 different registered dry beans \citepapp{koklu2020multiclass}. The seven types and their corresponding sample sizes are Barbunya (1322), Bombay (522), Cali (1630), Dermosan (3546), Horoz (1928), Seker (2027), and Sira (2636). The goal is to predict the bean type correctly. There are 16 predictors of the grains in total, consisting of 12 dimensions and four shape forms. The data is available on the UCI machine learning repository (\url{https://archive.ics.uci.edu/ml/datasets/Dry+Bean+Dataset}).

For convenience, we recode the bean types into classes 1 through 7. In each replication, we randomly split the data into 10\% training and 90\% test data per class. Consider the following NPMC problem
\begin{align}
	&\min_{\phi} \quad \frac{1}{4}\left[\tp_{X|Y=3}(\phi(X)\neq 3) + \tp_{X|Y=5}(\phi(X)\neq 5) + \tp_{X|Y=6}(\phi(X)\neq 6) + \tp_{X|Y=7}(\phi(X)\neq 7)\right]  \\
	&\text{s.t.} \quad \tp_{X|Y=1}(\phi(X)\neq 1) \leq 0.05,  \quad \tp_{X|Y=2}(\phi(X)\neq 2) \leq 0.01, \quad \tp_{X|Y=4}(\phi(X)\neq 4) \leq 0.03.
\end{align}

\begin{figure}[t]
	\centering
	\includegraphics[width=\textwidth]{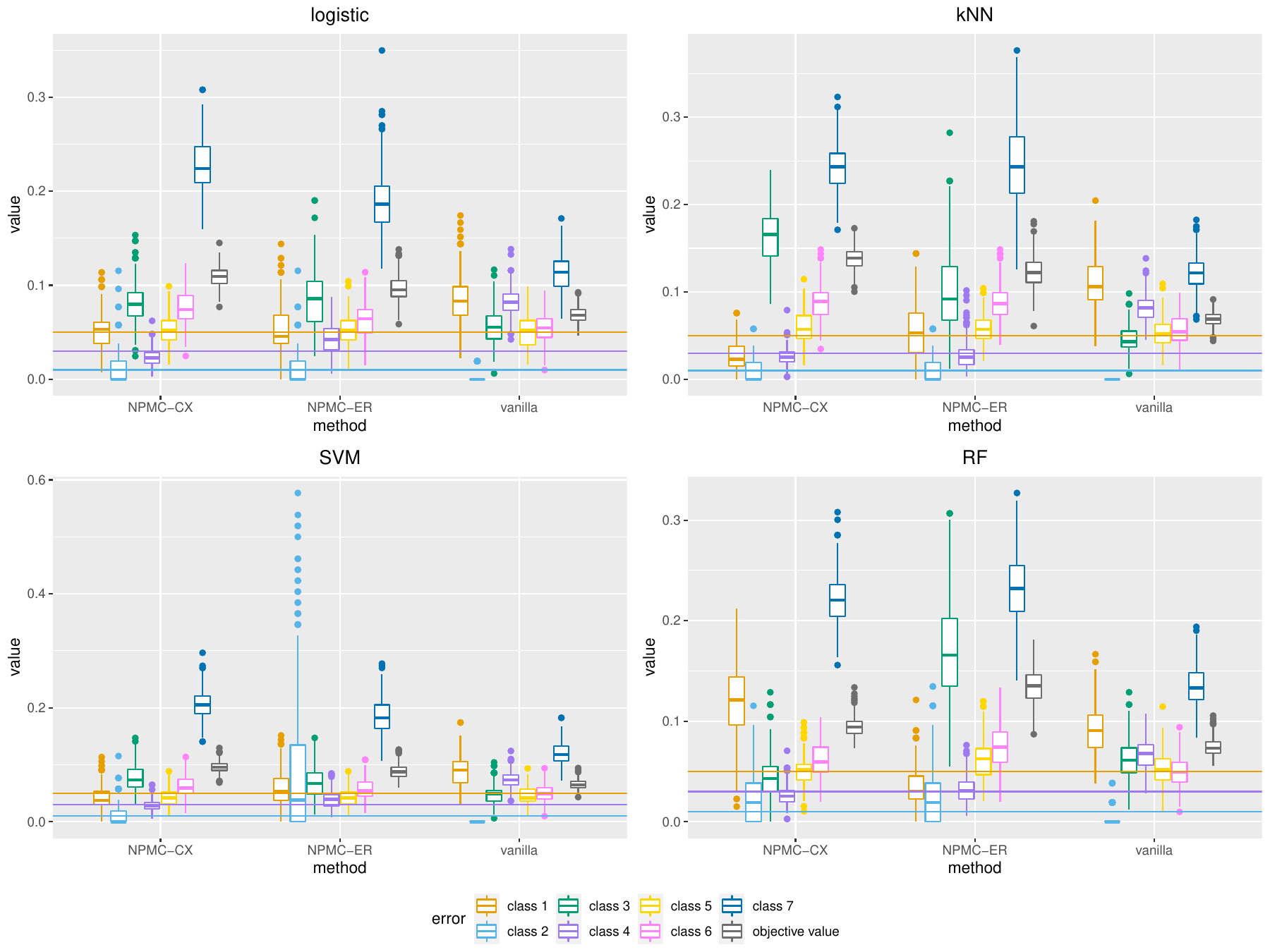}
	\caption{Per-class error rates and objective function values under each classifier for the dry bean dataset. Horizontal lines in corresponding colors mark the target control levels. In some graphs, additional values are displayed in brackets beneath the training sample size, $n$. These values represent the number of instances where the algorithms reported infeasibility during evaluation. }
	\label{fig: beans}
\end{figure}

We study NPMC-CX, NPMC-ER, and vanilla classifiers based on logistic regression, SVM, $k$NN, and random forest. The performance of these methods is summarized in Figure \ref{fig: beans}. Firstly, we can see that four vanilla classifiers can only control the error rate of class 2 while failing to control the error rates of classes 1 and 4. NPMC-CX and NPMC-ER work well to control the error rates around the target levels, except for NPMC-ER-SVM and NPMC-CX-RF. NPMC-ER-SVM leads to a large variance in the error rate of class 2, which might be caused by the limited sample size of class 2. Furthermore, NPMC-CX-RF fails to control the class 1 error rate. This may be caused by overfitting, because the random forest is a very complex model and the training data is used both in fitting the model and searching for $\hblambda$ in Algorithm \ref{algo}).

\subsubsection{Statlog (Landsat satellite) dataset}\label{subsubsec: sat}
This dataset contains the multi-spectral values of pixels in $3 \times 3$ neighborhoods in satellite images. We aim to predict the central pixel label in each neighborhood. There are 36 predictors for each of the 6435 observations, representing the multi-spectral values. Central pixel labels and their corresponding sample sizes are red soil (1533), cotton crop (703), grey soil (1358), damp grey soil (626), soil with vegetation stubble (707), and very damp grey soil (1508). We recode the six classes into classes 1 to 6, respectively. In each replication, we randomly split the data into 10\% training and 90\% test data per class. 

We consider the following NPMC problem
\begin{align}
	&\min_{\phi} \quad \frac{1}{6}\sum_{k=1}^6\tp_{X|Y=k}(\phi(X)\neq k)  \\
	&\text{s.t.} \quad \tp_{X|Y=3}(\phi(X)\neq 3) \leq 0.15, \quad \tp_{X|Y=4}(\phi(X)\neq 4) \leq 0.2, \quad \tp_{X|Y=5}(\phi(X)\neq 5) \leq 0.1.
\end{align}

\begin{figure}[t]
	\centering
	\includegraphics[width=\textwidth]{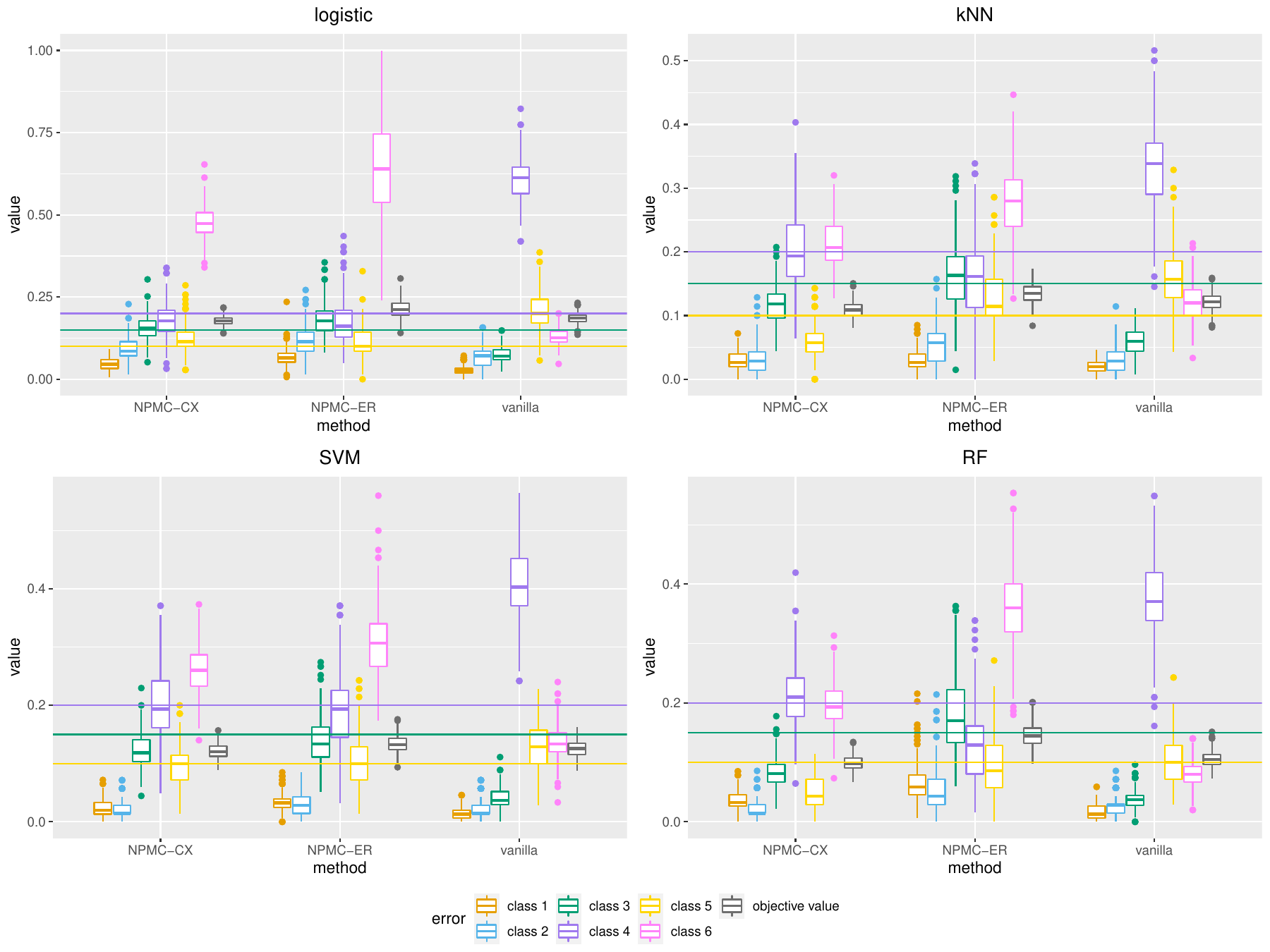}
	\caption{Per-class error rates under each classifier for the statlog dataset. Horizontal lines in corresponding colors mark the target control levels. In some graphs, additional values are displayed in brackets beneath the training sample size, $n$. These values represent the number of instances where the algorithms reported infeasibility during evaluation.}
	\label{fig: sat}
\end{figure}

As in Section \ref{subsubsec: beans}, we explore NPMC-CX, NPMC-ER, and vanilla classifiers based on logistic regression, SVM, $k$NN, and random forest. The results are available in Figure \ref{fig: sat}. It can be seen that vanilla-logistic and vanilla-$k$NN only control $\tp_{X|Y=3}(\phi(X)\neq 3)$ well, while vanilla-SVM and vanilla-RF successfully control $\tp_{X|Y=3}(\phi(X)\neq 3)$ and $\tp_{X|Y=5}(\allowbreak\phi(X)\neq 5)$ around the target levels. NPMC-CX and NPMC-ER successfully control all three error rates around the target levels in all cases. In addition, it is interesting that all vanilla methods over-control $\tp_{X|Y=3}(\phi(X)\neq 3)$, which might damage the performance on other classes. NPMC-CX and NPMC-ER can fix this issue and relax this control by increasing $\tp_{X|Y=3}(\phi(X)\neq 3)$ while still controlling other classes' error rates around the target levels. Thanks to this, compared to the vanilla methods, we observe that NPMC-CX and NPMC-ER can create classifiers that control the error rates around the levels approximately without increasing the objective function value too much.

\subsubsection{Dementia dataset}
Worldwide, the prevention, treatment, and precise diagnosis of subtypes of dementia is a top healthcare priority and a critical clinical focus. This dataset comes from a preliminary study based on medical and neuropathology records from participants enrolled in an NIH-funded AD research center (ADRC) at New York University. Each participant signed an IRB-approved form to donate the brain for post-mortem examination. Their clinical evaluation included an interview according to the Brief Cognitive Rating Scale, rating on the Global Deterioration Scale (GDS) \citepapp{reisberg1993empirical}, and Geriatric Depression Scale. Subjects with brain pathology, such as tumors, neocortical infarction, or diabetes, were excluded. 
 
 The selection of records that included post-mortem dementia diagnosis yielded a total of 302 observations. The original dataset contains 10 dementia subtypes. Since sample sizes of some subtypes are too small, we keep subtypes Normal (class 1) and Alzheimer's disease (class 2), and merge the other eight subtypes into one class (class 3). And the final sample sizes of them are 103,  89, and 110, respectively. For each observation, we retrieved information from the most recent clinic visit. There are 13 predictors, including age, sex, race, education, and the nine most relevant clinical measures after list-wise deletion.
 
Our goal is to solve the following NPMC problem
\begin{align}
	&\min_{\phi} \quad \tp_{X|Y=3}(\phi(X)\neq 3) \\
	&\text{s.t.} \quad \tp_{X|Y=1}(\phi(X)\neq 1) \leq 0.1,  \quad \tp_{X|Y=2}(\phi(X)\neq 2) \leq 0.02.
\end{align}

\begin{figure}[!h]
	\centering
	\includegraphics[width=\textwidth]{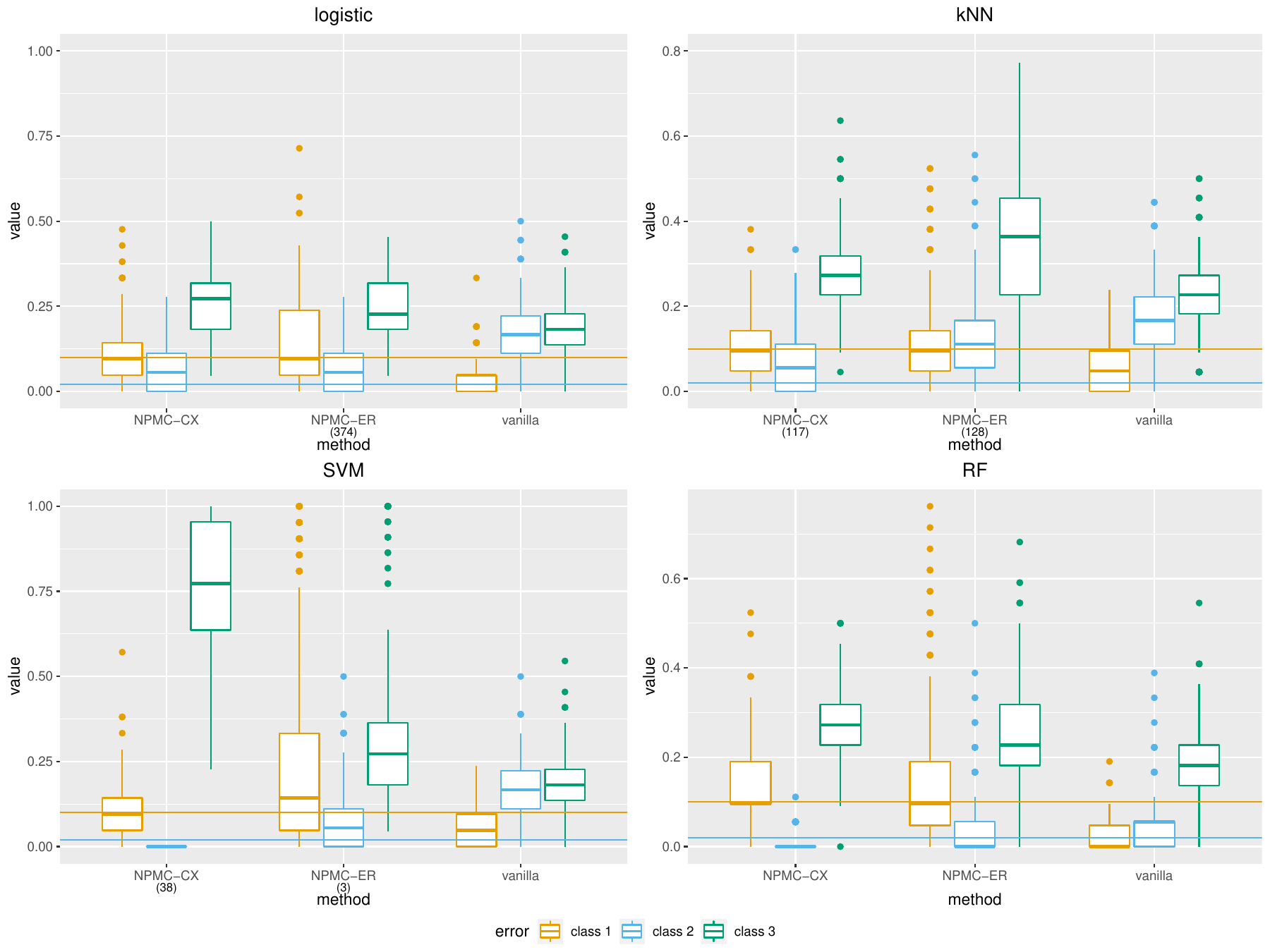}
	\caption{Per-class error rates under each classifier for the dementia dataset without 0.5-SMOTE. Horizontal lines in corresponding colors mark the target control levels. In some graphs, additional values are displayed in brackets beneath the training sample size, $n$. These values represent the number of instances where the algorithms reported infeasibility during evaluation. }
	\label{fig: yang}
\end{figure}

\begin{figure}[!h]
	\centering
	\includegraphics[width=\textwidth]{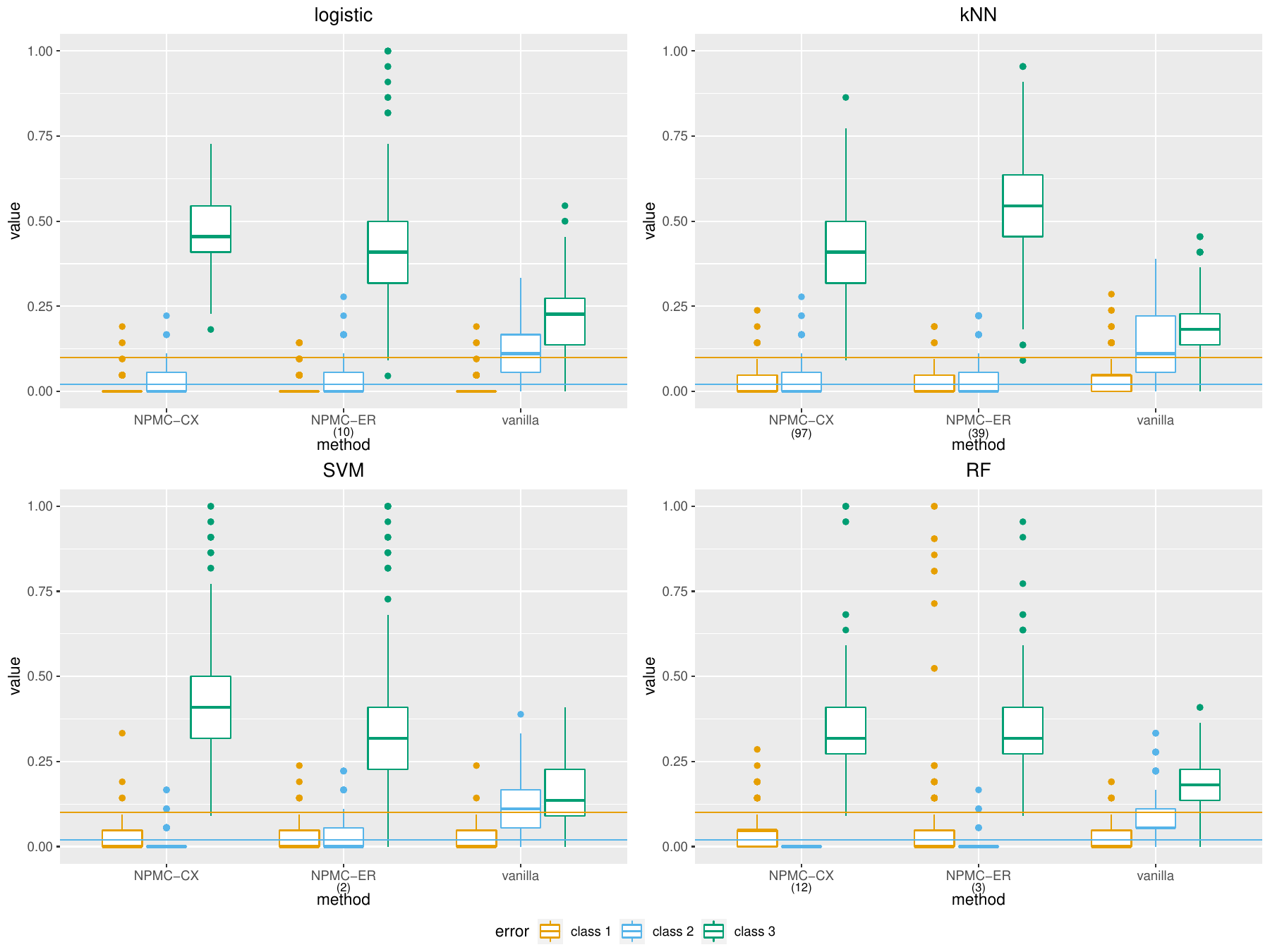}
	\caption{Per-class error rates and objective function values for the dementia dataset with 0.5-SMOTE. Horizontal lines in corresponding colors mark the target control levels. In some graphs, additional values are displayed in brackets beneath the training sample size, $n$. These values represent the number of instances where the algorithms reported infeasibility during evaluation. }
	\label{fig: yang-smote}
\end{figure}

Similar to the previous two real data studies, we fit NPMC-CX, NPMC-ER, and vanilla classifiers based on logistic regression, SVM, $k$NN, and random forest. The results are available in Figure \ref{fig: yang}. It can be seen that NPMC-CX-logistic, NPMC-CX-SVM, NPMC-ER-SVM, NPMC-CX-RF, NPMC-ER-RF, and vanilla-RF approximately control the $\tp_{X|Y=1}(\phi(X)\neq 1)$ and $\tp_{X|Y=2}(\phi(X)\neq 2)$ around the target levels, while the other methods fail. Besides, NPMC-ER-logistic often fails to give a feasible solution. These issues may be due to the limited sample size.

Motivated by the over-sampling strategy often used in imbalance classification to create synthetic observations for minor classes \citepapp{feng2021imbalanced}, next, we try to enlarge the training dataset before running NPMC algorithms. One of the most popular over-sampling is the synthetic minority over-sampling technique (SMOTE) \citepapp{chawla2002smote}, which creates synthetic samples via nearest neighbors. We can briefly describe the SMOTE algorithm with the number of nearest neighbors $\tilde{k}$ as follows. To enlarge the sample size of class $k$, for each class-$k$ sample $\bx_0$, randomly choose one of its $\tilde{k}$ nearest neighbors $\bx_1$ and generate a uniform random variable $u \sim \textrm{Unif}(0, 1)$. Then a new synthetic observation of class $k$ is generated as $\tilde{x} = u\bx_1 + (1-u)\bx_0$. Compared to other over-sampling methods with replacement, SMOTE benefits from more variations and uncertainty. 

In our case, we have limited observations for all classes. Therefore, we need to enlarge the whole dataset instead of a single class. To make our over-sampling procedure less aggressive, we adjusted the original SMOTE algorithm and conducted a conservative version called ``0.5-SMOTE" by replacing the $\textrm{Unif}(0, 1)$ with $\textrm{Unif}(0, 0.5)$. Compared to the original SMOTE, the synthetic samples generated by 0.5-SMOTE are closer to the real samples.

Next, in each of the 500 replications, we conduct 0.5-SMOTE with 5NN to generate a new training set with five times the sample size of the original data, then run NPMC and vanilla algorithms on this new training set. We summarize the results in Figure \ref{fig: yang-smote}. Compared to the results without 0.5-SMOTE, the performance of NPMC algorithms improves significantly, and all of them successfully control the error rates around the corresponding levels. At the same time, all vanilla approaches fail to control $\tp_{X|Y=2}(\phi(X)\neq 2)$. It can also be seen that NPMC methods tend to be conservative when controlling $\tp_{X|Y=1}(\phi(X)\neq 1)$, which might be caused by overfitting. When the sample size is small, doing 0.5-SMOTE can help NPMC methods succeed, making our algorithms more useful in practice.


\section{Technical Lemmas and Propositions}
\begin{lemma}\label{lem: finite bound}
	Consider Algorithm \ref{algo} (NPMC-CX). Under Assumptions \ref{asmp: second order info} and \ref{asmp: rademacher}, for any bounded $\Lambda \subseteq \mathbb{R}_+^{|\mathcal{A}|}$, we have
	\begin{equation}
		\tp\left(\sup_{\blambda \in \Lambda}\norm{\hL(\blambda, \hphi_{\blambda}) - L(\blambda, \phi_{\blambda}^*)} > \tau \right) 
		\lesssim \exp\{-n(\tau^2\wedge 1)\} + \tau^{-1}\max_k \te\norma{\widehat{\tp}_{Y|X}(Y=k)-\tp_{Y|X}(Y=k)},
	\end{equation}
	for any $\tau \gtrsim C_{\textup{Rad}}(n) + \max_k \te\norm{\widehat{\tp}_{Y|X}(Y=k)-\tp_{Y|X}(Y=k)}$.
\end{lemma}


\begin{lemma}\label{lem: uniform bdd 1}
	Consider Algorithm \ref{algo} (NPMC-CX). Suppose Assumptions \ref{asmp: posterior distance as}, \ref{asmp: second order info}, \ref{asmp: error rate continuity} hold. For any bounded set $\Lambda \subseteq \mathbb{R}^{|\mathcal{A}|}_+$, $\lim_{n \rightarrow \infty}\sup_{\blambda \in \Lambda}\norm{\hL(\blambda, \hphi_{\blambda}) - L(\blambda, \hphi_{\blambda})} = 0$ a.s..
\end{lemma}

\begin{lemma}\label{lem: uniform bdd 2}
	Consider Algorithm \ref{algo} (NPMC-CX). Suppose Assumptions \ref{asmp: posterior distance as}, \ref{asmp: second order info}, \ref{asmp: error rate continuity} hold. For any bounded set $\Lambda \subseteq \mathbb{R}^{|\mathcal{A}|}_+$, $\lim_{n \rightarrow \infty}\sup_{\blambda \in \Lambda}\norm{L(\blambda, \hphi_{\blambda}) - L(\blambda, \phi^*)} = 0$ a.s..
\end{lemma}

\begin{lemma}\label{lem: uniform conv algorithm 2}
	Consider Algorithm \ref{algo: erm} (NPMC-ER). We define $L(\blambda, \phi)$ as in \eqref{eq: f eq 2}. Under Assumptions \ref{asmp: strong duality} and \ref{asmp: second order info}, for any bounded set $\Lambda \subseteq \mathbb{R}_+^{|\mathcal{A}|}$,
	\begin{equation}
		\tp\left(\sup_{\blambda \in \Lambda}\norm{\hL(\blambda, \hphi_{\blambda}) - L(\blambda, \phi_{\blambda}^*)} >  \tau\right) 
		\lesssim \exp\{-Cn(\tau^2 \wedge 1)\} + \tau^{-1}\max_k \te\norma{\widehat{\tp}_{Y|X}(Y=k)-\tp_{Y|X}(Y=k)},
	\end{equation}
	if $\tau \gtrsim \sqrt{\frac{1}{n}}$.
\end{lemma}


\section{Proofs}
We mean ``without loss of generality" by writing ``WLOG".  
\subsection{Proof of lemmas}

\subsubsection{Proof of Lemmas \ref{lem: feasibility toy example} and \ref{lem: strong duality toy example}}
Denote the density of $N(\bmu_k, \bm{I}_p)$ as $f_k$. First, let us consider a classifier $\varphi_{\blambda}^*$ indexed by $\blambda = (\lambda_1, \lambda_2)^T \in \mathbb{R}^2_+$ as
\begin{equation}
	\varphi_{\blambda}^*: \bx \mapsto \argmax_{k=1:2}\{\lambda_kf_k(\bx)\} = \argmax_{k=1:2}\bigg\{
	\bmu_k^T\bx - \frac{1}{2}\twonorm{\bmu_k}^2 + \log \lambda_k\bigg\}.
\end{equation}

Note that the misclassification error rate of $\varphi_{\blambda}^*$ for class 1 can be written as
\begin{align}
	R_1(\varphi^*_{\blambda}) &= \tp_{X|Y=1}(\varphi^*_{\blambda}(X) \neq 1) \\
	&= \tp\bigg(\bmu_1^T\bx - \frac{1}{2}\twonorm{\bmu_1}^2 + \log \lambda_1 \leq \bmu_2^T\bx - \frac{1}{2}\twonorm{\bmu_2}^2 + \log \lambda_2\bigg) \\
	&= \tp\bigg((\bmu_1-\bmu_2)^T(\bx-\bmu_1) \leq \frac{1}{2}\twonorm{\bmu_1-\bmu_2}^2 + \log \lambda_2 - \log \lambda_1\bigg) \\
	&= \Phi\bigg(\frac{1}{2}\twonorm{\bmu_1 - \bmu_2} + \frac{\log \lambda_2 - \log \lambda_1}{\twonorm{\bmu_1 - \bmu_2}}\bigg).
\end{align}
Similarly, the misclassification error rate of $\varphi_{\blambda}^*$ for class 2 is
\begin{equation}
	R_2(\varphi^*_{\blambda}) = \tp_{X|Y=2}(\varphi^*_{\blambda}(X) \neq 2) = \Phi\bigg(\frac{1}{2}\twonorm{\bmu_1 - \bmu_2} + \frac{\log \lambda_1 - \log \lambda_2}{\twonorm{\bmu_1 - \bmu_2}}\bigg).
\end{equation}
Next, consider the CS classifier $\phi_{\blambda}^*$ indexed by $\blambda = (\lambda_1, \lambda_2)^T \in \mathbb{R}^2_+$ as
\begin{align}
	\phi_{\blambda}^*: \bx \mapsto &\argmax\bigg\{\frac{\lambda_1}{\pi_1^*}\tp_{Y|X=\bx}(Y=1), \frac{\lambda_2}{\pi_2^*}\tp_{Y|X=\bx}(Y=2), \frac{1}{\pi_3^*}\tp_{Y|X=\bx}(Y=3)\bigg\}  \\
	&= \argmax\{\lambda_1f_1(\bx), \lambda_2f_2(\bx), f_3(\bx)\}.
\end{align}

\noindent(\rom{1}) When $\twonorm{\bmu_1 - \bmu_2} > \Phi^{-1}(1-\alpha_1) + \Phi^{-1}(1-\alpha_2)$: There exists $\blambda = (\lambda_1, \lambda_2)^T \in \mathbb{R}^2_+$ such that 
\begin{equation}
	-\frac{1}{2}\twonorm{\bmu_1 - \bmu_2} - \Phi^{-1}(\alpha_1) < \frac{\log \lambda_1 - \log \lambda_2}{\twonorm{\bmu_1 - \bmu_2}} < \frac{1}{2}\twonorm{\bmu_1 - \bmu_2} + \Phi^{-1}(\alpha_2).
\end{equation}
Therefore $R_k(\varphi^*_{\blambda}) < \alpha_k$ for $k = 1:2$, which implies the feasibility of the NPMC problem. Consider $\blambda^{(m)}= (\lambda_1^{(m)}, \lambda_2^{(m)})^T$ with $\lambda_1^{(m)} = m\lambda_1$ and $\lambda_2^{(m)} = m\lambda_2$, for $m \geq 1$. Then $\varphi^*_{\blambda^{(m)}}(\bx)  = \varphi^*_{\blambda}(\bx)$, for all $\bx \in \mathbb{R}^p$, $k = 1:2$, and $\phi^*_{\blambda^{(m)}}(\bx) - \varphi^*_{\blambda^{(m)}}(\bx) \rightarrow 0$, $ \varphi^*_{\blambda^{(m)}}(\bx) \rightarrow \varphi^*_{\blambda}(\bx)$, as $m \rightarrow \infty$, for a.s. $\bx \sim \sum_{k=1}^3\pi_kf_k$. Hence by dominated convergence theorem, $R_k(\phi^*_{\blambda^{(m)}}) \rightarrow R_k(\varphi^*_{\blambda})$ for $k = 1:2$ when $m \rightarrow \infty$. Therefore, when $m > M$ where $M$ is a large constant, we must have $R_k(\phi^*_{\blambda^{(m)}}) \leq \alpha_k$ for $k = 1:2$. Then, the strong duality holds due to Theorem \ref{thm: strong duality}.(\rom{1}).

\noindent(\rom{2}) When $\twonorm{\bmu_1 - \bmu_2} = \Phi^{-1}(1-\alpha_1) + \Phi^{-1}(1-\alpha_2)$: There exists a unique $\blambda = (\lambda_1, \lambda_2)^T \in \mathbb{R}^2_+$ such that  
\begin{equation}
	-\frac{1}{2}\twonorm{\bmu_1 - \bmu_2} - \Phi^{-1}(\alpha_1) \leq \frac{\log \lambda_1 - \log \lambda_2}{\twonorm{\bmu_1 - \bmu_2}} \leq \frac{1}{2}\twonorm{\bmu_1 - \bmu_2} + \Phi^{-1}(\alpha_2),
\end{equation}
where the ``=" holds for both inequalities. Hence we have $R_k(\varphi^*_{\blambda}) = \alpha_k$ for $k = 1:2$ and this $\blambda = (\lambda_1, \lambda_2)^T \in \mathbb{R}^2_+$ is the only  $\blambda \in \mathbb{R}^2_+$ satisfying $R_k(\varphi^*_{\blambda}) \leq \alpha_k$ for $k = 1:2$. This proves the feasibility of the NPMC problem. On the other hand, if the strong duality holds, then by Theorem \ref{thm: strong duality}, there exists a finite $\tilde{\blambda} = (\tilde{\lambda}_1, \tilde{\lambda}_2)\in \mathbb{R}^2_+$ such that $R_k(\phi^*_{\tilde{\blambda}}) \leq \alpha_k$ for $k = 1:2$. 

Note that we must have $\tilde{\blambda} \neq \blambda$ because $R_k(\phi^*_{\blambda}) > R_k(\varphi^*_{\blambda}) = \alpha_k$ for $k = 1:2$ which violates the feasibility of $\phi^*_{\tilde{\blambda}}$. But then we have a contradiction because $R_k(\varphi^*_{\tilde{\blambda}}) \leq R_k(\phi^*_{\tilde{\blambda}}) \leq \alpha_k$ for $k = 1:2$, which violates the uniqueness of $\blambda$ such that $R_k(\varphi^*_{\blambda}) \leq \alpha_k$ for $k = 1:2$. Therefore, the strong duality must fail.

\noindent(\rom{3}) When $\twonorm{\bmu_1 - \bmu_2} < \Phi^{-1}(1-\alpha_1) + \Phi^{-1}(1-\alpha_2)$: The inequality
\begin{equation}
	-\frac{1}{2}\twonorm{\bmu_1 - \bmu_2} - \Phi^{-1}(\alpha_1) \leq \frac{\log \lambda_1 - \log \lambda_2}{\twonorm{\bmu_1 - \bmu_2}} \leq \frac{1}{2}\twonorm{\bmu_1 - \bmu_2} + \Phi^{-1}(\alpha_2)
\end{equation}
must fail. 

If $-\frac{1}{2}\twonorm{\bmu_1 - \bmu_2} - \Phi^{-1}(\alpha_1) < 0$, consider $\blambda = (\lambda_1, \lambda_2)^T \in \mathbb{R}^2_+$ such that 
\begin{equation}\label{eq: alpha less}
	0 \leq \lambda_1[\alpha_1 - \tp_{X|Y=1}(\phi^*_{\blambda}(X) \neq 1)] < \lambda_2[\tp_{X|Y=2}(\phi^*_{\blambda}(X) \neq 2) - \alpha_2].
\end{equation}
Such a $\blambda$ exists because we can consider $\blambda = (\lambda_1, \lambda_2)^T \in \mathbb{R}^2_+ \backslash \{(0, 0)^T\}$ with
\begin{equation}
	 -\frac{1}{2}\twonorm{\bmu_1 - \bmu_2} - \Phi^{-1}(\alpha_1), \frac{1}{2}\twonorm{\bmu_1 - \bmu_2} + \Phi^{-1}(\alpha_2) \leq \frac{\log \lambda_1 - \log \lambda_2}{\twonorm{\bmu_1 - \bmu_2}},
\end{equation}
and $-\frac{1}{2}\twonorm{\bmu_1 - \bmu_2} - \Phi^{-1}(\alpha_1)$ very close to $\frac{\log \lambda_1 - \log \lambda_2}{\twonorm{\bmu_1 - \bmu_2}}$. Since
\begin{equation}
	\bigg|\bigg[-\frac{1}{2}\twonorm{\bmu_1 - \bmu_2} - \Phi^{-1}(\alpha_1)\bigg] - \bigg[\frac{1}{2}\twonorm{\bmu_1 - \bmu_2} + \Phi^{-1}(\alpha_2)\bigg]\bigg| > 0,
\end{equation}
we have
\begin{equation}
	0 \leq \alpha_1 - \tp_{X|Y=1}(\phi^*_{\blambda}(X) \neq 1) < \tp_{X|Y=2}(\phi^*_{\blambda}(X) \neq 2) - \alpha_2.
\end{equation}
\eqref{eq: alpha less} follows with this $\blambda$ because the fact that $\frac{\log \lambda_1 - \log \lambda_2}{\twonorm{\bmu_1 - \bmu_2}}$ is very close to $-\frac{1}{2}\twonorm{\bmu_1 - \bmu_2} - \Phi^{-1}(\alpha_1) < 0$ implies $\lambda_1 < \lambda_2$. Going back to \eqref{eq: alpha less}, consider a sequence $\{\blambda^{(m)}\}_{m=1}^{\infty}$ with $\lambda^{(m)}_k = m\lambda_k$ for $m \geq 1$ and $k =1:2$. Then $\varphi^*_{\blambda^{(m)}}(\bx) = \varphi^*_{\blambda}(\bx)$ for all $\bx \in \mathbb{R}^p$ and $m \geq 1$. And
\begin{align}
	F_{\blambda^{(m)}}(\varphi_{\blambda^{(m)}}^*) &= \tp_{X|Y=3}(\varphi_{\blambda^{(m)}}^*(X) \neq 3) + \sum_{k=1}^2 \lambda_k^{(m)}[\tp_{X|Y=k}(\varphi_{\blambda^{(m)}}^*(X) \neq k) - \alpha_k] \\
	&=\tp_{X|Y=3}(\varphi_{\blambda^{(m)}}^*(X) \neq 3) + m\sum_{k=1}^2 \lambda_k[\tp_{X|Y=k}(\varphi_{\blambda^{(m)}}^*(X) \neq k) - \alpha_k] \\
	&\rightarrow +\infty,
\end{align}
as $m \rightarrow \infty$, due to \eqref{eq: alpha less}. This implies $G(\blambda^{(m)}) = F_{\blambda^{(m)}}(\phi_{\blambda^{(m)}}^*)\rightarrow +\infty$ as $m \rightarrow \infty$ because $R_k(\phi^*_{\blambda^{(m)}}) \geq R_k(\varphi^*_{\blambda^{(m)}})$ for $k = 1:2$ therefore $m\sum_{k=1}^2 \lambda_k[\tp_{X|Y=k}(\phi_{\blambda^{(m)}}^*(X) \neq k) - \alpha_k] \rightarrow  \infty$. Hence by weak duality, the NPMC problem is infeasible. And by definition, the strong duality holds. 

On the other hand, if $-\frac{1}{2}\twonorm{\bmu_1 - \bmu_2} - \Phi^{-1}(\alpha_1) \geq 0$, we must have $\frac{1}{2}\twonorm{\bmu_1 - \bmu_2} + \Phi^{-1}(\alpha_2) > -\frac{1}{2}\twonorm{\bmu_1 - \bmu_2} - \Phi^{-1}(\alpha_1) \geq 0$. By switching classes $1$ and $2$ in our analysis above, we can obtain the same conclusion that the NPMC problem is infeasible and the strong duality holds. 

\subsubsection{Proof of Lemma \ref{lemma: cs oracle classifier}}
We can easily write the cost function of any classifier $\phi$ as
\begin{align}
	\text{Cost}(\phi) &= \sum_{k=1}^K\pi^*_kc_k - \te[c_Y\mathds{1}(\phi(X) = Y)] \\
	&= \sum_{k=1}^K\pi^*_kc_k- \te_{X}\left\{\te_{Y|X}[c_Y\mathds{1}(\phi(X) = Y)]\right\} \\
	&= \sum_{k=1}^K\pi^*_kc_k- \te_{X}\left\{\sum_{k=1}^K[\mathds{1}(\phi(X) = k)\cdot c_k\tp_{Y|X}(Y=k)]\right\}.
\end{align}
By the last expression and the definition of $\phi^*$, we have $\text{Cost}(\phi) \geq \text{Cost}(\phi^*)$ for any $\phi$.

\subsubsection{Proof of Lemma \ref{lemma: cs oracle classifier confusion matrix}}
Similar to the proof of Lemma \ref{lemma: cs oracle classifier}, let's first simplify the cost function of any classifier $\phi$ as
\begin{align}
	\text{Cost}(\phi) &= \te\left[\sum_{r \neq Y}c_{Y,r}\mathds{1}(\phi(X) = r)\right] \\
	&= \te_{X}\left\{\te_{Y|X}\left[\sum_{r \neq Y}c_{Y, r}\mathds{1}(\phi(X) = r)\right]\right\} \\
	&= \te_{X}\left\{\sum_{k=1}^K\sum_{r \neq k}\left[\mathds{1}(\phi(X) = r)\cdot c_{kr}\tp_{Y|X}(Y=k)\right]\right\}.
\end{align}
Therefore by the definition of $\phi^*$, we have $\text{Cost}(\phi) \geq \text{Cost}(\phi^*)$ for any $\phi$.

\subsubsection{Proof of Lemma \ref{lem: npmc-max}}
(\rom{1}) If $\sum_{k \notin \mathcal{A}} \lambda_k^* > 1$, we have $G(\blambda^*) = -\sum_{k=1}^K\lambda^*_k \tp_{X|Y = k}(\phi^*_{\blambda^*}(X) = k) + \sum_{k \in \mathcal{A}}\lambda^*_k(1-\alpha_k) + 1$. Consider $\tilde{\blambda} = \{\tilde{\lambda}_k\}_{k=1}^K$ with $\tilde{\lambda}_k = \frac{\lambda^*_k}{\sum_{k \notin \mathcal{A}} \lambda_k^*} < \lambda_k^*$ for all $k = 1:K$. Then it can be seen that $\phi^*_{\blambda^*}(\bx) = \phi^*_{\tilde{\blambda}}(\bx)$ for all $\bx \in \mathbb{R}^p$. By weak duality, since NPMC-max problem is feasible, we must have $G(\blambda^*) \leq 1$, which implies that
\begin{equation}
	-\sum_{k=1}^K\lambda^*_k \tp_{X|Y = k}(\phi^*_{\blambda^*}(X) = k) + \sum_{k \in \mathcal{A}}\lambda^*_k(1-\alpha_k) \leq 0.
\end{equation}
Hence 
\begin{equation}
	-\sum_{k=1}^K\lambda^*_k \tp_{X|Y = k}(\phi^*_{\blambda^*}(X) = k) + \sum_{k \in \mathcal{A}}\lambda^*_k(1-\alpha_k) \leq -\sum_{k=1}^K\tilde{\lambda}_k \tp_{X|Y = k}(\phi^*_{\tilde{\blambda}}(X) = k) + \sum_{k \in \mathcal{A}}\tilde{\lambda}_k(1-\alpha_k),
\end{equation}
which leads to $G(\blambda^*) \leq G(\tilde{\blambda})$.

If $0 < \sum_{k \notin \mathcal{A}} \lambda_k^* < 1$, we have $G(\blambda^*) = -\sum_{k=1}^K\lambda^*_k \tp_{X|Y = k}(\phi^*_{\blambda^*}(X) = k) + \sum_{k \in \mathcal{A}}\lambda^*_k(1-\alpha_k) + \sum_{k \notin \mathcal{A}}\lambda_k^*$. Consider $\tilde{\blambda} = \{\tilde{\lambda}_k\}_{k=1}^K$ with $\tilde{\lambda}_k = \frac{\lambda^*_k}{\sum_{k \notin \mathcal{A}} \lambda_k^*} > \lambda_k^*$ for all $k = 1:K$. We also have $\phi^*_{\blambda^*}(\bx) = \phi^*_{\tilde{\blambda}}(\bx)$ for all $\bx \in \mathbb{R}^p$.  By strong duality, since NPMC-max problem is feasible, then $G(\blambda^*) \geq 0$, which implies that
\begin{align}
	&-\sum_{k=1}^K\lambda^*_k \tp_{X|Y = k}(\phi^*_{\blambda^*}(X) = k) + \sum_{k \in \mathcal{A}}\lambda^*_k(1-\alpha_k) + \sum_{k \notin \mathcal{A}}\lambda_k^* \\
	&\leq -\sum_{k=1}^K\tilde{\lambda}_k \tp_{X|Y = k}(\phi^*_{\tilde{\blambda}}(X) = k) + \sum_{k \in \mathcal{A}}\tilde{\lambda}_k(1-\alpha_k) + \sum_{k \notin \mathcal{A}}\tilde{\lambda}_k,
\end{align}
i.e. $G(\blambda^*) \leq G(\tilde{\blambda})$. 

If $\sum_{k \notin \mathcal{A}} \lambda_k^* = 0$, then $G(\blambda^*) = -\sum_{k \in \mathcal{A}}\lambda^*_k \tp_{X|Y = k}(\phi^*_{\blambda^*}(X) = k) + \sum_{k \in \mathcal{A}}\lambda^*_k(1-\alpha_k) \geq 0$ by strong duality. Therefore we can construct $\tilde{\blambda}$ with $\tilde{\blambda} = \{\tilde{\lambda}_k\}_{k=1}^K$, $\tilde{\lambda}_k = 2\lambda^*_k > \lambda_k^*$ for $k \in \mathcal{A}$, and $\tilde{\lambda}_k = 0$ for $k \notin \mathcal{A}$. It is easy to see that $\phi^*_{\blambda^*}(\bx) = \phi^*_{\tilde{\blambda}}(\bx)$ for all $\bx \in \mathbb{R}^p$, hence $G(\tilde{\blambda}) > G(\blambda^*)$ if $G(\blambda^*) > 0$, which is contradicted. If $G(\blambda^*) = 0$, then by strong duality, $\phi^*_{\blambda^*}$ defined in \eqref{eq: cs classifier npmc-max} is feasible and $\max_{k \notin \mathcal{A}}\tp_{X|Y = k}(\phi^*_{\blambda^*}(X) \neq k) = G(\blambda^*) = 0$. However, by the definition of $\phi^*_{\blambda^*}$ in \eqref{eq: cs classifier npmc-max}, since $\lambda_k^* = 0$ for all $k \notin \mathcal{A}$, we must have $\tp_{X|Y = k}(\phi^*_{\blambda^*}(X) \neq k) = 1$ for some $k \notin \mathcal{A}$, which is again contradited.

Therefore (\rom{1}) holds.

\noindent(\rom{2}) By strong duality, there exists a sequence $\{\blambda^{(m)}\}_{m=1}^{\infty}$ such that $G(\blambda^{(m)}) \rightarrow +\infty$. WLOG, assume $\lambda_k^{(m)} \leq C < \infty$ and $\sum_{k\notin \mathcal{A}}\lambda^{(m)}_k > 0$, for all $k \notin \mathcal{A}$ and $m \geq 1$ (otherwise we can construct such a sequence from $\{\blambda^{(m)}\}_{m=1}^{\infty}$). Hence 
\begin{equation}
	-\sum_{k \in \mathcal{A}}\lambda_k^{(m)} \tp_{X|Y = k}(\phi^*_{\blambda^{(m)}}(X) = k) + \sum_{k \in \mathcal{A}}\lambda^{(m)}_k(1-\alpha_k) \rightarrow +\infty,
\end{equation}
as $m \rightarrow \infty$. Consider a new sequence $\{\tilde{\blambda}^{(m)}\}_{m=1}^{\infty}$ with $\tilde{\lambda}^{(m)}_k = \frac{\lambda^{(m)}_k}{\sum_{k \notin \mathcal{A}}\lambda^{(m)}_k} \geq \frac{1}{C(K-|\mathcal{A}|)}\lambda_k^{(m)}$ for all $k \in \mathcal{A}$. It is easy to see that $\phi^*_{\blambda^{(m)}}(\bx) = \phi^*_{\tilde{\blambda}^{(m)}}(\bx)$ for all $\bx \in \mathbb{R}^p$ and $m \geq 1$. Therefore, we have
\begin{align}
	&-\sum_{k \in \mathcal{A}}\tilde{\lambda}_k^{(m)} \tp_{X|Y = k}(\phi^*_{\tilde{\blambda}^{(m)}}(X) = k) + \sum_{k \in \mathcal{A}}\tilde{\lambda}^{(m)}_k(1-\alpha_k) \\
	&\geq \frac{1}{C(K-|\mathcal{A}|)}\bigg[-\sum_{k \in \mathcal{A}}\lambda_k^{(m)} \tp_{X|Y = k}(\phi^*_{\blambda^{(m)}}(X) = k) + \sum_{k \in \mathcal{A}}\lambda^{(m)}_k(1-\alpha_k)\bigg] \\
	&\rightarrow +\infty,
\end{align}
as $m \rightarrow \infty$. Hence $G(\tilde{\blambda}^{(m)}) \rightarrow +\infty$ as $m \rightarrow \infty$ and $\{\tilde{\blambda}^{(m)}\}_{m=1}^{\infty}$ satisfies $\sum_{k \notin \mathcal{A}}\tilde{\lambda}^{(m)}_k = 1$ for all $m \geq 1$.

\subsubsection{Proof of Lemma \ref{lem: finite bound}}
\begin{lemma}[Corollary 1 in \citealpapp{maurer2016vector}]\label{lem: vec contraction}
	Suppose $\{\epsilon_{ik}\}_{i\in [n], k \in [K]}$ and $\{\epsilon_i\}_{i=1}^n$ are independent Rademacher variables. Let $\mathcal{F}$ be a class of functions $f: \mathbb{R}^d \rightarrow  \mathcal{S}\subseteq \mathbb{R}^K$ and $h: \mathcal{S} \rightarrow\mathbb{R}$ is $L$-Lipschitz under $\ell_2$-norm, i.e., $\norm{h(\bm{y}) - h(\bm{y}')} \leq L\twonorm{\bm{y} - \bm{y}'}$, where $\bm{y} = (y_1,\ldots,y_K)^T$, $\bm{y}' = (y_1',\ldots,y_K')^T \in \mathcal{S}$. Then
	\begin{equation}
		\te \sup_{f \in \mathcal{F}} \sum_{i=1}^n \epsilon_i h(f(x_i)) \leq \sqrt{2}L\te \sup_{f \in \mathcal{F}} \sum_{i=1}^n\sum_{k=1}^K \epsilon_{ik}f_k(x_i), 
	\end{equation}
	where $f_k(x_i)$ is the $k$-th component of $f(x_i) \in \mathcal{S}\subseteq \mathbb{R}^K$.
\end{lemma}

First, we prove that for any bounded sets $\Lambda \subseteq \mathbb{R}^{|\mathcal{A}|}_+$, we have 
\begin{align}
	&\tp\bigg(\sup_{\blambda \in \Lambda}\norm{\hL(\blambda, \hphi_{\blambda})-L(\blambda, \hphi_{\blambda})} > \tau + CC_{\textup{Rad}} +  C'\max_{k=1:K}\te|\widehat{\tp}_{Y|X}(Y = k) - \tp_{Y|X}(Y = k)|,\\
	&\quad \bigcap_{k=1}^K\{|\hpi_k - \pi^*_k|\leq \pi^*_k/2\}\bigg) \lesssim \exp\{-n\tau^2\}, 
\end{align}
for any $\tau > 0$. Denote the estimate $\widehat{\tp}_{Y|X=\bx}(Y = k) = \widehat{\tp}_{Y|X=\bx}(Y = k; \hbbeta)$, $\tilde{\phi}(\bx; \bbeta, \blambda, \bpi) = \argmax_k\{c_k(\blambda, \bpi)\widehat{\tp}_{Y|X=\bx}(Y=k;\bbeta)\}$, and $\Pi = \{\bpi \in S^{K-1}: \pi_k \geq \pi^*_k/2 \textup{ for all }k\}$, where $S^{K-1}$ is the $\ell_1$ unit sphere in $\mathbb{R}^K$. Note that 
\begin{align}
	\sup_{\blambda \in \Lambda}\norm{\hL(\blambda, \hphi_{\blambda})-L(\blambda, \hphi_{\blambda})} &\leq \sup_{\blambda \in \Lambda} \sup_{\bpi \in \Pi} \Bigg|\frac{1}{n}\sum_{i=1}^n c_{\tilde{\phi}(\bx_i)}(\blambda,\bpi)\widehat{\tp}_{Y|X=\bx_i}(Y = \tilde{\phi}(\bx_i;\bbeta, \blambda, \bpi);\hbbeta)  \\
	&\quad\quad\quad\quad\quad -\te\left[c_{\tilde{\phi}(X)}(\blambda,\bpi)\tp_{Y|X}(Y = \tilde{\phi}(X;\bbeta, \blambda, \bpi))\right]\Bigg| \\
	&\leq \sup_{\blambda \in \Lambda} \sup_{\bpi \in \Pi} \Bigg|\frac{1}{n}\sum_{i=1}^n c_{\tilde{\phi}(\bx_i)}(\blambda,\bpi)\widehat{\tp}_{Y|X=\bx_i}(Y = \tilde{\phi}(\bx_i;\bbeta, \blambda, \bpi);\hbbeta)  \\
	&\quad\quad\quad\quad\quad -\te\left[c_{\tilde{\phi}(X)}(\blambda,\bpi)\widehat{\tp}_{Y|X}(Y = \tilde{\phi}(X;\bbeta, \blambda, \bpi); \hbbeta)\right]\Bigg| \\
	&\quad + \sup_{\blambda \in \Lambda} \sup_{\bpi \in \Pi}\Bigg|\te\left[c_{\tilde{\phi}(X)}(\blambda,\bpi)\widehat{\tp}_{Y|X}(Y = \tilde{\phi}(X;\bbeta, \blambda, \bpi); \hbbeta)\right] \\
	&\quad\quad\quad\quad\quad\quad -\te\left[c_{\tilde{\phi}(X)}(\blambda,\bpi)\tp_{Y|X}(Y = \tilde{\phi}(X;\bbeta, \blambda, \bpi))\right]\Bigg|\\
	&\leq \sup_{\blambda \in \Lambda}\sup_{\bbeta \in \mathcal{B}} \sup_{\bpi \in \Pi} \Bigg|\frac{1}{n}\sum_{i=1}^n c_{\tilde{\phi}(\bx_i)}(\blambda,\bpi)\widehat{\tp}_{Y|X=\bx_i}(Y = \tilde{\phi}(\bx_i;\bbeta, \blambda, \bpi);\bbeta)  \\
	&\quad\quad\quad\quad\quad\quad\quad -\te\left[c_{\tilde{\phi}(X)}(\blambda,\bpi)\widehat{\tp}_{Y|X}(Y = \tilde{\phi}(X;\bbeta, \blambda, \bpi);\bbeta)\right]\Bigg| \\
	& + C'\max_{k=1:K}\te|\widehat{\tp}_{Y|X}(Y = k) - \tp_{Y|X}(Y = k)|.
\end{align}
Let 
\begin{equation}
	g(\bx;\bbeta, \blambda, \bpi) = \max_k\{c_k(\blambda, \bpi)\widehat{\tp}_{Y|X=\bx}(Y=k;\bbeta)\} - \te[\max_k\{c_k(\blambda, \bpi)\widehat{\tp}_{Y|X=\bx}(Y=k;\bbeta)\}].
\end{equation}
Then $\te g(X) = 0$ and $\sup_{\blambda \in \Lambda}\sup_{\bbeta \in \mathcal{B}} \sup_{\bpi \in \Pi}\infnorm{g(X)} \leq C$ a.s. with some large constant $C > 0$. It suffices to upper bound $U(\{X_i\}_{i=1}^n) \coloneqq \sup_{\blambda \in \Lambda}\sup_{\bbeta \in \mathcal{B}} \sup_{\bpi \in \Pi}|n^{-1}\sum_{i=1}^n g(X_i;\bbeta, \blambda, \bpi)|$. 

Consider $\{X_i'\}_{i=1}^n$ which differs from $\{X_i\}_{i=1}^n$ at only one single entry $i$. Then $|U(\{X_i\}_{i=1}^n) - U(\{X_i'\}_{i=1}^n)| \lesssim 1/n$ by boundedness of $\Lambda$, $\mathcal{B}$, and $\Pi$. Applying bounded difference inequality (Corollary 2.21 in \citealpapp{wainwright2019high}),
\begin{equation}
	\tp(U - \te U \geq \tau) \lesssim \exp\{-Cn\tau^2\}.
\end{equation}
It remains to bound $\te U$. By standard symmetrization argument, with $\bm{\epsilon} = (\epsilon_1, \ldots, \epsilon_n)^T$ independent Rademacher variables, we have
\begin{align}
	\te U (\{X_i\}_{i=1}^n) &\lesssim \te_{\bm{\epsilon}}\te_X \sup_{\blambda \in \Lambda}\sup_{\bbeta \in \mathcal{B}} \sup_{\bpi \in \Pi} \Bigg|\frac{1}{n}\sum_{i=1}^n \epsilon_i \max_k\{c_k(\blambda, \bpi)\widehat{\tp}_{Y|X=\bx}(Y=k;\bbeta)\}\Bigg| \\
	&\lesssim \te_{\bm{\epsilon}}\te_X \sup_{\infnorm{\bm{a}} \leq C}\sup_{\bbeta \in \mathcal{B}} \Bigg|\frac{1}{n}\sum_{i=1}^n \epsilon_i \max_k\{a_k\widehat{\tp}_{Y|X=\bx}(Y=k;\bbeta)\}\Bigg|, \label{eq: proof lemma 7 eq 1}
\end{align}
where $\bm{a} = (a_1, \ldots, a_K)^T$ and $C$ is some constant. It is easy to see that $\max_{k=1:K}x_k$ is a 1-Lipschitz function of $\bx = (x_1, \ldots, x_K)^T$ w.r.t. $\ell_2$-norm. Then applying the vectorized Rademacher contraction inequality (Lemma \ref{lem: vec contraction}), we can further bound the RHS of \eqref{eq: proof lemma 7 eq 1} as
\begin{align}
	&\te_{\bm{\epsilon}}\te_X \sup_{\infnorm{\bm{a}} \leq C}\sup_{\bbeta \in \mathcal{B}} \Bigg|\frac{1}{n}\sum_{i=1}^n \epsilon_i \max_k\{a_k\widehat{\tp}_{Y|X=\bx_i}(Y=k;\bbeta)\}\Bigg| \\
	&\lesssim \te_{\bm{\epsilon}'}\te_X \sup_{\infnorm{\bm{a}} \leq C}\sup_{\bbeta \in \mathcal{B}}\Bigg|\frac{1}{n}\sum_{k=1}^K\sum_{i=1}^n \epsilon_{ik}'a_k\widehat{\tp}_{Y|X=\bx_i}(Y=k;\bbeta)\Bigg| \\
	&\lesssim \sum_{k=1}^K\te_{\bm{\epsilon}'}\te_X \sup_{\bbeta \in \mathcal{B}}\Bigg|\frac{1}{n}\sum_{i=1}^n \epsilon_{ik}'\widehat{\tp}_{Y|X=\bx_i}(Y=k;\bbeta)\Bigg| \\
	&\lesssim \max_{k=1:K}\te_{\bm{\epsilon}}\te_X \sup_{\bbeta \in \mathcal{B}}\Bigg|\frac{1}{n}\sum_{i=1}^n \epsilon_{i}\widehat{\tp}_{Y|X=\bx_i}(Y=k;\bbeta)\Bigg| \\
	&= C_{\textup{Rad}}(n),
\end{align}
where $\bm{\epsilon}' = \{\epsilon_{ik}\}_{i=1:n, k = 1:K}$ contains independent Rademacher variables. Putting all the pieces together, we have 
\begin{align}
	&\tp\bigg(\sup_{\blambda \in \Lambda}\norm{\hL(\blambda, \hphi_{\blambda})-L(\blambda, \hphi_{\blambda})} > \tau + CC_{\textup{Rad}} +  C'\max_{k=1:K}\te|\widehat{\tp}_{Y|X}(Y = k) - \tp_{Y|X}(Y = k)|,\\
	&\quad \bigcap_{k=1}^K\{|\hpi_k - \pi^*_k|\leq \pi^*_k/2\}\bigg) \lesssim \exp\{-n\tau^2\}, 
\end{align}
for any $\tau > 0$. Note that by Hoeffding's inequality and union bound,
\begin{equation}
	\tp\bigg(\bigcup_{k=1}^K\{|\hpi_k - \pi^*_k|> \pi^*_k/2\}\bigg) \lesssim \exp\{-Cn\}.
\end{equation}
Therefore, for any $\tau > 0$,
\begin{equation}\label{eq: lemma 7 proof eq 2}
	\tp\bigg(\sup_{\blambda \in \Lambda}\norm{\hL(\blambda, \hphi_{\blambda})-L(\blambda, \hphi_{\blambda})} > \tau + CC_{\textup{Rad}} +  C'\max_{k=1:K}\te|\widehat{\tp}_{Y|X}(Y = k) - \tp_{Y|X}(Y = k)|\bigg) \leq  \exp\{-n(\tau^2 \wedge 1)\}.
\end{equation}

Next, we will show that
\begin{equation}\label{eq: lemma finite bound ineq 2}
	\tp\left(\sup_{\blambda \in \Lambda}\norm{L(\blambda, \hphi_{\blambda})-L(\blambda, \phi^*)} > \tau \right) \lesssim \tau^{-1}\max_{k}\te \norma{\widehat{\tp}_{Y|X}(Y=k) - \tp_{Y|X}(Y=k)} + \exp\{-Cn\tau^2\}.
\end{equation}
By the proof of Lemma \ref{lem: uniform bdd 2}, combined with Markov inequality and union bounds,
\begin{align}
	\tp\left(\sup_{\blambda \in \Lambda}\norm{L(\blambda, \hphi_{\blambda})-L(\blambda, \phi^*)} > \tau \right) &\leq \sum_{k=1}^K\tp\left(\te_X\norm{\widehat{\tp}_{Y|X}(Y=k) - \tp_{Y|X}(Y=k)}> \frac{\tau}{2CK}\right) \\
	&\quad + \sum_{k=1}^K \tp\left(|\hpi_k - \pi_k|>\frac{\tau}{2CK}\right) \\
	&\lesssim \tau^{-1}\max_k\te\norm{\widehat{\tp}_{Y|X}(Y=k) - \tp_{Y|X}(Y=k)} + \exp\{-Cn\tau^2\}. \label{eq: finite bound lemma step 2}
\end{align}

Finally, combining \eqref{eq: lemma 7 proof eq 2} and \eqref{eq: lemma finite bound ineq 2}, we get the desired conclusion, which completes the proof of Lemma \ref{lem: finite bound}.

\subsubsection{Proof of Lemma \ref{lem: uniform bdd 1}}
Denote $\btheta = (\bbeta, \blambda, \bpi)^T$, $U(\bx; \btheta) = c_{\tilde{\phi}(\bx)}\widehat{\tp}_{Y|X=\bx}(Y = \tilde{\phi}(\bx;\btheta);\bbeta) - \te\left[c_{\tilde{\phi}(X)}\widehat{\tp}_{Y|X}(Y = \tilde{\phi}(X;\btheta);\bbeta)\right]$, and $\tilde{\phi}(\bx;\bbeta, \blambda, \bpi) = \argmax_k \{c_k(\blambda, \bpi)\widehat{\tp}_{Y|X=\bx}(Y=k;\bbeta)\}$. First, we prove that for any compact sets $\Lambda \subseteq \mathbb{R}^{|\mathcal{A}|}_+$, $\mathcal{B} \subseteq \mathbb{R}^p$ and $\Pi \subseteq (0,1)^K$, it holds
\begin{equation}\label{eq: conv p lemma 1}
  \lim_{n \rightarrow \infty}\sup_{\blambda \in \Lambda}\sup_{\bbeta \in \mathcal{B}} \sup_{\bpi \in \Pi} \norma{\frac{1}{n}\sum_{i=1}^n U(\bx_i; \btheta)} = 0, \quad a.s.,
\end{equation}
We follow the proof idea of Theorem 5.14 in \citeapp{van2000asymptotic}, which was first stated in \citeapp{wald1949note}. We first check the following two conditions:
\begin{enumerate}[(i)]
	\item $U(\bx; \btheta)$ is a continuous function of $\btheta = (\bbeta, \blambda, \bpi)$ for a.s. $\bx$ w.r.t. the distribution of $X$.
	\item There is a function $m(\bx)$ satisfying 
		\begin{equation}
			\sup_{\blambda \in \Lambda}\sup_{\bbeta \in \mathcal{B}} \sup_{\bpi \in \Pi}|c_{\tilde{\phi}(\bx)}(\blambda,\bpi)\widehat{\tp}_{Y|X=\bx}(Y = \tilde{\phi}(\bx;\bbeta, \blambda, \bpi);\bbeta)| \leq m(\bx), \te m(\bx) < \infty.
		\end{equation} 
\end{enumerate}
First, (\rom{2}) is trivial according to the fact that $\widehat{\tp}_{Y|X=\bx}(Y = \tilde{\phi}(\bx;\bbeta, \blambda, \bpi);\bbeta)$ is bounded. For (\rom{1}), note that $U(\bx; \btheta)$ can be written as a maximum of $K$ continuous functions of $\btheta$ by the definition of $\tilde{\phi}$, then the continuity of the maximum follows.

Define $W(\bx; r, \btheta) = \sup_{\btheta': \twonorm{\btheta'-\btheta} \leq r}U(\bx; \btheta')$. By the continuity of $U(\bx; \btheta)$ w.r.t. $\btheta$, $W(\bx; r, \btheta)$ is continuous w.r.t. $r$. In addition, by the dominated convergence theorem,
\begin{equation}
	\lim_{r \rightarrow 0}\te[W(X; r, \btheta)] = \te \left[\lim_{r \rightarrow 0}W(X; r, \btheta)\right] = 0.
\end{equation}
Then for any $\btheta \in \mathcal{B}\otimes\Lambda\otimes\Pi$, any $\epsilon > 0$, $\exists r_{\epsilon}(\btheta)$, such that $\te[W(X; r_{\epsilon}(\btheta), \btheta)] \leq \epsilon$. Because $\mathcal{B}\otimes\Lambda\otimes\Pi$ is compact, there exists a finite subcover of $\bigcup_{\btheta \in \mathcal{B}\otimes\Lambda\otimes\Pi}\mathcal{B}_{r_{\epsilon}(\btheta)}(\btheta)$, which we denoted as $\bigcup_{l=1}^L\mathcal{B}_{r_l}(\btheta_l)$. Then
\begin{equation}
	\sup_{\btheta \in \mathcal{B}\otimes\Lambda\otimes\Pi} \frac{1}{n}\sum_{i=1}^n U(\bx_i;\btheta) \leq \sup_{l=1,\ldots,L}\frac{1}{n}\sum_{i=1}^n W(\bx_i;r_l,\btheta_l) \overset{a.s.}{\rightarrow} \sup_{l=1,\ldots,L}\te [W(X;r_l,\btheta_l)] \leq \epsilon.
\end{equation}
Constructing a vanishing series $\{\epsilon_r\}_{r=1}^{\infty} \rightarrow 0$ leads to 
\begin{equation}\label{eq: sup u}
	\tp\left(\limsup_{n \rightarrow \infty}\sup_{\btheta \in \mathcal{B}\otimes\Lambda\otimes\Pi}\frac{1}{n}\sum_{i=1}^n U(\bx_i;\btheta) \leq 0\right) =\lim_{r\rightarrow \infty}\tp\left(\limsup_{n \rightarrow \infty}\sup_{\btheta \in \mathcal{B}\otimes\Lambda\otimes\Pi}\frac{1}{n}\sum_{i=1}^n U(\bx_i;\btheta) \leq \epsilon_r\right) =1.
\end{equation}
On the other hand, we can show $\tp\left(\liminf_{n \rightarrow \infty}\inf_{\btheta \in \mathcal{B}\otimes\Lambda\otimes\Pi}\frac{1}{n}\sum_{i=1}^n U(\bx_i;\btheta) \geq 0\right) = 1$ in the same way, which combines with \eqref{eq: sup u} implies \eqref{eq: conv p lemma 1}. Therefore, by plugging $\bbeta = \hbbeta$ and $\bpi = \hbpi$ into \eqref{eq: conv p lemma 1}, we have
\begin{equation}\label{eq: lemma xx proof eq 1}
	\lim_{n\rightarrow \infty}\norma{\hL(\blambda, \hphi_{\blambda}) - \te_X\left[c_{\hphi_{\blambda}(X)}(\blambda,\hbpi)\widehat{\tp}_{Y|X}(Y = \hphi_{\blambda}(X)) \right]} = 0, a.s..
\end{equation}

Next, we want to show
\begin{equation}\label{eq: lemma xx proof eq 2}
	\limsup_{n\rightarrow \infty}\sup_{\blambda \in \Lambda} \norma{\te_X\left[c_{\hphi_{\blambda}(X)}(\blambda,\hbpi)\widehat{\tp}_{Y|X}(Y = \hphi_{\blambda}(X)) - c_{\hphi_{\blambda}(X)}(\blambda,\bpi)\tp_{Y|X}(Y = \hphi_{\blambda}(X))\right]} = 0, a.s..
\end{equation}
Note that the left-hand side is no larger than
\begin{equation}
	\limsup_{n\rightarrow \infty}\max_k \te_X\left[\norma{\widehat{\tp}_{Y|X}(Y = k)-\tp_{Y|X}(Y = k)}\right]\cdot \max_k c_k + 2\limsup_{n\rightarrow \infty}\sup_{\blambda \in \Lambda}\max_k \norm{c_k(\blambda,\hbpi) - c_k(\blambda,\bpi)} = 0,
\end{equation}
a.s., which is derived by Assumption \ref{asmp: posterior distance as} with dominated convergence theorem combined with the strong consistency of $\hbpi$. Combining \eqref{eq: lemma xx proof eq 1} and \eqref{eq: lemma xx proof eq 2}, we finish the proof of Lemma \ref{lem: uniform bdd 1}.

\subsubsection{Proof of Lemma \ref{lem: uniform bdd 2}}
\begin{align}
	&\lim_{n\rightarrow \infty}\sup_{\blambda \in \Lambda}\norm{L(\blambda, \hphi_{\blambda}) - L(\blambda, \phi^*)} \\
	&= \lim_{n\rightarrow \infty}\sup_{\blambda \in \Lambda}\norma{\te_X\left[c_{\phi^*_{\blambda}(X)}(\blambda,\bpi^*)\tp_{Y|X}(Y = \phi^*_{\blambda}(X))-c_{\hphi_{\blambda}(X)}(\blambda,\bpi^*)\tp_{Y|X}(Y = \hat{\phi}_{\blambda}(X))\right]} \\
	&\leq \lim_{n\rightarrow \infty}\sum_{k=1}^K \left[\te_X\norma{\widehat{\tp}_{Y|X}(Y=k) - \tp_{Y|X}(Y=k)}\cdot \sup_{\blambda \in \Lambda}\max_k c_k(\blambda,\bpi^*) + \sup_{\blambda \in \Lambda}\norm{c_k(\blambda,\hbpi)-c_k(\blambda,\bpi^*)}\right]\\
	&= 0,
\end{align}
a.s., where the last equation holds because of Assumption \ref{asmp: posterior distance as} and the strong consistency of $\hbpi$. It suffices to verify the intermediate inequality. For any $X = \bx$ and $\blambda \in \Lambda$, denote $\hat{k} = \hat{k}(\bx) = \hphi_{\blambda}(\bx)$, $k^* = k^*(\bx) = \phi^*_{\blambda}(\bx)$. Then by the definition of $\hphi_{\blambda}$ and $\phi^*_{\blambda}$,
\begin{align}
	0 &\leq c_{\phi^*_{\blambda}(\bx)}(\blambda,\bpi^*)\tp_{Y|X=\bx}(Y = \phi^*_{\blambda}(\bx)) - c_{\hphi_{\blambda}(\bx)}(\blambda,\bpi^*)\tp_{Y|X=\bx}(Y = \hat{\phi}_{\blambda}(\bx)) \\
	&\leq [\tp_{Y|X=\bx}(Y = k^*)c_{k^*}(\blambda,\bpi^*) - \widehat{\tp}_{Y|X=\bx}(Y = k^*)c_{k^*}(\blambda,\hbpi)] \\
	&\quad+ [\widehat{\tp}_{Y|X=\bx}(Y = k^*)c_{k^*}(\blambda,\hbpi) - \widehat{\tp}_{Y|X=\bx}(Y = \hat{k})c_{\hat{k}}(\blambda,\hbpi)] \\
	&\quad + [\widehat{\tp}_{Y|X=\bx}(Y = \hat{k})c_{\hat{k}}(\blambda,\hbpi) - \tp_{Y|X=\bx}(Y = \hat{k})c_{\hat{k}}(\blambda,\bpi^*)] \\
	&\leq 2\sum_{k=1}^K \norma{\widehat{\tp}_{Y|X}(Y=k)c_k(\blambda,\hbpi) - \tp_{Y|X}(Y=k)c_k(\blambda,\bpi^*)} \\
	&\leq 2\sum_{k=1}^K \norma{\widehat{\tp}_{Y|X}(Y=k) - \tp_{Y|X}(Y=k)}\cdot \max_k c_k(\blambda,\bpi^*) + 2\norm{c_k(\blambda,\hbpi)-c_k(\blambda,\bpi^*)}, \\ \label{eq: an interesting argument}
\end{align}
where we used the fact that $\tp_{Y|X=\bx}(Y = k^*)c_{k^*}(\blambda,\bpi^*) - \tp_{Y|X=\bx}(Y = \hat{k})c_{\hat{k}}(\blambda,\bpi^*) \geq 0$ and $\widehat{\tp}_{Y|X=\bx}(Y = k^*)c_{k^*}(\blambda,\hbpi) - \widehat{\tp}_{Y|X=\bx}(Y = \hat{k})c_{\hat{k}}(\blambda,\hbpi) \leq 0$. Taking the supremum w.r.t. $\blambda$ and the limit $n \rightarrow \infty$ leads to the desired conclusion.

\subsubsection{Proof of Lemma \ref{lem: uniform conv algorithm 2}}
Let's fix $\mathcal{D}_2$ and $n_k$ first. Denote $\widehat{R}_k(\hphi_{\blambda}) = n^{-1}_k \sum_{i=1}^{n_k}\mathds{1}(\hphi(\bx_i^{(k)}) = k) = n^{-1}_k \sum_{i=1}^{n_k}\mathds{1}(\hat{g}^{(k)}_{\blambda}(\bx_i^{(k)}) > 0)$, where $\hat{g}^{(k)}_{\blambda}(\bx) = c_k(\blambda,\hbpi)\widehat{\tp}_{Y|X=\bx}(Y=k) - \max_{j \neq k}\big[c_j(\blambda,\hbpi)\widehat{\tp}_{Y|X=\bx}(Y=j)\big]$. Given any $\mathcal{D}_2$, we claim that the VC dimension of $\mathcal{A}_k = \{\mathds{1}(\hat{g}^{(k)}_{\blambda}(\bx) > 0): \blambda \succeq \bm{0} \}$ is finite for any $k$. 

The proof is straightforward. Recall that given $\mathcal{D}_2$ and $\blambda$, 
\begin{equation}
	c_k(\blambda, \hbpi) = \begin{cases}
		w_k/\hpi_k, \quad &k \notin \mathcal{A};\\
		(w_k+\lambda_k)/\hpi_k, \quad &k \in \mathcal{A}.
	\end{cases}
\end{equation}
For $k \in \mathcal{A}$, $\hat{g}^{(k)}_{\blambda}(\bx) = (w_k+\lambda_k)/\hpi_k\cdot \widehat{\tp}_{Y|X=\bx}(Y=k) - \max\big\{\max_{j \in \mathcal{A}\backslash\{k\}}\big[(w_j+\lambda_j)/\hpi_j\cdot\widehat{\tp}_{Y|X=\bx}(Y=j)\big], \max_{j\notin \mathcal{A}}\big[w_j/\hpi_j\cdot \widehat{\tp}_{Y|X=\bx}(Y=j)\big]\big\}$. Note that 
\begin{align}
	\{\bx: \hat{g}^{(k)}_{\blambda}(\bx) > 0\} &= \bigcap_{j \in \mathcal{A}\backslash\{k\}}\{\bx: (w_j+\lambda_j)/\hpi_j\cdot\widehat{\tp}_{Y|X=\bx}(Y=j) < (w_k+\lambda_k)/\hpi_k\cdot \widehat{\tp}_{Y|X=\bx}(Y=k)\} \\
	&\quad \bigcap \bigcap_{j \notin \mathcal{A}}\{\bx: w_j/\hpi_j\cdot \widehat{\tp}_{Y|X=\bx}(Y=j) < (w_k+\lambda_k)/\hpi_k\cdot \widehat{\tp}_{Y|X=\bx}(Y=k)\},
\end{align}
where each of $\{\bx: (w_j+\lambda_j)/\hpi_j\cdot\widehat{\tp}_{Y|X=\bx}(Y=j) < (w_k+\lambda_k)/\hpi_k\cdot \widehat{\tp}_{Y|X=\bx}(Y=k)\}$ belongs to the classification decision boundary of a linear classifier with parameter $\lambda_j$ if we see $\{\widehat{\tp}_{Y|X=\bx}(Y=j)\}_j$ as the predictors. Denote $s_{\lambda_j}(\bx)= \mathds{1}((w_j+\lambda_j)/\hpi_j\cdot\widehat{\tp}_{Y|X=\bx}(Y=j) < (w_k+\lambda_k)/\hpi_k\cdot \widehat{\tp}_{Y|X=\bx}(Y=k))$ and $\tilde{s}_{\lambda_j}(\bx)= \mathds{1}(w_j/\hpi_j\cdot\widehat{\tp}_{Y|X=\bx}(Y=j) < (w_k+\lambda_k)/\hpi_k\cdot \widehat{\tp}_{Y|X=\bx}(Y=k))$. For any $\tilde{n}$ data points $\{\bx_i\}_{i=1}^{\tilde{n}}$, denote $\{\{s_{\lambda_j}(\bx_i)\}_{i=1}^{\tilde{n}}: \lambda_j \geq 0\}$ as $\mathcal{S}_j(\{\bx_i\}_{i=1}^{\tilde{n}})$ for $j \in \mathcal{A}\backslash \{k\}$, $\{\{\tilde{s}_{\lambda_j}(\bx_i)\}_{i=1}^{\tilde{n}}: \lambda_j \geq 0\}$ as $\widetilde{\mathcal{S}}_j(\{\bx_i\}_{i=1}^{\tilde{n}})$ for $j \notin \mathcal{A}$, $\{\{\mathds{1}(\hat{g}^{(k)}_{\blambda}(\bx_i) > 0)\}_{i=1}^{\tilde{n}}: \blambda \in \mathbb{R}^{|\mathcal{A}|}_+\}$ as $\mathcal{S}^{(k)}(\{\bx_i\}_{i=1}^{\tilde{n}})$, which include all possible classification result of $\{\bx_i\}_{i=1}^{\tilde{n}}$ for all possible $\lambda_j$ values. Since linear classifiers have finite VC dimension $d_k$, by Sauer's lemma, when $\tilde{n} > d_k$, $\norm{\mathcal{S}_j(\{\bx_i\}_{i=1}^{\tilde{n}})} \leq C\tilde{n}^{d_k}$, $\norm{\widetilde{\mathcal{S}}_{j'}(\{\bx_i\}_{i=1}^{\tilde{n}})}  \leq C\tilde{n}^{d_k}$. And it's easy to see that $\norm{\mathcal{S}^{(k)}(\{\bx_i\}_{i=1}^{\tilde{n}})} \leq \prod_{j \in \mathcal{A}\backslash\{k\}} \norm{\mathcal{S}_j(\{\bx_i\}_{i=1}^{\tilde{n}})}\cdot \prod_{j \notin \mathcal{A}} \norm{\widetilde{\mathcal{S}}_j(\{\bx_i\}_{i=1}^{\tilde{n}})} \leq C'\tilde{n}^{Kd_k}$. If $\textrm{VC}(\mathcal{A}_k) > \tilde{n}$, then we must have $\norm{\mathcal{S}_k(\{\bx_i\}_{i=1}^{\tilde{n}})} = 2^{\tilde{n}} > C'\tilde{n}^{Kd_k}$ when $\tilde{n}$ is larger than some constant, which is contradicted. Therefore $\textrm{VC}(\mathcal{A}_k)$ must be finite. The same arguments hold with $k \notin \mathcal{A}$.

Then the $\epsilon$-covering number of $\mathcal{G}_k(\Lambda) = \{(\mathds{1}(\hat{g}^{(k)}_{\blambda}(\bx^{(k)}_1) > 0), \ldots, \mathds{1}(\hat{g}^{(k)}_{\blambda}(\bx^{(k)}_{n_k}) > 0)): \blambda \in \Lambda\}$ w.r.t. $\ell_{n_k}^2$-norm satisfies $\mathcal{N}(\epsilon, \mathcal{G}_k, \ell_{n_k}^2) \leq (C/\epsilon)^{V}$ with $\|\bx\|_{\ell_n^2} \coloneqq \sqrt{n^{-1}\sum_{i=1}^n x_i^2}$ for $\bx = (x_1, \ldots, x_n)^T$, where $V$ is a universal constant for any $k$. By Lemma 26.2 in \citeapp{shalev2014understanding},

\begin{equation}
	\te\left[\sup_{\blambda \in \Lambda}\norm{\widehat{R}_k(\hphi_{\blambda}) - R_k(\hphi_{\blambda})}\Big|\mathcal{D}_2, n_k\right] \leq 2\te \,\text{Rad}_n(\mathcal{G}_k(\Lambda)).
\end{equation}
where Rademacher complexity $\text{Rad}_n(A) = n^{-1}\te_{\bm{\sigma}}\sup_{\bm{a} \in A}|\bm{\sigma}^T\bm{a}|$ and $\bm{\sigma} = (\sigma_1, \ldots, \sigma_n)^T$ where each of $\sigma_i$ independently follows $\text{Unif}(\{+1, -1\})$. Denote $n_k = \#\{i = 1, \ldots, n: y_i = k\}$. Then by applying Dudley's entropy integral (e.g., Theorem 3.1 in \citealpapp{koltchinskii2011oracle}), for any $\{\bx_1^{(k)}, \ldots, \bx_{n_k}^{(k)}\}$, we get

\begin{equation}
	\text{Rad}_n(\mathcal{G}_k(\Lambda))\lesssim \int_0^{1/2}\sqrt{\frac{\log \mathcal{N}(\epsilon, \mathcal{G}_k, \ell_{n_k}^2)}{n_k}} d\epsilon \lesssim n_k^{-1/2}\int_0^{1/2} \sqrt{\log (C/\epsilon)} d\epsilon \lesssim \sqrt{\frac{1}{n_k}}, 
\end{equation}
leading to
\begin{equation}
	\te\left[\sup_{\blambda \in \Lambda}\norm{\widehat{R}_k(\hphi_{\blambda}) - R_k(\hphi_{\blambda})}\Big|\mathcal{D}_2, n_k\right] \lesssim \sqrt{\frac{1}{n_k}}.
\end{equation}
Then by the bounded difference inequality, 
\begin{equation}
	\tp\left(\sup_{\blambda \in \Lambda}\norm{\widehat{R}_k(\hphi_{\blambda}) - R_k(\hphi_{\blambda})} > \tau + Cn_k^{-1/2}\Big| \mathcal{D}_2, n_k\right) \lesssim \exp\{-Cn_k\tau^2\},
\end{equation}
Thus,
\begin{align}
	&\tp\left(\sup_{\blambda \in \Lambda}\norm{\widehat{R}_k(\hphi_{\blambda}) - R_k(\hphi_{\blambda})} > \tau + Cn^{-1/2}\Big| \mathcal{D}_2\right) \\
	&\leq \te\left[\tp\left(\sup_{\blambda \in \Lambda}\norm{\widehat{R}_k(\hphi_{\blambda}) - R_k(\hphi_{\blambda})} > \tau + Cn_k^{-1/2}\Big| \mathcal{D}_2, n_k \geq \frac{1}{2}n\pi_k^* \right)\Big| \mathcal{D}_2\right] + \tp\left(n_k < \frac{1}{2}n\pi_k^*\right) \\
	&\lesssim \exp\{-Cn\tau^2\} + \exp\{-Cn\}, 
\end{align}
where the constants are not related to $\mathcal{D}_2$. By taking the expectation w.r.t. $\mathcal{D}_2$, we get
\begin{equation}
	\tp\left(\sup_{\blambda \in \Lambda}\norm{\widehat{R}_k(\hphi_{\blambda}) - R_k(\hphi_{\blambda})} > \tau + Cn^{-1/2}\right)\lesssim \exp\{-Cn(\tau^2\wedge 1)\}
\end{equation}
Since $\hL(\blambda, \hphi_{\blambda})-L(\blambda, \hphi_{\blambda})$ is a linear combination of $\widehat{R}_k(\hphi_{\blambda}) - R_k(\hphi_{\blambda})$ with different $k$'s, by union bounds, we have
\begin{equation}
	\tp\left(\sup_{\blambda \in \Lambda}\norm{\hL(\blambda, \hphi_{\blambda})-L(\blambda, \hphi_{\blambda})} > \tau + Cn^{-1/2}\right)\lesssim \exp\{-Cn(\tau^2 \wedge 1)\}. \label{eq: finite bound alg 2 step 1}
\end{equation}
Applying similar arguments in \eqref{eq: finite bound lemma step 2}, we get
\begin{align}
	\tp\left(\sup_{\blambda \in \Lambda}\norm{L(\blambda, \hphi_{\blambda}) - L(\blambda, \phi_{\blambda}^*)} > \tau + Cn^{-1/2}\right) &\lesssim \tau^{-1}\max_k\te\norm{\widehat{\tp}_{Y|X}(Y=k) - \tp_{Y|X}(Y=k)} \\
	&\quad + \exp\{-Cn\tau^2\}. \label{eq: finite bound alg 2 step 2}
\end{align}
Combine \eqref{eq: finite bound alg 2 step 1} and \eqref{eq: finite bound alg 2 step 2}, we obtain
\begin{equation}
	\tp\left(\sup_{\blambda \in \Lambda}\norm{\hL(\blambda, \hphi_{\blambda}) - L(\blambda, \phi_{\blambda}^*)} > \tau \right) \lesssim \tau^{-1}\max_k\te\norm{\widehat{\tp}_{Y|X}(Y=k) - \tp_{Y|X}(Y=k)}  + \exp\{-Cn(\tau^2 \wedge 1)\}.
\end{equation}
when $\tau \gtrsim n^{-1/2}$.

\subsection{Proof of propositions}
\subsubsection{Proof of Proposition \ref{prop: concave G CX}}
Because $G(\blambda) = \min_{\phi}L(\blambda, \phi)$ and $L(\blambda, \phi)$ is an affine function in $\blambda$, by definition $G(\blambda)$ is concave. Similarly, by definition,  $\widehat{G}(\blambda) = \min_{\phi}\widehat{F}_{\blambda}(\phi)$, where $\widehat{F}_{\blambda}(\phi)$ is an affine function in $\blambda$. Therefore $\widehat{G}(\blambda)$ is concave as well, which completes our proof.

\subsection{Proof of theorems}

\subsubsection{Proof of Theorem \ref{thm: strong duality}}
(\rom{1}) When the strong duality holds, suppose $\phi^*$ is a solution to the NPMC problem \eqref{eq: NPMC}. We claim that 
\begin{equation}
	\phi^* \in \argmin_{\phi}L(\blambda^*, \phi).
\end{equation}
To see this, note that
\begin{align}
	J(\phi^*) &= L(\blambda^*, \phi_{\blambda^*}^*) \\
	&= \min_{\phi}L(\blambda^*, \phi) \\
	&\leq L(\blambda^*, \phi^*) \\
	&= J(\phi^*) + \sum_{k \in \mathcal{A}}\lambda_k[R_k(\phi^*) - \alpha_k] \\
	&\leq J(\phi^*),
\end{align}
which implies that $\sum_{k \in \mathcal{A}}\lambda_k[R_k(\phi^*) - \alpha_k] = 0$ and $J(\phi^*) = L(\blambda^*, \phi_{\blambda^*}^*) = L(\blambda^*, \phi^*)$. Hence, our claim is correct. In addition, by the proof of Lemma \ref{lemma: cs oracle classifier}, any classifier $\phi_{\blambda^*}$ having form $\bx \mapsto \argmax_{k}\{c_k(\blambda^*, \bpi^*)\tp_{Y|X=\bx}(Y=k)\}$ \footnote{If there is a tie, then the $\argmax_k$ can be any $k$ within that tie.} is ``pointwise" optimal in the CS problem \eqref{eq: dual of NPMC}, in the sense that
\begin{equation}
	\sum_{k=1}^K \pi_k^* c_k(\blambda, \bpi^*)\mathds{1}(\phi(\bx) \neq k) \geq \sum_{k=1}^K \pi_k^* c_k(\blambda^*, \bpi^*)\mathds{1}(\phi^*_{\blambda^*}(\bx) \neq k),
\end{equation}
for a.s. $\bx$ w.r.t. $\tp_X$ and any measurable deterministic function $\phi: \mathcal{X} \rightarrow 1:K$. Since $X|Y=k$ is absolutely continous, for any $\tilde{\phi} \in \argmin_{\phi}L(\blambda^*, \phi)$, $\tilde{\phi}$ must be of form $\bx \mapsto \argmax_{k}\{c_k(\blambda^*, \bpi^*)\tp_{Y|X=\bx}(Y=k)\}$, a.s. w.r.t. $\tp_X$. Combine this with the fact that $\phi^* \in \argmin_{\phi}L(\blambda^*, \phi)$, there must exist a classifier $\phi_{\blambda^*}$ having form $\bx \mapsto \argmax_{k}\{c_k(\blambda^*, \bpi^*)\tp_{Y|X=\bx}(Y=k)\}$ coincides with $\phi^*$ for a.s. $\bx$ w.r.t. $\tp_X$, which proves the ``only if" part.

In the following, we will prove the ``if" part by assuming such an $\blambda^{(0)}$ exists. 

WLOG, suppose $\tp_{X|Y=k}(\phi_{\blambda^{(0)}}^*(X)\neq k) < \alpha_k$ for all $k \in \mathcal{A}$. In fact, if some constraints hold with equality, then we can directly jump into Step 2. It is because, in each step, we either tune the corresponding $\lambda$ value to make the constraint hold with equality or shrink the $\lambda$ to zero. We can treat these classes in $\mathcal{A}$ for which the constraint holds with equality at the beginning in the same way as the classes for which the constraint holds with equality after tuning in Step 1.

(1) \underline{Step 1}: Let $\lambda_k = t\lambda_k^{(0)}+(t-1)w_k$ with $t \in \left[\max_{k\in \mathcal{A}}\{w_k(w_k+\lambda_k^{(0)})^{-1}\}, 1\right]$, for all $k \in \mathcal{A}$. As $t$ decreases from $1$, $\tp_{X|Y=k}(\phi^*_{\blambda}(X) \neq k)$ is non-decreasing for all $k \in \mathcal{A}$. To see this, note that $\{\bx: \phi^*_{\blambda}(\bx) = k_1, \phi^*_{\blambda^{(0)}}(\bx) = k_2\}$ does not change with $t$ for any $k_1$, $k_2 \in \mathcal{A}$, while event $\{\bx: \phi^*_{\blambda}(\bx) \in \mathcal{A}, \phi^*_{\blambda^{(0)}}(\bx) \notin \mathcal{A}\}$ is non-decreasing in $t$. Let's assume one of $\tp_{X|Y=k}(\phi^*_{\blambda}(X) \neq k) -\alpha_k$ will reach zero before $t$ hits $\max_{k}\{w_k(w_k+\lambda_k^{(0)})^{-1}\}$ for now, and we will revisit the case that $\tp_{X|Y=k}(\phi^*_{\blambda}(X) \neq k) < \alpha_k $ when $t = \max_{k}\{w_k(w_k+\lambda_k^{(0)})^{-1}\}$ by the end of the proof. Denote $t^{(0)}$ as the maximum $t$ such that at least one of equations $\tp_{X|Y=k}(\phi^*_{\blambda}(X) \neq k) -\alpha_k=0$ holds. WLOG, suppose $\tp_{X|Y=1}(\phi^*_{\blambda^{(1)}}(X) \neq 1) = \alpha_1$ and $\tp_{X|Y=k}(\phi^*_{\blambda^{(1)}}(X) \neq k) < \alpha_k$ when $k \in \mathcal{A}\backslash \{1\}$, where $\lambda^{(1)}_k = t^{(0)}\lambda^{(0)}_k + (t^{(0)}-1)w_k$.

(2) \underline{Step 2}: Let $\lambda_k = t\lambda_k^{(1)}+(t-1)w_k$ with $t \in \left[\max_{k\in \mathcal{A}\backslash \{1\}}\{w_k(w_k+\lambda_k^{(1)})^{-1}\}, 1\right]$, for all $k \in \mathcal{A}\backslash \{1\}$. We would like $\lambda_1$ to satisfy 
\begin{align}
	&\tp_{X|Y=1}(\phi_{\blambda}^*(X) \neq 1) \\
	&= \tp_{X|Y=1}\bigg(\frac{\lambda_1+w_1}{\pi_1}\tp_{Y|X}(Y=1) < \max\bigg\{\max_{k \in \mathcal{A}\backslash \{1\}}\left[\frac{\lambda_k+w_k}{\pi_k}\tp_{Y|X}(Y=k)\right], \\
	&\hspace{7.6cm} \max_{k \notin \mathcal{A}}\left[\frac{w_k}{\pi_k}\tp_{Y|X}(Y=k)\right]\bigg\}\bigg)\\
	&= \alpha_1. \label{eq: lambda_1 alpha_1}
\end{align}
Therefore we can solve $\lambda_1 = \lambda_1(t)$ from \eqref{eq: lambda_1 alpha_1} as an increasing function of $t$. Note that when $t = 1$, $\lambda_1 = \lambda_1^{(1)}$. Similar to Step 1, as $t$ decreases from $1$, it can be shown that $\tp_{X|Y=k}(\phi^*_{\blambda}(X) \neq k)$ are non-decreasing for all $k \in \mathcal{A}\backslash \{1\}$. Again, we assume one of $\tp_{X|Y=k}(\phi^*_{\blambda}(X) \neq k) -\alpha_k$ will be zero before $t$ hits $\max_{k \in \mathcal{A}\backslash \{1\}}\{w_k(w_k+\lambda_k^{(0)})^{-1}\}$, and denote $t^{(1)}$ as the maximum $t$ such that at least one of equations $\tp_{X|Y=k}(\phi^*_{\blambda}(X) \neq k) -\alpha_k=0$ holds. WLOG, suppose $\tp_{X|Y=1}(\phi^*_{\blambda^{(2)}}(X) \neq 1) = \alpha_1$, $\tp_{X|Y=2}(\phi^*_{\blambda^{(2)}}(X) \neq 2) = \alpha_2$ and $\tp_{X|Y=k}(\phi^*_{\blambda^{(2)}}(X) \neq k) < \alpha_k$ when $k \in \mathcal{A}\backslash \{1, 2\}$, where $\lambda^{(2)}_k = t^{(1)}\lambda^{(1)}_k + (t^{(1)}-1)w_k$, $\lambda_1^{(2)} = \lambda_1(t^{(1)})$.

(3) \underline{Step 3}: Let $\lambda_k = t\lambda_k^{(2)}+(t-1)w_k$ with $t \in \left[\max_{k\in \mathcal{A}\backslash \{1,2\}}\{w_k(w_k+\lambda_k^{(1)})^{-1}\}, 1\right]$, for all $k \in \mathcal{A}\backslash \{1,2\}$ \ldots

Continue this process, until the final classifier $\phi^*_{\blambda}$ corresponding to the final $\blambda$ satisfies all constraints with equality. That is, in the final step, we obtain $\tilde{\blambda}$ such that $\tp_{X|Y=k}(\phi^*_{\tilde{\blambda}}(X) \neq k) = \alpha_k$ for all $k \in \mathcal{A}$. Define the Lagrangian function $L(\blambda, \phi) = \sum_{k \in \mathcal{A}}\lambda_k[\tp_{X|Y=k}(\phi(X)\neq k) -\alpha_k] + \sum_{k=1}^K w_k\tp_{X|Y=k}(\phi(X)\neq k)$. Then since $\phi^*_{\tilde{\blambda}}$ is feasible in NP problem,
\begin{align}
	L(\tilde{\blambda}, \phi^*_{\tilde{\blambda}}) &= \inf_{\phi}L(\blambda, \phi) \\
	&= \sum_{k=1}^K w_k\tp_{X|Y=k}(\phi^*_{\tilde{\blambda}}(X)\neq k) \\
	&\geq \sum_{k=1}^K w_k\tp_{X|Y=k}(\phi^*(X)\neq k) \\
	&= \inf_{\phi}\sup_{\blambda \in \mathbb{R}^{|\mathcal{A}|}_+} L(\blambda, \phi).
\end{align}
However, by definition, 
\begin{equation}\label{eq: saddle point}
	L(\tilde{\blambda}, \phi^*_{\tilde{\blambda}}) \leq \sup_{\blambda \in \mathbb{R}^{|\mathcal{A}|}_+}\inf_{\phi} L(\blambda, \phi) \leq \inf_{\phi}\sup_{\blambda \in \mathbb{R}^{|\mathcal{A}|}_+} L(\blambda, \phi).
\end{equation}
Therefore $\sup_{\blambda \in \mathbb{R}^{|\mathcal{A}|}_+} \inf_{\phi}L(\blambda, \phi) = \inf_{\phi}\sup_{\blambda \in \mathbb{R}^{|\mathcal{A}|}_+} L(\blambda, \phi)$, which implies the strong duality.

The last thing left is to discuss the issue we mentioned in Step 1, i.e. what happens if $\tp_{X|Y=k}(\phi^*_{\blambda}(X) \neq k) < \alpha_k $ when $t = \max_{k}\{w_k(w_k+\lambda_k^{(0)})^{-1}\}$. At this time, at least one $\lambda_k$ will be zero. WLOG, suppose $\lambda_1 = 0$ while the other $\lambda_k > 0$. Let $\lambda_k'(t) = t\lambda_k+(t-1)w_k$, where $t \in \left[\max_{k\in \mathcal{A}\backslash\{1\}}\{w_k(w_k+\lambda_k)^{-1}\}, 1\right]$, for $k \in \mathcal{A}\backslash\{1\}$. Again, it can be shown that as $t$ decreases from $1$, $\tp_{X|Y=k}(\phi^*_{\blambda}(X) \neq k)$ are non-decreasing for all $k \in \mathcal{A}$. Then we will either find some $t$ such that  $\tp_{X|Y=k}(\phi^*_{\blambda'}(X) \neq k) =\alpha_k$ holds for some $k \in \mathcal{A}\backslash\{1\}$, or get $\tp_{X|Y=k}(\phi^*_{\blambda}(X) \neq k) < \alpha_k$ for all $k$ when $t = \max_{k \in \mathcal{A}\backslash\{1\}}\{w_k(w_k+\lambda_k)^{-1}\}$. Repeating the process will lead to two possible outcomes. One is that $\tp_{X|Y=k}(\phi^*_{\blambda}(X) \neq k) = \alpha_k$ holds for at least one $k$ with some $\blambda$. The other one is that $\tp_{X|Y=k}(\phi^*_{\bm{0}}(X) \neq k) < \alpha_k$ for all $k \in \mathcal{A}$. In the first case, we can continue the steps above to finally get some $\blambda''$ such that $\tp_{X|Y=k}(\phi^*_{\blambda''}(X) \neq k) < \alpha_k$ if and only if $\lambda_k'' = 0$. This implies
\begin{align}
	\sup_{\blambda \in \mathbb{R}^{|\mathcal{A}|}_+}\inf_{\phi} L(\blambda, \phi) &= L(\blambda'', \phi^*_{\blambda''}) \\
	&\geq \sum_{k=1}^K w_k\tp_{X|Y=k}(\phi^*_{\blambda''}(X)\neq k) \\
	&\geq \sum_{k=1}^K w_k\tp_{X|Y=k}(\phi^*(X)\neq k)\\
	&= \inf_{\phi}\sup_{\blambda \in \mathbb{R}^{|\mathcal{A}|}_+}L(\blambda, \phi) 
\end{align}
Combining this with \eqref{eq: saddle point}, $\sup_{\blambda \in \mathbb{R}^{|\mathcal{A}|}_+} \inf_{\phi} L(\blambda, \phi) = \inf_{\phi}\sup_{\blambda \in \mathbb{R}^{|\mathcal{A}|}_+} L(\blambda, \phi)$, i.e. the strong duality holds. The case that $\tp_{X|Y=k}(\phi^*_{\bm{0}}(X) \neq k) < \alpha_k$ for all $k \in \mathcal{A}$ can be justified to imply strong duality with the same reason. Therefore, the case we omit in Step 1 leads to strong duality as well, so do the same cases in other steps.

\vspace{0.5cm}
\noindent(\rom{2}) When the strong duality holds, if there exists some $\blambda$ such that $\phi^*_{\blambda}$ is feasible for NP problem, then by (\rom{1}) the NP problem should be feasible as well, which is a contradiction. Therefore the ``only if" part holds. Next, we will prove the ``if" part, where we assume that for any $\blambda \in \mathbb{R}^{|\mathcal{A}|}_+$, $\exists$ at least one $k \in \mathcal{A}$ such that $R_k(\phi^*_{\blambda})=\tp_{X|Y=k}(\phi_{\blambda}^*(X) \neq k) > \alpha_k$.

Define the cost
\begin{equation}
	c_k = c_k(\blambda) = \begin{cases}
		w_k/\pi_k^*, \quad &k \notin \mathcal{A};\\
		(w_k+\lambda_k)/\pi_k^*, \quad &k \in \mathcal{A}.
	\end{cases}
\end{equation}

(1) \underline{Step 1}: We arbitrarily pick one $\blambda \in \mathbb{R}^{|\mathcal{A}|}_+$. Suppose $R_{k_1}(\phi^*_{\blambda}) > \alpha_{k_1}$ for some $k_1$. Due to the assumption that $\min_{k \in \mathcal{A}}\tp_{Y|X=\bx}(Y=k)\geq a > 0$ for a.s. $\bx$ w.r.t. $\tp_X$ and the assumption that $Y|X = k$ is absolutely continuous, if we increase $\lambda_{k_1}$ and keep other $\lambda_k$'s fixed, then $R_{k_1}(\phi^*_{\blambda})$ will decrease and finally equal $\alpha_{k_1}$. Denote the current $\blambda$ as $\blambda^{(1)}$.

(2) \underline{Step 2}: By the condition, there exists $k_2 \in \mathcal{A}\backslash \{k_1\}$ such that $R_{k_2}(\phi^*_{\blambda^{(1)}}) > \alpha_{k_2}$, $R_{k_1}(\phi^*_{\blambda^{(1)}}) = \alpha_{k_1}$. Next, we increase $\lambda^{(1)}_{k_2}$ and $\lambda^{(1)}_{k_1}$ at the same time, to decrease $R_{k_2}(\phi^*_{\blambda^{(1)}})$ while keeping $R_{k_1}(\phi^*_{\blambda^{(1)}}) = \alpha_{k_1}$. Here we cannot increase $\lambda^{(1)}_{k_2}$ only and leave all the other $\lambda^{(1)}_k$'s unchanged, otherwise we must have $R_{k_1}(\phi^*_{\blambda^{(1)}}) > \alpha_{k_1}$. Consider all the possible $(\lambda_{k_2}, \lambda_{k_1})$'s which belongs to $\mathcal{Q} = \{(\lambda_{k_2}, \lambda_{k_1}) \in \mathbb{R}^2: \lambda_{k_2} \geq \lambda^{(1)}_{k_2}, \lambda_{k_1} \geq \lambda^{(1)}_{k_1}, R_{k_2}(\phi^*_{\tilde{\blambda}}) = \alpha_{k_2}, R_{k_1}(\phi^*_{\tilde{\blambda}}) = \alpha_{k_1} \textup{ with } \tilde{\blambda} \textup{ satisfying } \tilde{\lambda}_{k_1} = \lambda_{k_1}, \tilde{\lambda}_{k_2} = \lambda_{k_2}, \tilde{\lambda}_k = \lambda_{k}^{(1)} \textup{ for other } k \neq k_1, k_2\}$. If $\mathcal{Q} \neq \emptyset$, then we can arbitrarily pick a pair of $(\lambda_{k_2}, \lambda_{k_1})$ in $\mathcal{Q}$, denote the corresponding $\blambda$ as $\blambda^{(2)}$, and proceed with Step 3. If $\mathcal{Q} = \emptyset$, then for any $\blambda$ with $\lambda_{k} = \lambda^{(1)}_k$ for $k \in \mathcal{A} \backslash \{k_1, k_2\}$ and $R_{k_2}(\phi^*_{\tilde{\blambda}}) = \alpha_{k_2}$, we must have $R_{k_1}(\phi^*_{\tilde{\blambda}}) > \alpha_{k_1}$, no matter what $\lambda_{k_1}$ is. Taking one such $\blambda$ such that $\tp(\phi_{\blambda}^*(X) = k|Y=k_1) = \tp(\phi_{\blambda}^*(X) = k|Y=k_2) = 0$ for all $k \neq k_1, k_2$ \footnote{This is possible due to the assumption that $\min_{k \in \mathcal{A}}\tp_{Y|X=\bx}(Y=k) \geq a > 0$. In fact, any $\blambda$ satisfying $\min_{k=k_1,k_2}c_k(\blambda,\bpi^*) a > \max_{k\neq k_1,k_2}c_k(\blambda,\bpi^*) (1-a)$ would work.}, increasing $\lambda_{k_1}$ to $\tilde{\lambda}_{k_1}$, $\lambda_{k_2}$ to $\tilde{\lambda}_{k_2}$ while keeping other $\lambda_k$ fixed, we denote the new $\blambda$ as $\tilde{\blambda}$. We require such a $\tilde{\blambda}$ to satisfy $R_{k_2}(\phi^*_{\tilde{\blambda}}) = \alpha_{k_2}$. Then due to the absolute continuity of $X|Y=k$ for any $k$, we must have $R_{k_1}(\phi^*_{\tilde{\blambda}}) > R_{k_1}(\phi^*_{\blambda}) + \epsilon > \alpha_{k_1} + \epsilon$, for some $\epsilon > 0$. Then consider $\bar{\blambda} = \bar{\blambda}(t)$ satisfying $\bar{\lambda}_{k_1} = t\tilde{\lambda}_{k_1} + (t-1)w_{k_1}$, $\bar{\lambda}_{k_2} = t\tilde{\lambda}_{k_2} + (t-1)w_{k_2}$, $\bar{\lambda}_k = \tilde{\lambda}_k^{(1)}$ for $k \in \mathcal{A} \backslash \{k_1, k_2\}$, with $t \geq 1$. It is easy to see that when $t$ increases, we have $\tp(\phi_{\bar{\blambda}}^*(X) = k|Y=k_1) = \tp(\phi_{\bar{\blambda}}^*(X) = k|Y=k_2) = 0$ for all $k \neq k_1, k_2$, and $R_{k_1}(\phi^*_{\bar{\blambda}})$ and $R_{k_2}(\phi^*_{\bar{\blambda}})$ stay the same. Hence,
	\begin{align}
		G(\bar{\blambda}(t)) &=  F_{\bar{\blambda}}(\phi^*_{\bar{\blambda}}) \\
		&= \sum_{k \notin \mathcal{A}}w_k\tp_{X|Y=k}(\phi^*_{\bar{\blambda}}(X)\neq k) + \sum_{k \in \mathcal{A}}(w_k + \bar{\lambda}_k)\tp_{X|Y=k}(\phi^*_{\bar{\blambda}}(X)\neq k) -\sum_{k \in \mathcal{A}}\bar{\lambda}_k\alpha_k \\
		&\geq  C+(w_{k_1} + \bar{\lambda}_{k_1})\left[R_{k_1}(\phi^*_{\bar{\blambda}}) -\alpha_{k_1}\right]\\
		&= C+t(w_{k_1} + \tilde{\lambda}_{k_1})\epsilon \\
		&\rightarrow +\infty,
	\end{align}
	as $t \rightarrow +\infty$, where $C$ is a constant which does not depend on $t$.
	
(3) \underline{Step 3}: We can continue the procedure described in Step 2, and it will finally terminate with an unbounded $G(\bar{\blambda}(t))$ as $t \rightarrow +\infty$. Otherwise, we will have $R_k(\phi) \leq \alpha_k$ holds with some classifier $\phi$ for all $k \in \mathcal{A}$, which is contradicted. This completes our proof.

\subsubsection{Proof of Theorem \ref{thm: error rate bound}}
Part (\rom{1}) of the proof of Theorem \ref{thm: error rate bound} is the same as part (\rom{1}) of the proof of Theorem \ref{thm: error bound alg 2}. So we sketch the main procedure here and omit the details. 

Due to Assumption \ref{asmp: second order info}, for any sufficiently small $\tau_0 > 0$, when $\blambda \in \clmball{2\tau_0}{\blambda^*}$, $\nabla^2 G(\blambda) \preceq \frac{1}{2}\nabla^2 G(\blambda^*) \prec 0$. Then through Taylor expansion,
\begin{align}
	G(\blambda) - G(\blambda^*) &= \nabla G(\blambda^*)^T(\blambda - \blambda^*) +  \frac{1}{2}(\blambda - \blambda^*)^T\nabla^2 G(\blambda^* + t_{\blambda}(\blambda - \blambda^*)) (\blambda - \blambda^*) \\
	&\leq \frac{1}{4}(\blambda - \blambda^*)^T\lambda_{\max}(\nabla^2 G(\blambda^*)) (\blambda - \blambda^*),
\end{align}
where $t_{\blambda} \in (0, 1)$. For $\blambda \in \clmball{2\tau_0}{\blambda^*}\backslash \mball{\tau_0}{\blambda^*}$, $G(\blambda) - G(\blambda^*) \leq \frac{1}{4}\tau_0^2\lambda_{\max}(\nabla^2 G(\blambda^*))$. Therefore, for any $\blambda \notin \clmball{2\tau_0}{\blambda^*}$, $\exists t_{\blambda} \in (0, 1)$ such that $(1-t_{\blambda})\blambda^* + t_{\blambda}\blambda \in \clmball{2\tau_0}{\blambda^*}\backslash \mball{\tau_0}{\blambda^*}$, which combines with concavity leading to
\begin{equation}
	(1-t_{\blambda})G(\blambda^*) + t_{\blambda}G(\blambda) \leq G((1-t_{\blambda})\blambda^* + t_{\blambda}\blambda) \leq G(\blambda^*) + \frac{1}{4}\tau^2_0 \lambda_{\max}(\nabla^2 G(\blambda^*)).
\end{equation}
It follows that $G(\blambda) \leq G(\blambda^*) + \frac{1}{4}\tau^2_0 \lambda_{\max}(\nabla^2 G(\blambda^*))$ for any $\blambda \notin \clmball{2\tau_0}{\blambda^*}$. Besides, for $\tau \leq 2\tau_0$,
\begin{align}
	&\tp\left(\sup_{\blambda \in \clmball{2\tau_0}{\blambda^*}\backslash \mball{\tau}{\blambda^*}}\widehat{G}(\blambda) -\widehat{G}(\blambda^*) \geq 0\right) \\
	&\leq \tp\left(\sup_{\blambda \in \clmball{2\tau_0}{\blambda^*}\backslash \mball{\tau}{\blambda^*}}G(\blambda) + 2\sup_{\blambda \in \clmball{2\tau_0}{\blambda^*}}\norm{\widehat{G}(\blambda)-G(\blambda)}\geq G(\blambda^*)\right)\\
	&\leq \tp\left(\sup_{\blambda \in \clmball{2\tau_0}{\blambda^*}}\norm{\widehat{G}(\blambda)-G(\blambda)} \geq -\frac{1}{8}\tau^2 \lambda_{\max}(\nabla^2 G(\blambda^*))\right).
\end{align}
Similarly, for any $\blambda \notin \clmball{2\tau_0}{\blambda^*}$, $\exists t_{\blambda} \in (0, 1)$ such that $(1-t_{\blambda})\blambda^* + t_{\blambda}\blambda \in \clmball{2\tau_0}{\blambda^*}\backslash \mball{\tau_0}{\blambda^*}$. It implies that
\begin{align}
	(1-t_{\blambda})\widehat{G}(\blambda^*) + t_{\blambda}\widehat{G}(\blambda) &\leq  \widehat{G}((1-t_{\blambda})\blambda^* + t_{\blambda}\blambda) \\
	&\leq G((1-t_{\blambda})\blambda^* + t_{\blambda}\blambda) + \sup_{\blambda \in \clmball{2\tau_0}{\blambda^*}}\norm{\widehat{G}(\blambda)-G(\blambda)} \\
	&\leq G(\blambda^*) + \frac{1}{4}\tau^2_0 \lambda_{\max}(\nabla^2 G(\blambda^*)) + \sup_{\blambda \in \clmball{2\tau_0}{\blambda^*}}\norm{\widehat{G}(\blambda)-G(\blambda)} \\
	&\leq \widehat{G}(\blambda^*) + \frac{1}{4}\tau^2_0 \lambda_{\max}(\nabla^2 G(\blambda^*)) + 2\sup_{\blambda \in \clmball{2\tau_0}{\blambda^*}}\norm{\widehat{G}(\blambda)-G(\blambda)},
\end{align}
implying
\begin{equation}
	\widehat{G}(\blambda) \leq \widehat{G}(\blambda^*) + t_{\blambda}^{-1}\left[\frac{1}{4}\tau^2_0 \lambda_{\max}(\nabla^2 G(\blambda^*)) + 2\sup_{\blambda \in \clmball{2\tau_0}{\blambda^*}}\norm{\widehat{G}(\blambda)-G(\blambda)}\right].
\end{equation}

By the proof of Theorem \ref{thm: consistency}, for any $\tau \in [C_{\textup{Rad}}(n) + \max_k \te\norm{\widehat{\tp}_{Y|X}(Y=k)-\tp_{Y|X}(Y=k)}]^{1/2}, 2\tau_0]$, similar to \eqref{eq: limsup 1} and \eqref{eq: limsup 2}, we can obtain
	\begin{align}
		\tp(\twonorm{\hblambda - \blambda^*} > \tau) &\leq \tp\left(\sup_{\blambda \notin \clmball{2\tau_0}{\blambda^*}}\widehat{G}(\blambda) \geq \widehat{G}(\blambda^*)\right) + \tp\left(\sup_{\blambda \in \clmball{2\tau_0}{\blambda^*}\backslash \mball{\tau}{\blambda^*}}\widehat{G}(\blambda) \geq \widehat{G}(\blambda^*)\right) \\
		&\leq \tp\left(\sup_{\blambda \in \clmball{2\tau_0}{\blambda^*}}\norma{\widehat{G}(\blambda) - G(\blambda)} \geq -\frac{1}{8}\tau^2\lambda_{\max}(\nabla^2 G(\blambda^*))\right) \\
		&\quad + \tp\left(\sup_{\blambda \in \clmball{2\tau_0}{\blambda^*}}\norm{\widehat{G}(\blambda)-G(\blambda)} \geq -\frac{1}{8}\left(\inf_{\blambda \notin \clmball{2\tau_0}{\blambda^*}}t_{\blambda}^{-1}\right)\tau^2 \lambda_{\max}(\nabla^2 G(\blambda^*)) \right)\\
		&\leq 2 \tp\left(\sup_{\blambda \in \clmball{2\tau_0}{\blambda^*}}\norma{\widehat{G}(\blambda) - G(\blambda)} \geq -\frac{1}{8}\tau^2\lambda_{\max}(\nabla^2 G(\blambda^*))\right) \\
		&\lesssim \tp\left(\sup_{\blambda \in \clmball{2\tau_0}{\blambda^*}}\norm{\hL(\blambda, \hphi_{\blambda}) - L(\blambda, \phi_{\blambda}^*)} > C'\tau^2\right) \\ 
		&\lesssim \exp\{-Cn(\tau^4\wedge 1)\} + \tau^{-2}\max_k \te\norma{\widehat{\tp}_{Y|X}(Y=k)-\tp_{Y|X}(Y=k)}, \\ \label{eq: thm 2 prop 2}
	\end{align}
	where the last inequality comes from Lemma \ref{lem: finite bound} and the third last inequality comes from the fact $t_{\blambda}^{-1} > 1$ for any $\blambda \notin \clmball{2\tau}{\blambda^*}$. Note that $t_{\blambda} \in (0, 1)$ is defined as the largest $t \in (0, 1)$ value such that $(1-t)\blambda^* + t\blambda \in \bar{\mathcal{B}}_{2\tau}(\blambda^*) \backslash \mathcal{B}_{\tau}(\blambda^*)$. On the other hand, if $\tau > 2\tau_0$, by a similar analysis as in \eqref{eq: thm 2 prop 2}, we have
	\begin{align}
		\tp(\twonorm{\hblambda - \blambda^*} > \tau) &\leq \tp(\twonorm{\hblambda - \blambda^*} > \tau_0) \\
		&\lesssim \exp\{-Cn\tau_0^4\} + \tau_0^{-2}\max_k \te\norma{\widehat{\tp}_{Y|X}(Y=k)-\tp_{Y|X}(Y=k)}. \\ \label{eq: thm 2 prop 2 eq 2}
	\end{align}
	Combining \eqref{eq: thm 2 prop 2} and \eqref{eq: thm 2 prop 2 eq 2}, we have
	\begin{equation}
		\tp(\twonorm{\hblambda - \blambda^*} > \tau) \lesssim \exp\{-Cn(\tau^4\wedge 1)\} + (\tau \wedge 1)^{-2} \max_k \te\norma{\widehat{\tp}_{Y|X}(Y=k)-\tp_{Y|X}(Y=k)}.
	\end{equation}
	Then we follow \eqref{eq: proof eq R bound 1} and \eqref{eq: proof eq R bound 2} in the proof of Theorem \ref{thm: error bound alg 2} to get the desired bound.

For part (\rom{2}), by recalling part (\rom{2}) in the proof of Theorem \ref{thm: consistency} and letting $M = 2+\delta$, there exists a compact set $\Lambda \subseteq \mathbb{R}^{|\mathcal{A}|}_+$, such that $\sup_{\blambda \in \Lambda}G(\blambda) > 2+\delta$. Therefore, by Lemma \ref{lem: finite bound},
\begin{align}
	\tp\left(\sup_{\blambda \in \mathbb{R}^{|\mathcal{A}|}_+}\widehat{G}(\blambda) > 1 + \delta\right) &\geq  \tp\left(\sup_{\blambda \in \Lambda}\norm{G(\blambda)-\widehat{G}(\blambda)} \leq 1, \sup_{\blambda \in \Lambda}G(\blambda) > 2+\delta\right) \\
	&= \tp\left(\sup_{\blambda \in \Lambda}\norm{G(\blambda)-\widehat{G}(\blambda)} \leq 1\right) \\
	&\geq 1- C \left(\max_k\te\norm{\widehat{\tp}_{Y|X}(Y=k) - \tp_{Y|X}(Y=k)} + \exp\{-Cn\}\right),
\end{align}
when $\delta \gtrsim [C_{\textup{Rad}}(n) + \max_k \te\norm{\widehat{\tp}_{Y|X}(Y=k)-\tp_{Y|X}(Y=k)}]^{1/2}$, which completes the proof.

\subsubsection{Proof of Theorem \ref{thm: consistency}}

\noindent(\rom{1}) By Lemmas \ref{lem: uniform bdd 1} and \ref{lem: uniform bdd 2}, for any bounded set $\Lambda \subseteq \mathbb{R}^{|\mathcal{A}|}_+$, 
\begin{equation}\label{eq: a.s. convergence function g}
	\lim_{n \rightarrow \infty}\sup_{\blambda \in \Lambda} \norm{\widehat{G}(\blambda)-G(\blambda)} = 0, a.s..
\end{equation}
Due to Assumption \ref{asmp: second order info}, for any sufficiently small $\tau_0 > 0$, when $\blambda \in \clmball{2\tau_0}{\blambda^*}$, $\nabla^2 G(\blambda) \preceq \frac{1}{2}\nabla^2 G(\blambda^*) \prec 0$. Then by Taylor expansion,
\begin{align}
	G(\blambda) - G(\blambda^*) &= \nabla G(\blambda^*)^T(\blambda - \blambda^*) +  \frac{1}{2}(\blambda - \blambda^*)^T\nabla^2 G(\blambda^* + t_{\blambda}(\blambda - \blambda^*)) (\blambda - \blambda^*) \\
	&\leq \frac{1}{4}(\blambda - \blambda^*)^T\lambda_{\max}(\nabla^2 G(\blambda^*)) (\blambda - \blambda^*),\label{eq: proof of theorem 7 eq 1}
\end{align}
where $t_{\blambda} \in (0, 1)$. For $\blambda \in \clmball{2\tau_0}{\blambda^*}\backslash \mball{\tau_0}{\blambda^*}$, $G(\blambda) - G(\blambda^*) \leq \frac{1}{4}\tau_0^2\lambda_{\max}(\nabla^2 G(\blambda^*))$. Therefore, for any $\blambda \notin \clmball{2\tau_0}{\blambda^*}$, $\exists t_{\blambda} \in (0, 1)$ such that $(1-t_{\blambda})\blambda^* + t_{\blambda}\blambda \in \clmball{2\tau_0}{\blambda^*}\backslash \mball{\tau_0}{\blambda^*}$, which combines with concavity leading to
\begin{equation}
	(1-t_{\blambda})G(\blambda^*) + t_{\blambda}G(\blambda) \leq G((1-t_{\blambda})\blambda^* + t_{\blambda}\blambda) \leq G(\blambda^*) + \frac{1}{4}\tau^2_0 \lambda_{\max}(\nabla^2 G(\blambda^*)).
\end{equation}
It follows that $G(\blambda) \leq G(\blambda^*) + \frac{1}{4}\tau^2_0 \lambda_{\max}(\nabla^2 G(\blambda^*))$ for any $\blambda \notin \clmball{2\tau_0}{\blambda^*}$. Besides,
\begin{align}
	&\tp\left(\limsup_{n \rightarrow \infty}\left[\sup_{\blambda \in \clmball{2\tau_0}{\blambda^*}\backslash \mball{\tau_0}{\blambda^*}}\widehat{G}(\blambda) -\widehat{G}(\blambda^*)\right] \geq 0\right) \\
	&\leq \tp\left(\sup_{\blambda \in \clmball{2\tau_0}{\blambda^*}\backslash \mball{\tau_0}{\blambda^*}}G(\blambda) + 2\limsup_{n \rightarrow \infty}\sup_{\blambda \in \clmball{2\tau_0}{\blambda^*}}\norm{\widehat{G}(\blambda)-G(\blambda)}\geq G(\blambda^*)\right)\\
	&\leq \tp\left(\limsup_{n \rightarrow \infty}\sup_{\blambda \in \clmball{2\tau_0}{\blambda^*}}\norm{\widehat{G}(\blambda)-G(\blambda)} \geq -\frac{1}{8}\tau_0^2 \lambda_{\max}(\nabla^2 G(\blambda^*))\right)\\
	&= 0. \label{eq: limsup 1}
\end{align}
Similarly, for any $\blambda \notin \clmball{2\tau_0}{\blambda^*}$, $\exists t_{\blambda} \in (0, 1/2]$ such that $(1-t_{\blambda})\blambda^* + t_{\blambda}\blambda \in \clmball{2\tau_0}{\blambda^*}\backslash \mball{\tau_0}{\blambda^*}$. Combining this fact and \eqref{eq: proof of theorem 7 eq 1} with the concavity of $\widehat{G}(\blambda)$, it implies that
\begin{align}
	(1-t_{\blambda})\widehat{G}(\blambda^*) + t_{\blambda}\widehat{G}(\blambda) &\leq  \widehat{G}((1-t_{\blambda})\blambda^* + t_{\blambda}\blambda) \\
	&\leq G((1-t_{\blambda})\blambda^* + t_{\blambda}\blambda) + \sup_{\blambda \in \clmball{2\tau_0}{\blambda^*}}\norm{\widehat{G}(\blambda)-G(\blambda)} \\
	&\leq G(\blambda^*) + \frac{1}{4}\tau^2_0 \lambda_{\max}(\nabla^2 G(\blambda^*)) + \sup_{\blambda \in \clmball{2\tau_0}{\blambda^*}}\norm{\widehat{G}(\blambda)-G(\blambda)} \\
	&\leq \widehat{G}(\blambda^*) + \frac{1}{4}\tau^2_0 \lambda_{\max}(\nabla^2 G(\blambda^*)) + 2\sup_{\blambda \in \clmball{2\tau_0}{\blambda^*}}\norm{\widehat{G}(\blambda)-G(\blambda)},
\end{align}
implying
\begin{equation}
	\widehat{G}(\blambda) \leq \widehat{G}(\blambda^*) + t_{\blambda}^{-1}\left[\frac{1}{4}\tau^2_0 \lambda_{\max}(\nabla^2 G(\blambda^*)) + 2\sup_{\blambda \in \clmball{2\tau_0}{\blambda^*}}\norm{\widehat{G}(\blambda)-G(\blambda)}\right].
\end{equation}
Therefore,
\begin{align}
	&\tp\left(\limsup_{n \rightarrow \infty}\left[\sup_{\blambda \notin \clmball{2\tau_0}{\blambda^*}}\widehat{G}(\blambda) -\widehat{G}(\blambda^*)\right] \geq 0\right) \\
	&\leq \tp\left(\limsup_{n \rightarrow \infty}\sup_{\blambda \notin \clmball{2\tau_0}{\blambda^*}}\left\{t_{\blambda}^{-1}\left[\frac{1}{4}\tau^2_0 \lambda_{\max}(\nabla^2 G(\blambda^*)) + 2\sup_{\blambda \in \clmball{2\tau_0}{\blambda^*}}\norm{\widehat{G}(\blambda)-G(\blambda)}\right]\right\} \geq 0\right) \\
	&\leq \tp\left(\sup_{\blambda \notin \clmball{2\tau_0}{\blambda^*}}\left\{t_{\blambda}^{-1}\left[\frac{1}{4}\tau^2_0 \lambda_{\max}(\nabla^2 G(\blambda^*)) + 2\limsup_{n \rightarrow \infty}\sup_{\blambda \in \clmball{2\tau_0}{\blambda^*}}\norm{\widehat{G}(\blambda)-G(\blambda)}\right]\right\} \geq 0\right) \\
	&\leq \tp\left(\sup_{\blambda \notin \clmball{2\tau_0}{\blambda^*}}\left\{t_{\blambda}^{-1}\cdot\frac{1}{4}\tau^2_0 \lambda_{\max}(\nabla^2 G(\blambda^*))\right\} \geq 0\right) \\
	&= 0. \label{eq: limsup 2}
\end{align}
Note that the second inequality holds because the supremum over $\blambda \notin \clmball{2\tau_0}{\blambda^*}$ is unrelated to training data, therefore it's independent of the process $n \rightarrow \infty$ and we can switch the order of limit with supremum. Due to \eqref{eq: limsup 1} and \eqref{eq: limsup 2},
\begin{align}
	\tp\left(\limsup_{n \rightarrow \infty}\twonorm{\hblambda -\blambda^*} > \tau_0\right) &\leq \tp\left(\limsup_{n \rightarrow \infty}\left[\sup_{\blambda \in \clmball{2\tau_0}{\blambda^*}\backslash \mball{\tau_0}{\blambda^*}}\widehat{G}(\blambda) -\widehat{G}(\blambda^*)\right] \geq 0\right) \\
	&\quad + \tp\left(\limsup_{n \rightarrow \infty}\left[\sup_{\blambda \notin \clmball{2\tau_0}{\blambda^*}}\widehat{G}(\blambda) -\widehat{G}(\blambda^*)\right] \geq 0\right) \\
	& = 0.
\end{align}
Because the conclusion holds for arbitrarily small $\tau_0$, by letting $\tau_0 \rightarrow 0$, we have $\lim_{n\rightarrow \infty}\hblambda =\blambda^*$ a.s.. Recall that by strong law of large numbers, $\lim_{n\rightarrow \infty}\hbpi = \bpi^*$ a.s.. And by Assumption \ref{asmp: posterior distance as}, $\lim_{n\rightarrow \infty}\widehat{\tp}_{Y|X=\bx}(Y=k) = \tp_{Y|X=\bx}(Y=k)$ and $\bx$ a.s., w.r.t. the distribution of $X$ (as well as the distribution of $X|Y=k$ for any $k$, since $\pi_k^* > 0$ which implies $\tp_{X|Y=k} \ll \tp_{X}$), for all $k$'s. 

Denote $\varphi_k(\bx;\blambda, \bpi, \tilde{\tp}_{Y|X=\bx}) = c_k(\blambda, \bpi)\tilde{\tp}_{Y|X=\bx}(Y=k) -\max_{j \neq k}c_j(\blambda, \bpi)\tilde{\tp}_{Y|X=\bx}(Y=j)$, where $\tilde{\tp}_{Y|X}$ can be any posterior distribution of $Y|X$. Then by dominated convergence theorem and the continuity of $\varphi_k(\bx;\blambda, \bpi, \tilde{\tp}_{Y|X=\bx})$ w.r.t. $(\blambda, \bpi, \tilde{\tp}_{Y|X=\bx}(Y=k))$,
\begin{align}
	\lim_{n \rightarrow \infty}R_k(\hphi) &= \lim_{n \rightarrow \infty}\tp_{X|Y=k}(\varphi_k(X;\hblambda, \hbpi, \widehat{\tp}_{Y|X}) < 0) \\
	&= \tp_{X|Y=k}\left(\lim_{n \rightarrow \infty}\varphi_k(X;\hblambda, \hbpi, \widehat{\tp}_{Y|X}) < 0\right) \\
	&=  \tp_{X|Y=k}\left(\lim_{n \rightarrow \infty}\varphi_k(X;\blambda^*, \bpi^*, \tp_{Y|X}) < 0\right) \\
	&= R_k(\phi^*), \quad a.s.,
\end{align}
for any $k$. Followed by basic calculations, part (\rom{1}) of Theorem \ref{thm: consistency} is proved.

Furthermore, if $\tp(\hlambda_k > \tau_n ) \rightarrow 1$ for any vanishing sequence $\{\tau_n\}_{n=1}^{\infty} \rightarrow 0$, then by the consistency $\lambda^*_k > 0$, which implies $R_k(\phi^*) = \alpha_k$ by complementary slackness \citepapp{luenberger1997optimization, boyd2004convex}.

\noindent(\rom{2}) By strong duality, the infeasibility of the NP problem leads to $\sup_{\blambda \succeq \bm{0}}G(\blambda) = +\infty$. There exists a sequence of compact sets $\{\Lambda_j\}_{j=1}^{\infty}$ satisfying $\sup_{\blambda \in \Lambda_j}G(\blambda) \rightarrow +\infty$ as $j \rightarrow \infty$. Then for any $M > 0$, $\exists$ a positive integer $J=J(M)$, such that when $j \geq J$, $\sup_{\blambda \in \Lambda_j}G(\blambda) > 2M$. It follows that
\begin{equation}
	\tp\left(\liminf_{n \rightarrow \infty}\sup_{\blambda \succeq \bm{0}}\widehat{G}(\blambda) \geq M\right) \geq  \tp\left(\limsup_{n \rightarrow \infty}\sup_{\blambda \in \Lambda_J}\norm{G(\blambda)-\widehat{G}(\blambda)} \leq M, \sup_{\blambda \in \Lambda_J}G(\blambda) > 2M\right) =1,
\end{equation}
due to \eqref{eq: a.s. convergence function g}. Specially, by letting $M = 1+\delta$, we have proved part (\rom{2}).

\subsubsection{Proof of Theorem \ref{thm: error bound alg 2}}
Since all norms are equivalent in a finite-dimensional space, there exists $R_1$, $R_2 > 0$ such that 
\begin{equation}
	\{\blambda: \twonorm{\blambda} \leq R_1\} \subseteq \{\blambda: \infnorm{\blambda} \leq R\} \leq \{\blambda: \twonorm{\blambda} \leq R_2\}.
\end{equation}
(\rom{1}) Due to Assumption \ref{asmp: second order info}, for any sufficiently small $\tau > 0$, when $\blambda \in \clmball{2\tau}{\blambda^*}$, $\nabla^2 G(\blambda) \preceq \frac{1}{2}\nabla^2 G(\blambda^*) \prec 0$. Then by Taylor expansion,
\begin{align}
	G(\blambda) - G(\blambda^*) &= \nabla G(\blambda^*)^T(\blambda - \blambda^*) +  \frac{1}{2}(\blambda - \blambda^*)^T\nabla^2 G(\blambda^* + t_{\blambda}(\blambda - \blambda^*)) (\blambda - \blambda^*) \\
	&\leq \frac{1}{4}(\blambda - \blambda^*)^T\lambda_{\max}(\nabla^2 G(\blambda^*)) (\blambda - \blambda^*).
\end{align}
Denote the $\ell_{\infty}$-ball centered at $\blambda^*$ with radius $R$ as $\bar{\mathcal{B}}_{R}^{\infty}(\blambda^*)$. For $\blambda \in \bar{\mathcal{B}}_{R}^{\infty}(\blambda^*)\backslash \mball{\tau}{\blambda^*} \subseteq \clmball{R_2}{\blambda^*}\backslash \mball{\tau}{\blambda^*}$, $G(\blambda) - G(\blambda^*) \leq \frac{1}{4}\tau^2\lambda_{\max}(\nabla^2 G(\blambda^*))$. Therefore, for any $\blambda \notin \clmball{2\tau}{\blambda^*}$, $\exists t_{\blambda} \in (0, 1)$ such that $(1-t_{\blambda})\blambda^* + t_{\blambda}\blambda \in \clmball{2\tau}{\blambda^*}\backslash \mball{\tau}{\blambda^*}$, which combines with concavity leading to
\begin{equation}
	(1-t_{\blambda})G(\blambda^*) + t_{\blambda}G(\blambda) \leq G((1-t_{\blambda})\blambda^* + t_{\blambda}\blambda) \leq G(\blambda^*) + \frac{1}{4}\tau^2 \lambda_{\max}(\nabla^2 G(\blambda^*)).
\end{equation}
It follows that $G(\blambda) \leq G(\blambda^*) + \frac{1}{4}\tau^2 \lambda_{\max}(\nabla^2 G(\blambda^*))$ since $t_{\blambda}^{-1} > 1$. Therefore, for $\tau \in (0, R_2]$,
\begin{align}
\tp(\twonorm{\hblambda - \blambda^*} > \tau) &=\tp\left(\sup_{\blambda \in \clmball{R_2}{\blambda^*}\backslash \mball{\tau}{\blambda^*}}\widehat{G}(\blambda) \geq \widehat{G}(\blambda^*)\right) \\
	&\leq \tp\left(\sup_{\blambda \in \clmball{R_2}{\blambda^*}\backslash \mball{\tau}{\blambda^*}}G(\blambda) + 2\sup_{\blambda \in \mball{R_2}{\blambda^*}}\norm{\widehat{G}(\blambda)-G(\blambda)}\geq G(\blambda^*)\right)\\
	&\leq \tp\left(\sup_{\blambda \notin \clmball{\tau}{\blambda^*}}G(\blambda) - \frac{1}{4}\tau^2 \lambda_{\max}(\nabla^2 G(\blambda^*)) \geq G(\blambda^*)\right) \\
	&\quad + \tp\left(\sup_{\blambda \in \clmball{R_2}{\blambda^*}}\norm{\widehat{G}(\blambda)-G(\blambda)} \geq -\frac{1}{8}\tau^2 \lambda_{\max}(\nabla^2 G(\blambda^*))\right)\\
	&\lesssim  \exp\{-Cn(\tau^4 \wedge 1)\} + \tau^{-2}\max_k\te\norm{\widehat{\tp}_{Y|X}(Y=k) - \tp_{Y|X}(Y=k)}, \\ \label{eq: proof eq lambda concentration}
\end{align}
where the last inequality comes from Lemma \ref{lem: uniform conv algorithm 2}. Denote $\hat{g}^{(k)}_{\blambda}(\bx) = c_k(\blambda, \hbpi)\widehat{\tp}_{Y|X=\bx}(Y=k) - \max_{j \neq k}[c_j(\blambda, \hbpi)\widehat{\tp}_{Y|X=\bx}(Y=j)]$ and $g^{(k)}_{\blambda}(\bx) = c_k(\blambda, \bpi^*)\tp_{Y|X=\bx}(Y=k) - \max_{j \neq k}\allowbreak [c_j(\blambda, \bpi^*)\tp_{Y|X=\bx}(Y=j)]$. Note that $g^{(k)}_{\blambda^*}(\bx) = \varphi_k(\bx)$ for all $\bx$. We have
\begin{align}
	\norm{R_k(\hphi) - R_k(\phi^*)} &= \tp_{X|Y = k}(\hphi(X) \neq \phi^*(X)) \\
	&= \tp_{X|Y = k}\Big(\{\bx:\hat{g}^{(k)}_{\hblambda}(\bx)<0\} \triangle \{\bx:g^{(k)}_{\blambda^*}(\bx)<0\}\Big) \\
	&\leq \tp_{X|Y= k}\Big(\norm{g^{(k)}_{\blambda^*}(X)} \leq \norm{\hat{g}^{(k)}_{\hblambda}(X)-g^{(k)}_{\blambda^*}(X)}\Big).
\end{align}
By the triangle inequality,
\begin{align}
	\norm{\hat{g}^{(k)}_{\hblambda}(X)-g^{(k)}_{\blambda^*}(X)} &\leq \sum_{k=1}^K\norm{c_k(\hblambda, \hbpi)\widehat{\tp}_{Y|X=\bx}(Y=k) - c_k(\blambda^*, \bpi^*)\tp_{Y|X=\bx}(Y=k)} \\
	&\lesssim \onenorm{\hbpi - \bpi^*} + \onenorm{\hblambda - \blambda^*} + \sum_{k}|\widehat{\tp}_{Y|X=\bx}(Y=k)-\tp_{Y|X=\bx}(Y=k)|,
\end{align}
when $\infnorm{\hblambda} \leq C''$ with a constant $C'' \geq (2\infnorm{\blambda^*}) \vee (\infnorm{\blambda^*}+1)$. Therefore, by Assumption \ref{asmp: marginal finite sample},
\begin{align}
	\norm{R_k(\hphi) - R_k(\phi^*)} &\leq \tp_{X|Y=k}(\norm{g^{(k)}_{\blambda^*}(X)} \leq C\onenorm{\hbpi - \bpi^*}) + \tp_{X|Y=k}(\norm{g^{(k)}_{\blambda^*}(X)} \leq C\onenorm{\hblambda - \blambda^*})\\
	&\quad + \tp_{X|Y= k}\bigg(\norm{g^{(k)}_{\blambda^*}(X)} \leq C\sum_{k}|\widehat{\tp}_{Y|X=\bx}(Y=k)-\tp_{Y|X=\bx}(Y=k)|\bigg) \\
	&\lesssim \onenorm{\hbpi - \bpi^*}^{\bar{\gamma}} + \onenorm{\hblambda - \blambda^*}^{\bar{\gamma}}\\
	&\quad + \tp_{X|Y= k}\bigg(\norm{g^{(k)}_{\blambda^*}(X)} \leq C\sum_{k}|\widehat{\tp}_{Y|X=\bx}(Y=k)-\tp_{Y|X=\bx}(Y=k)|\bigg),
\end{align}
when $\infnorm{\hblambda} \leq C''$ with a constant $C'' \geq (2\infnorm{\blambda^*}) \vee (\infnorm{\blambda^*}+1)$. Hence,
\begin{align}
	&\tp(\norm{R_k(\hphi) - R_k(\phi^*)} > \tau)\\
	&\leq \tp(\onenorm{\hbpi - \bpi^*}^{\bar{\gamma}} > \tau/3) + \tp(\onenorm{\hblambda - \blambda^*}^{\bar{\gamma}} > \tau/3) + \tp(\infnorm{\hblambda} > C'')\\
	&\quad + \tp\bigg(\tp_{X|Y= k}\bigg(\norm{g^{(k)}_{\blambda^*}(X)} \leq C\sum_{k}|\widehat{\tp}_{Y|X=\bx}(Y=k)-\tp_{Y|X=\bx}(Y=k)|\bigg) > \tau/3\bigg) \\
	&\lesssim \exp\{-n\tau^{4/\bar{\gamma}}\} + \tau^{-2}\max_k\te\norm{\widehat{\tp}_{Y|X}(Y=k) - \tp_{Y|X}(Y=k)} \\
	&\quad + \tp\bigg(\tp_{X|Y= k}\bigg(\norm{g^{(k)}_{\blambda^*}(X)} \leq C\sum_{k}|\widehat{\tp}_{Y|X=\bx}(Y=k)-\tp_{Y|X=\bx}(Y=k)|\bigg) > \tau/3\bigg), \\ \label{eq: proof eq R bound 1}
\end{align}
when $1 \geq \tau \gtrsim n^{-\bar{\gamma}/4}$, where the last inequality comes from the standard concentration results of $\onenorm{\hbpi - \bpi^*}$ and the concentration of $\twonorm{\hblambda - \blambda^*}$ in \eqref{eq: proof eq lambda concentration}. Note that we assume fixed $K$ in this paper, therefore the concentration rates of $\twonorm{\hblambda - \blambda^*}$ and $\onenorm{\hblambda - \blambda^*}$ only differ by constants. Finally, it suffices to bound the last term above. By Markov inequality, with $t = (\tau/12)^{1/\bar{\gamma}}$, we have
\begin{align}
	&\tp\bigg(\tp_{X|Y= k}\bigg(\norm{g^{(k)}_{\blambda^*}(X)} \leq C\sum_{k}|\widehat{\tp}_{Y|X=\bx}(Y=k)-\tp_{Y|X=\bx}(Y=k)|\bigg) > \tau/3\bigg) \\
	&\leq \tp(\tp_{X|Y= k}(\norm{g^{(k)}_{\blambda^*}(X)} \leq t) > \tau/6) \\
	&\quad + \tp\bigg(\tp_{X|Y=k}\bigg(C\sum_{k}|\widehat{\tp}_{Y|X=\bx}(Y=k)-\tp_{Y|X=\bx}(Y=k)| \geq t\bigg) > \tau/6\bigg) \\
	&\leq \tp(t^{\bar{\gamma}} > \tau/6) + \tp\Big(CKt^{-1}\max_{k}\te_{X|Y=k}|\widehat{\tp}_{Y|X=\bx}(Y=k)-\tp_{Y|X=\bx}(Y=k)| > \tau/6\Big) \\
	&\lesssim t^{-1}\tau^{-1} \max_{k}\te|\widehat{\tp}_{Y|X=\bx}(Y=k)-\tp_{Y|X=\bx}(Y=k)| \\
	&\lesssim \tau^{-\frac{1+\bar{\gamma}}{\bar{\gamma}}}\max_{k}\te|\widehat{\tp}_{Y|X=\bx}(Y=k)-\tp_{Y|X=\bx}(Y=k)|. \label{eq: proof eq R bound 2}
\end{align}
Combining \eqref{eq: proof eq R bound 1} and \eqref{eq: proof eq R bound 2}, we get
\begin{equation}\label{eq: error rare bound last proof}
	\tp(\norm{R_k(\hphi) - R_k(\phi^*)} > \tau) \leq \exp\{-n\tau^{4/\bar{\gamma}}\} + \tau^{-\frac{2\vee  (1+\bar{\gamma})}{\bar{\gamma}}}\max_k \te \norma{\widehat{\tp}_{Y|X}(Y=k)-\tp_{Y|X}(Y=k)},
\end{equation}
when $1 \geq \tau \gtrsim n^{-\bar{\gamma}/4}$, which completes our proof.

\noindent(\rom{2}) Recall part (\rom{2}) in the proof of Theorem \ref{thm: consistency}. When $R$ is sufficiently large such that $\exists R_1 > 0$ satisfying $\{\blambda: \twonorm{\blambda} \leq R_1\} \subseteq \{\blambda: \infnorm{\blambda} \leq R\}$ and $\sup_{\blambda \in \mathbb{R}^{|\mathcal{A}|}_+, \twonorm{\blambda} \leq R_1}G(\blambda) > 2+\delta$. Therefore,
\begin{align}
	\tp\left(\sup_{\blambda \in \mathbb{R}^{|\mathcal{A}|}_+, \infnorm{\blambda} \leq R}\widehat{G}(\blambda) > 1+\delta \right) &\geq  \tp\left(\sup_{\blambda \in \mathbb{R}^{|\mathcal{A}|}_+, \twonorm{\blambda} \leq R_1}\norm{G(\blambda)-\widehat{G}(\blambda)} \leq 1, \sup_{\blambda \in \mathbb{R}^{|\mathcal{A}|}_+, \twonorm{\blambda} \leq R_1}G(\blambda) > 2+\delta\right) \\
	&= \tp\left(\sup_{\blambda \in \mathbb{R}^{|\mathcal{A}|}_+, \twonorm{\blambda} \leq R_1}\norm{G(\blambda)-\widehat{G}(\blambda)} \leq 1\right) \\
	&\geq 1- C \left(\max_k\te\norm{\widehat{\tp}_{Y|X}(Y=k) - \tp_{Y|X}(Y=k)} + \exp\{-Cn\}\right),
\end{align}
which completes the proof.

\subsubsection{Proof of Theorem \ref{thm: strong duality gnpmc}}
The proof is almost the same as the proof of Theorem \ref{thm: strong duality}, which is done by starting from an NP feasible classifier obtained by the cost-sensitive learning problem, manipulating the $\blambda$ step by step,  and reaching the desired conclusion. We omit the details here.

\subsubsection{Proof of Theorem \ref{thm: error rate bound gnpmc}}
The proof is almost identical to the proof of Theorem \ref{thm: error rate bound}. We can first prove a similar uniform concentration result as in Lemma \ref{lem: finite bound}, then apply a similar analysis used in the proof of Theorem \ref{thm: error rate bound}.

\subsubsection{Proof of Theorem \ref{thm: error bound alg 2 gnpmc}}
The proof is almost identical to the proof of Theorem \ref{thm: error bound alg 2}. We can first prove a similar uniform concentration result as in Lemma \ref{lem: uniform conv algorithm 2}, then apply a similar analysis used in the proof of Theorem \ref{thm: error rate bound}. The only difference is in the proof of Lemma \ref{lem: uniform conv algorithm 2}, we shall replace the inequality \eqref{eq: an interesting argument} with the following one: For any $X = \bx$ and $\blambda$, suppose $\hat{k} = \hat{\phi}_{\blambda}(\bx)$ and $k^* = \phi^*_{\blambda}(\bx)$,
\begin{align}
	0 &\leq \sum_{k \neq \hat{k}}c_{k\hat{k}}(\blambda, \hbpi)\tp_{Y|X=\bx}(Y=k) - \sum_{k \neq k^*}c_{kk^*}(\blambda, \bpi^*)\tp_{Y|X=\bx}(Y=k) \\
	&= \sum_{k \neq \hat{k}}c_{k\hat{k}}(\blambda, \hbpi)\tp_{Y|X=\bx}(Y=k) - \sum_{k \neq \hat{k}}c_{k\hat{k}}(\blambda, \hbpi)\widehat{\tp}_{Y|X=\bx}(Y=k) \\
	&\quad + \sum_{k \neq \hat{k}}c_{k\hat{k}}(\blambda, \hbpi)\widehat{\tp}_{Y|X=\bx}(Y=k) - \sum_{k \neq k^*}c_{kk^*}(\blambda, \hbpi)\widehat{\tp}_{Y|X=\bx}(Y=k) \\
	&\quad + \sum_{k \neq k^*}c_{kk^*}(\blambda, \hbpi)\widehat{\tp}_{Y|X=\bx}(Y=k) - \sum_{k \neq k^*}c_{kk^*}(\blambda, \bpi^*)\tp_{Y|X=\bx}(Y=k) \\
	&\leq \max_{r=1:K}\bigg|\sum_{k \neq r}[c_{kr}(\blambda, \hbpi)\widehat{\tp}_{Y|X=\bx}(Y=k)-c_{kr}(\blambda, \hbpi)\tp_{Y|X=\bx}(Y=k)]\bigg|\\
	&\quad + \max_{r=1:K}\bigg|\sum_{k \neq r}[c_{kr}(\blambda, \hbpi)\widehat{\tp}_{Y|X=\bx}(Y=k)-c_{kr}(\blambda, \bpi^*)\tp_{Y|X=\bx}(Y=k)]\bigg| \\
	&\leq 2\max_{r=1:K}\bigg|\sum_{k \neq r}[c_{kr}(\blambda, \hbpi)-c_{kr}(\blambda, \bpi^*)]\bigg| \\
	&\quad + 2\max_{r=1:K}\bigg|\sum_{k \neq r}[\widehat{\tp}_{Y|X=\bx}(Y=k)-\tp_{Y|X=\bx}(Y=k)]\bigg| \cdot \max_{k\neq r}c_{kr}(\blambda, \bpi^*) \\
	&\leq 2\max_{r=1:K}\bigg|\sum_{k \neq r}[c_{kr}(\blambda, \hbpi)-c_{kr}(\blambda, \bpi^*)]\bigg| \\
	&\quad + 2\max_{r=1:K}\bigg|\widehat{\tp}_{Y|X=\bx}(Y=r)-\tp_{Y|X=\bx}(Y=r)\bigg| \cdot \max_{k\neq r}c_{kr}(\blambda, \bpi^*) \\
	&\lesssim \max_{k=1:K}\norm{\hpi_k-\pi_k} + \max_{k=1:K}\bigg|\widehat{\tp}_{Y|X=\bx}(Y=k)-\tp_{Y|X=\bx}(Y=k)\bigg|,
\end{align}
where we used the fact that $\sum_{k \neq \hat{k}}c_{k\hat{k}}(\blambda, \hbpi)\widehat{\tp}_{Y|X=\bx}(Y=k) - \sum_{k \neq k^*}c_{kk^*}(\blambda, \hbpi)\widehat{\tp}_{Y|X=\bx}(Y=k) \leq 0$.

\subsubsection{Proof of Corollary \ref{cor: df check}}
\noindent (\rom{1}) ``$\Rightarrow$": The proof is quite similar to the proof of the ``only if" part in the proof of Theorem \ref{thm: strong duality}.(\rom{1}). Suppose $\phi^*$ is a solution of the NPMC problem. By the definition of strong duality, we have
\begin{align}
	L(\blambda^*, \phi^*_{\blambda^*}) &= \min_{\phi} F_{\blambda^*}(\phi^*_{\blambda}) \\
	&= \max_{\blambda \in \mathbb{R}_+^{|\mathcal{A}|}}\min_{\phi}L(\blambda, \phi) \\
	&= \min_{\phi \in \mathfrak{C}}J(\phi) \\
	&= J(\phi^*) \\
	&\geq  J(\phi^*) + \sum_{k \in \mathcal{A}}\lambda_k^*[\tp_{X|Y = k}(\phi^*(X) \neq k) - \alpha_k] \\
	&= L(\blambda^*, \phi^*) \\
	&\geq \min_{\phi} L(\blambda^*, \phi^*_{\blambda^*}).
\end{align}
Therefore, we must have $\phi^* \in \argmin_{\phi}F_{\blambda^*}(\phi)$. In the proof of the ``only if" part in the proof of Theorem \ref{thm: strong duality}.(\rom{1}), we have argued that for any $\tilde{\phi} \in \argmin_{\phi}L(\blambda^*, \phi)$, $\tilde{\phi}$ must be of form $\bx \mapsto \argmax_{k}\{c_k(\blambda^*, \bpi^*)\tp_{Y|X=\bx}(Y=k)\}$, a.s. w.r.t. $\tp_X$. Here we can further claim that $\argmin_{\phi}L(\blambda^*, \phi) = \phi^*_{\blambda^*}$ a.s. w.r.t. $\tp_X$. In fact, by Assumption \ref{asmp: marginal finite sample}, there is no mass on the decition boundary w.r.t. $\tp_X$, which implies that there is no tie in $\argmax_{k}\{c_k(\blambda^*, \bpi^*)\tp_{Y|X=\bx}(Y=k)\}$ a.s. w.r.t. $\bx \sim \tp_X$. Then the pointwise optimality of $\phi^*_{\blambda^*}$ guarantees that $\argmin_{\phi}L(\blambda^*, \phi) = \phi^*_{\blambda^*}$ a.s. w.r.t. $\tp_X$. Hence, we must have $\phi^*(\bx) = \phi^*_{\blambda^*}(\bx)$ a.s. w.r.t. $\bx \sim \tp_X$, which entails the feasibility of $\phi^*_{\blambda^*}$

``$\Leftarrow$": This is due to Theorem \ref{thm: strong duality}.(\rom{1}). 

\noindent (\rom{3}) ``$\Rightarrow$": Because of the definition of strong duality, $\sup_{\blambda \in \mathbb{R}^{|\mathcal{A}|}_+}G(\blambda) = +\infty$. And note that for any finite $\blambda$, $G(\blambda)$ must be finite. Hence $\blambda^*$ must be infinite.

``$\Leftarrow$": This is due to the weak duality.

\noindent (\rom{2}) ``$\Rightarrow$": By the equivalence in (\rom{1}), we must have $\blambda^*$ infinite or $\blambda^*$ finite but $\phi^*_{\blambda^*}$ infeasible. Since the NPMC problem is feasible, by weak duality, we must have $G(\blambda^*) \leq \min_{\phi \in \mathfrak{C}}J(\phi) \leq 1$.

``$\Leftarrow$": This is due to the equivalence in (\rom{1}).

\noindent (\rom{4}) ``$\Rightarrow$": By the equivalence in (\rom{2}) and the fact that NPMC is infeasible, we must have $\blambda^*$ infinite but $G(\blambda^*) < +\infty$ or $\blambda^*$ finite but $\phi^*_{\blambda^*}$ infeasible.

``$\Leftarrow$": The strong duality must fail due to (\rom{1}) and (\rom{2}). The NPMC infeasibility comes from the weak duality.

\subsubsection{Proof of Theorem \ref{thm: dk check}}
(\rom{1}) The bound comes from a similar argument as in \eqref{eq: error rare bound last proof}. Note that here \eqref{eq: error rare bound last proof} holds when $\delta \gtrsim \big[C_{\textup{Rad}}(n) + \max_k \te\norm{\widehat{\tp}_{Y|X}(Y=k)-\tp_{Y|X}(Y=k)}\big]^{\bar{\gamma}/2}$ which is true for sufficiently large $n$ and constant $\delta$. 

\noindent (\rom{2}) Note that the convergence of $\twonorm{\hblambda-\blambda^*}$ in the proof of Theorem \ref{thm: error rate bound} does not depend on the strong duality. Even if strong duality fails, we can still obtain the same high-probability upper bound for $\twonorm{\hblambda-\blambda^*}$, which implies the high-probability upper bound of $|R_k(\hphi) - R_k(\phi^*_{\blambda^*})|$ by a similar argument in \eqref{eq: error rare bound last proof}. Therefore for Algorithm \ref{algo: df check CX}, we must have
\begin{align}
	\tp\bigg(R_k(\hphi) > \alpha_k\Big(1+\frac{3}{2}\delta\Big)\bigg) &\leq  
	\tp\bigg(R_k(\hphi) - R_k(\phi^*_{\blambda^*}) > \frac{1}{2}\delta\bigg) \\
	&\lesssim \exp\{-C'n\delta^{4/\bar{\gamma}}\} + \delta^{-\frac{2\vee (1+\bar{\gamma})}{\bar{\gamma}}}\max_{k}\te |\widehat{\tp}_{Y|X}(Y = k) - \tp_{Y|X}(Y = k)|,
\end{align}
which proves part (a). The same results hold for Algorithm \ref{algo: df check ER} which proves part (b). Part (c) comes from a similar argument by replacing the inequality above with
\begin{align}
	\tp\bigg(\widehat{R}_k(\hphi) \leq \alpha_k(1+\delta)\bigg) &\leq  
	\tp\bigg(\widehat{R}_k(\hphi) - R_k(\phi^*_{\blambda^*}) < -\frac{1}{2}\delta\bigg) \\
	&\lesssim \exp\{-C'n\delta^{4/\bar{\gamma}}\} + \delta^{-\frac{2\vee (1+\bar{\gamma})}{\bar{\gamma}}}\max_{k}\te |\widehat{\tp}_{Y|X}(Y = k) - \tp_{Y|X}(Y = k)|, \label{eq: unknown number proof thm 2}
\end{align} 
and
\begin{align}
	\tp(\widehat{G}^{\textup{CX}}(\hlambda) > 1+\delta ) &\leq \tp(\widehat{G}^{\textup{CX}}(\hlambda) - G(\blambda^*) > \delta ) \\
	&\lesssim \exp\{-C'n\delta^{2}\} + \delta^{-1}\max_{k}\te |\widehat{\tp}_{Y|X}(Y = k) - \tp_{Y|X}(Y = k)| \\
\end{align}	
or
\begin{align}
	\tp(\widehat{G}^{\textup{ER}}(\hlambda) > 1+\delta ) &\leq \tp(\widehat{G}^{\textup{ER}}(\hlambda) - G(\blambda^*) > \delta ) \\
	&\lesssim \exp\{-C'n\delta^{2}\} + \delta^{-1}\max_{k}\te |\widehat{\tp}_{Y|X}(Y = k) - \tp_{Y|X}(Y = k)|.
\end{align}
by similar arguments in the proofs of Lemma \ref{lem: finite bound} and Lemma \ref{lem: uniform conv algorithm 2}. Inequality \eqref{eq: unknown number proof thm 2} is obtained by combining the high-probability upper bounds of $|\widehat{R}_k(\hphi) - R_k(\hphi)|$ and $|R_k(\hphi) - R_k(\phi^*_{\blambda^*})|$, where the first one is due to a similar argument in the proof of Lemma \ref{lem: finite bound} and the second one is due to a similar argument in \eqref{eq: error rare bound last proof}.

\noindent (\rom{3}) The bound is an immediate result by part (\rom{2}) of Theorem \ref{thm: error rate bound}.

\noindent (\rom{4}) Note that the convergence of $|\widehat{G}^{\textup{CX}}(\hblambda) - G(\blambda^*)|$ and $|\widehat{G}^{\textup{ER}}(\hblambda) - G(\blambda^*)|$ in the proofs of Theorems \ref{thm: error rate bound} and \ref{thm: error bound alg 2} does not depend on the strong duality. Therefore we must have
\begin{align}
	\tp(\widehat{G}^{\textup{CX}}(\hblambda) \leq 1+\delta) &\leq \tp(\widehat{G}^{\textup{CX}}(\hblambda) - G(\blambda^*) \leq -\delta/2) \\
	&\lesssim \exp\{-C'n\delta^{2}\} + \delta^{-1}\max_{k}\te |\widehat{\tp}_{Y|X}(Y = k) - \tp_{Y|X}(Y = k)|,
\end{align}
and 
\begin{align}
	\tp(\widehat{G}^{\textup{ER}}(\hblambda) \leq 1+\delta) &\leq \tp(\widehat{G}^{\textup{ER}}(\hblambda) - G(\blambda^*) \leq -\delta/2) \\
	&\lesssim \exp\{-C'n\delta^{2}\} + \delta^{-1}\max_{k}\te |\widehat{\tp}_{Y|X}(Y = k) - \tp_{Y|X}(Y = k)|.
\end{align}

\subsubsection{Proof of Theorem \ref{thm: obj prediction dk check}}
Suppose $\blambda^* = \argmax_{\blambda \in \mathbb{R}_+^{|\mathcal{A}|}}G(\blambda)$. Then by Assumption \ref{asmp: second order info},
\begin{align}
	\tp(|\widehat{G}^{\textup{CX}}(\hblambda) - G(\blambda^*)| > \tau_1) &\leq \tp(|\widehat{G}^{\textup{CX}}(\hblambda) - G(\hblambda)| > \tau_1/2) + \tp(|G(\hblambda) - G(\blambda^*)| > \tau_1/2) \\
	&\leq \tp\bigg(\sup_{\blambda \in \mathcal{B}_1(\blambda^*)}|\widehat{G}^{\textup{CX}}(\blambda) - G(\blambda)| > \tau_1/2\bigg) \\
	&\quad + \tp(\twonorm{\hblambda - \blambda^*} >1) + \tp(C\twonorm{\hblambda - \blambda^*} > \tau_1) \\
	&\leq \exp\{-Cn\tau_1^4\} + \tau_1^{-2}\max_k \te\norma{\widehat{\tp}_{Y|X}(Y=k)-\tp_{Y|X}(Y=k)},
\end{align}
where $1 \geq \tau_1 \gtrsim \sqrt{C_{\textup{Rad}}} + \big[\max_k \te\norm{\widehat{\tp}_{Y|X}(Y=k)-\tp_{Y|X}(Y=k)}\big]^{1/2}$, and the last inequality comes from \eqref{eq: thm 2 prop 2} and Lemma \ref{lem: finite bound}. Similarly we can prove the bound for $|\widehat{G}^{\textup{ER}}(\hblambda) - G(\blambda^*)|$.

\spacingset{0.87}
{\small\bibliographyapp{reference_np.bib}}
\bibliographystyleapp{apalike}



\end{document}